\documentclass[lettersize,journal]{IEEEtran}
\usepackage{amsmath,amsfonts}
\usepackage{amssymb} 
\usepackage{algorithmic}
\usepackage{algorithm}
\usepackage{textcomp}
\usepackage{stfloats}
\usepackage{url}
\usepackage{verbatim}
\usepackage{graphicx}
\usepackage{cite}
\usepackage{amsmath}
\usepackage[colorlinks,linkcolor=blue]{hyperref}
\usepackage{verbatim}
\usepackage{multirow} 
\usepackage{xcolor}
\usepackage{hyperref}
\usepackage{threeparttable} 
\usepackage{subcaption}
\usepackage{color}
\usepackage{booktabs}
\usepackage{makecell}
\usepackage{graphics}
\usepackage{array}

\newcommand{\etal}{\emph{et~al.~}}
\newcommand{\ie}{\emph{i.e.~}}
\newcommand{\wrt}{\emph{w.r.t.~}}
\newcommand{\eg}{\emph{e.g.~}}
\newcommand{\etc}{\emph{etc}}

\begin{document}

\title{CBA: Contextual Background Attack against \\Optical Aerial Detection in the Physical World}

\author{Jiawei~Lian,~\IEEEmembership{Graduate Student Member,~IEEE},
        Xiaofei~Wang, 
        Yuru~Su, 
        Mingyang~Ma,~\IEEEmembership{Graduate Student Member,~IEEE},
        and~Shaohui~Mei,~\IEEEmembership{Senior Member,~IEEE}

\thanks{This work was supported in part by the National Natural Science Foundation of China (62171381) and in part by the Fundamental Research Funds for the Central Universities. (Corresponding author: Shaohui Mei.)}

\thanks{Jiawei Lian, Xiaofei Wang, Yuru Su, Mingyang Ma, and Shaohui Mei are with the School of Electronics and Information, Northwestern Polytechnical University, Xi'an 710129, China (Email: lianjiawei@mail.nwpu.edu.cn; wangxiaofei2022@mail.nwpu.edu.cn; suyuru\underline{~}nwpu@mail.nwpu.edu.cn; mamingyang@mail.nwpu.edu.cn; meish@nwpu.edu.cn).}
}

\markboth{Journal of \LaTeX\ Class Files,~Vol.~14, No.~8, August~2021}%
{Shell \MakeLowercase{\textit{et al.}}: A Sample Article Using IEEEtran.cls for IEEE Journals}


\maketitle

\begin{abstract}

   Patch-based physical attacks have increasingly aroused concerns. 
   However, most existing methods focus on obscuring targets captured on the ground, and some of these methods are simply extended to deceive aerial detectors.
   They smear the targeted objects in the physical world with the elaborated adversarial patches, which can only slightly sway the aerial detectors' prediction and with weak attack transferability.
   To address the above issues, we propose to perform Contextual Background Attack (CBA), a novel physical attack framework against aerial detection, which can achieve strong attack efficacy and transferability in the physical world even without smudging the interested objects at all.
   Specifically, the targets of interest, \ie the aircraft in aerial images, are adopted to mask adversarial patches.
   The pixels outside the mask area are optimized to make the generated adversarial patches closely cover the critical contextual background area for detection, which contributes to gifting adversarial patches with more robust and transferable attack potency in the real world.
   To further strengthen the attack performance, the adversarial patches are forced to be outside targets during training, by which the detected objects of interest, both on and outside patches, benefit the accumulation of attack efficacy. 
   Consequently, the sophisticatedly designed patches are gifted with solid fooling efficacy against objects both on and outside the adversarial patches simultaneously.
   Extensive proportionally scaled experiments are performed in physical scenarios, demonstrating the superiority and potential of the proposed framework for physical attacks.
   We expect that the proposed physical attack method will serve as a benchmark for assessing the adversarial robustness of diverse aerial detectors and defense methods.
   The code has been released at \url{https://github.com/JiaweiLian/CBA}.

\end{abstract}


\section{Introduction}

\IEEEPARstart{D}{eep} neural networks (DNNs) have shown great potency over the past few years. 
However, some works \cite{szegedy2014intriguing,ian2015explaining} have demonstrated that adversarial examples can easily deceive DNNs.
When some elaborated human-unperceivable perturbations are added to the clean image \cite{dong2019efficient,deng2022frequency,xie2017adversarial,chen2017zoo,dong2018boosting,mahmood2021robustness,moosavi2016deepfool,carlini2017towards,madry2018towards,chen2022adversarial}, DNNs will generate a completely different wrong prediction, which poses extreme concerns for some security-critical applications.
Consequently, the adversarial attack has increasingly garnered attention since it helps to further understand the vulnerability and interpretability of DNNs by delving into negative examples.
Moreover, the study of malicious examples also provides ideas and data for improving the adversarial robustness of DNNs.

\begin{figure}
  \centering
  \includegraphics[width=0.98\linewidth]{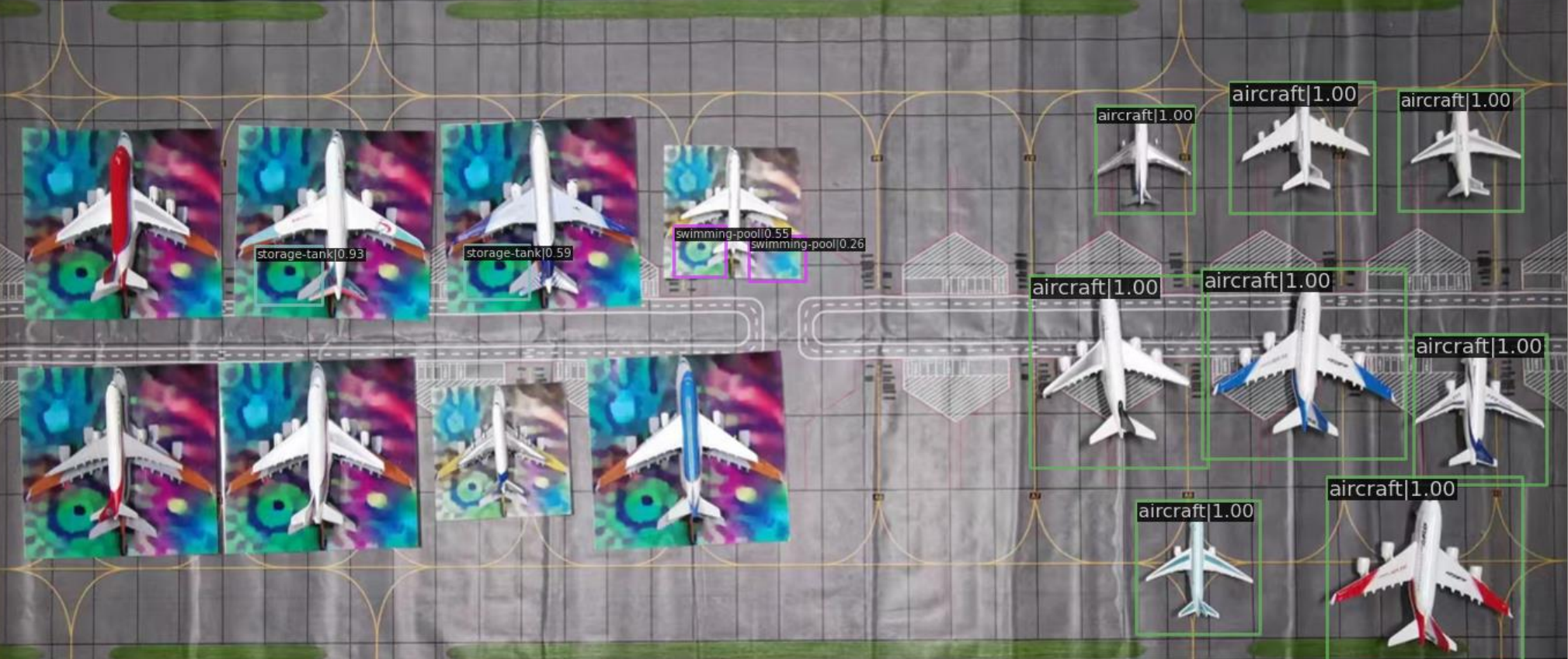}
  \caption{Adversarial attack performance against aerial detection in the physical world, in which the specified targets (left part) are hidden from being detected, while the unmodified targets (right part) are detected correctly.}
  \label{fig:detection_results_comparison}
\end{figure}

\begin{figure}[!t]
\centering
\includegraphics[width=0.98\linewidth]{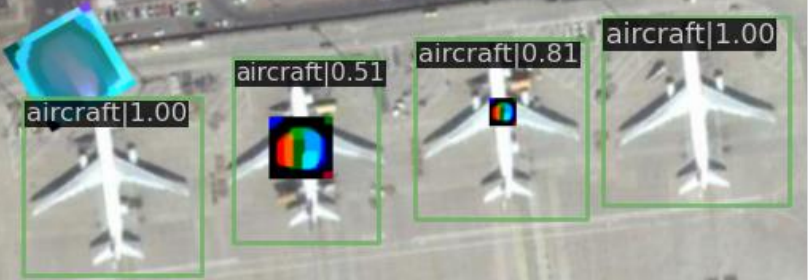}
\caption{
The comparison of different patch settings. From left to right represent patch outside target, big patch, small patch, and no patch, respectively.
}
\label{fig:patch_small_big_outside}
\end{figure}

Nowadays, aerial detection is indispensable and widely used in real scenarios, such as environmental surveillance \cite{mao2021remote}, aerial search and rescue \cite{mei2023rotation}, surveying and mapping \cite{liu2021finer}, \etc. 
Unfortunately, the vulnerability toward adversarial samples also exists in aerial detectors \cite{du2022physical,xu2022universal,lian2022benchmarking}, as shown in Fig. \ref{fig:detection_results_comparison}. 
Nonetheless, most of the existing attacks \cite{xu2020assessing,xu2022universal} are designed for digital attack \cite{chen2022adversarial,vellaichamy2022detectordetective,cai2022zero,shi2022query}.
A few physical attacks against aerial detection \cite{du2022physical,lian2022benchmarking} are directly derived from general real-scenario attack methods \cite{wang2019advpattern,hu2021naturalistic,thys2019fooling}, which focus on hiding particular objects captured on the ground from being detected by placing the malicious patch on the targeted objects, such as persons, traffic signs, cars, \etc.
Regarding aerial detection, most targets are smaller compared to other natural images.
Hence, capturing the tiny patch's negative pattern is challenging. Meanwhile, enlarging the patch size would cause severe occlusion issues, as shown in Fig. \ref{fig:patch_small_big_outside}. 
Recently, some works \cite{lian2022benchmarking,du2022physical} have attempted to perform attacks with patches outside the target, which may be successful but the attack efficacy is unsatisfied and unstable.

To solve the above problems, we propose an innovative physical attack approach called Contextual Background Attack (CBA),
where the targeted objects are protected from being identified by using contextual background adversarial patches.
Specifically, the shape of the protected targets, such as aircraft in aerial detection, is extracted to mask the protected object to design a contextual background patch embedded with the interested target, in which the pixels of the background area are optimized iteratively during the training process. 
Moreover, we devise a novel training strategy in which the patches are put outside targets so that the perceived targets, both in and outside patches, are adopted to calculate the gradients and optimize the contextual adversarial patch. 
Given the shortage of a standardized benchmark to assess physical attacks against aerial detection, extensive experiments in both digital and physical domains are conducted to verify the effectiveness of the proposed CBA and evaluate the adversarial robustness of various aerial detectors.

In summary, our contributions are four-fold as follows:
\begin{itemize}
\setlength{\itemsep}{1pt }
  \item A brand-new Contextual Background Attack (CBA) framework is devised to deceive aerial detection methods in the physical world, which can gift the contextual background patches with SOTA attack performance in both white-box and black-box settings and no need to smear the protected targets.
  \item A novel training strategy is proposed to elaborate adversarial perturbations in the contextual background area. The generated contextual adversarial patches are masked by the interested objects and can simultaneously hide objects both on and outside the adversarial patch from being recognized.
  \item To the best of our knowledge, we are the first to benchmark physical attacks against aerial detection, in which rigorous and exhaustive tests are conducted to evaluate the adversarial robustness of various aerial detectors, and the adversarial patch set is public.
  \item Comprehensive proportionally scaled experiments are conducted in the physical world, demonstrating the substantial physical attack effect and transferability of the elaborated adversarial patches in the contextual background.
\end{itemize}

This work extends our previous conference version \cite{lian2023contextual} in the following aspects.
First, we provide more details and a deep analysis of our CBA.
Second, comprehensive experiments are conducted in both white-box and black-box settings, and proxy models are extended from 4 to 20.
Third, the attack efficacy is extensively validated in both the digital and physical domains.
Finally, the physical robustness is thoroughly verified via exhaustive experiments in the physical world.

The rest part of this article is organized as follows. Section II reviews the related work of 
 physical adversarial attacks in detail. Next, we introduce the details of the proposed Contextual Background Attack framework CBA for generating contextual background adversarial patches against aerial detection tasks in Section III. Then, we verify the effectiveness of the proposed CBA and demonstrate the advantages of the generated contextual background patches in Section IV. Finally, we conclude our proposed CBA method and discuss potential future work concerning patch-based physical attacks in Section V.

\section{Related Work}

In this part, the related works concerning digital attacks are briefly introduced at first.
Subsequently, we review the typical physical attack methods and physical attacks against aerial detection in detail.

\subsection{Digital Attack}

Digital attack methods can be categorized as optimization-based and gradient-based according to how the adversarial perturbations are crafted.
Optimization-based Limited memory Broyden-Fletcher-Goldfarb-Shanno (L-BFGS) \cite{szegedy2014intriguing}, Deepfool \cite{moosavi2016deepfool}, C\&W \cite{carlini2017towards}, \etc. conduct attacks via box-constrained mechanisms.
Gradient-based methods, \eg fast gradient sign method (FGSM) \cite{ian2015explaining}, iterative FGSM (I-FGSM) \cite{kurakin2018adversarial}, momentum iterative (MI-FGSM) \cite{dong2018boosting}, and projected gradient descent (PGD) \cite{madry2018towards}, design adversarial perturbations based on the gradient information of models.
The approaches mentioned above are all performed in the white-box conditions, \ie the training data, victim model structure, and victim model parameters are available to the attackers.
However, the imperceptible adversarial perturbations generated by digital attack algorithms are utterly useless for physical attacks, for the reason that imaging devices can barely capture the indistinguishable noises. 
Consequently, physical attack methods are progressively garnering attention.

\subsection{Physical Attack}

\subsubsection{General Physical Attack Methods}

Adversarial patches \cite{brown2017adversarial} are widely used for physical attacks, such as face recognition \cite{sharif2016accessorize,zheng2023robust,wei2022adversarial}, object detection \cite{labarbarie2022benchmarking,wang2022fca,hu2021naturalistic,thys2019fooling,wei2022simultaneously}, autonomous driving \cite{han2022physical,cheng2022physical}, \etc. 
We review the related work according to application domains as follows:

\textbf{Face recognition}: 
Sharif \etal \cite{sharif2016accessorize} developed a systematic method to generate attacks realized by printing a pair of eyeglass frames. 
In \cite{wei2022adversarial}, the authors proposed another kind of adversarial patch: Meaningful Adversarial Sticker, a physically feasible and stealthy attack method by using actual stickers existing in our life.
Dubbed PadvFace \cite{zheng2023robust} framework was devised to model the challenging physical variations precisely. 
In addition, some other physical attacks \cite{kaziakhmedov2019real,pautov2019adversarial,komkov2021advhat,nguyen2020adversarial} are also devised to deceive face recognition systems.

\textbf{Object detection}:
Hu \etal \cite{hu2021naturalistic} proposed a method to craft natural-looking adversarial patches by leveraging the learned image manifold of a pretrained GAN upon real-world images.
In \cite{labarbarie2022benchmarking}, the authors proposed an evaluation framework for patch attacks against object detectors. 
\cite{thys2019fooling} introduced an approach to generate a patch that can successfully hide a person from a person detector. 
To bridge the gap between digital and physical attacks, \cite{wang2022fca} exploited the entire 3D vehicle surface to propose a robust Full-coverage Camouflage Attack. 
In addition, there are some works \cite{zhu2021fooling,zhu2022infrared} focus on fooling thermal infrared pedestrian detection methods.

\textbf{Autonomous driving}:
Cheng \etal \cite{cheng2022physical} proposed an optimization-based method to generate stealthy physical-object-oriented adversarial patches to attack depth estimation.
In \cite{han2022physical}, the authors realized the first physical backdoor attacks on the lane detection system, including two attack methodologies (poison-annotation and clean-annotation) to generate poisoned samples.
\cite{cheng2022physical} adopt an optimization-based approach to craft stealthy physical-object-oriented adversarial patches to fool depth estimation algorithms.
Besides, \cite{huang2020universal,hu2021naturalistic} also delve into camouflaging adversarial patches in the physical world.

\subsubsection{Physical Attacks against Aerial Detection}

DNNs have been broadly adopted to process aerial imagery \cite{mei2021accelerating,mei2021hyperspectral,tian2021adversarial}.
Consequently, Delving into adversarial attacks against aerial detection paves a critical path to better explaining and improving model robustness.
However, most adversarial attack methods \cite{xu2022universal,xu2020assessing,burnel2021generating,cheng2021perturbation,zhang2022adversarial} against aerial detection concentrate on the digital domain.
In contrast, physical attacks against aerial detection are somewhat scarce, while it is more critical and practical.
Du \etal \cite{du2022physical} demonstrated one of the first efforts at physical adversarial attacks on aerial imagery, whereby malicious patches were optimized, fabricated, and installed on or near target objects to reduce the efficacy of an object detector applied on overhead images. 
In \cite{lian2022benchmarking}, a novel adaptive-patch-based physical attack (APPA) framework was proposed to generate adversarial patches adapted to both physical dynamics and varying scales. Furthermore, they devised a new loss to optimize the adversarial patch by entirely using the detection results, which can significantly accelerate the optimizing process.
However, the above attacks against aerial detection are derived from the aforementioned general physical attack methods, which are not aggressive enough and need to smear targets.

\section{Methodology}

\begin{figure*}[!t]
\centering
\includegraphics*[width=0.995\linewidth]{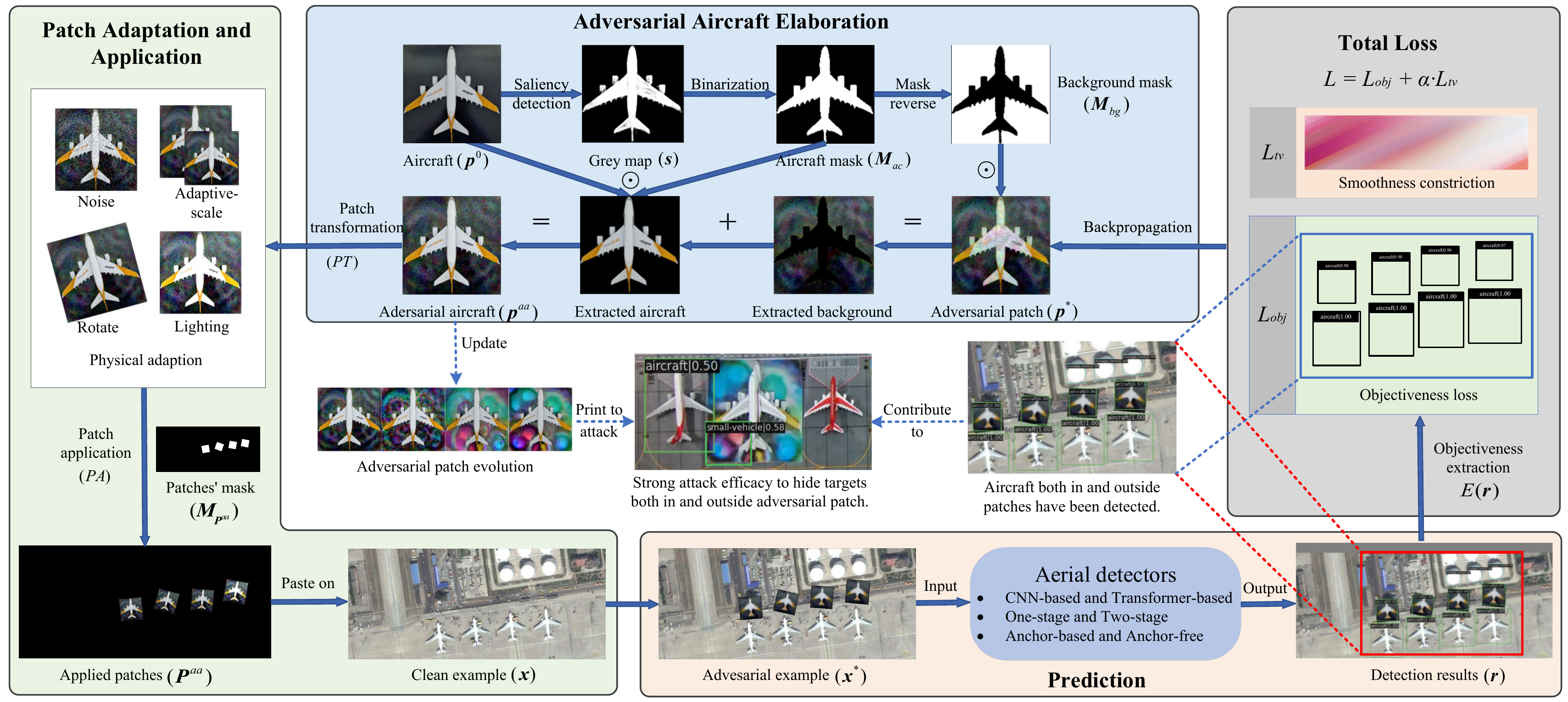}
\caption{
The illustration of the proposed contextual physical attack method.
Firstly, we extract the protected object's saliency map as a mask to separate the foreground and background areas. 
Secondly, augmentations are adopted to adapt the physical dynamics. 
Thirdly, we place the adversarial patches on the clean image in the proper size and location to generate adversarial examples in the digital domain.
Then, the adversarial examples are fed into an aerial detector, and the recognized targets both in and outside the patch are used to optimize the contextual adversarial patch.
Finally, we use the background pixels of the adversarial patch plus the extracted foreground to update the contextual adversarial patch, \ie we only optimize the background pixels of the adversarial aircraft during the training process.
}
\label{fig:pipeline}
\end{figure*}

In this section, we formulate the problem and then elaborate on the proposed CBA with aircraft as the targets of interest.
It is believed that CBA can also work for other interested targets similarly.
The overall pipeline is displayed in Fig. \ref{fig:pipeline}.

\subsection{Problem Formulation}

Given a benign aerial image $\boldsymbol{x}$, the attack purpose in aerial detection is to hide the specified targets from being detected by pasting elaborated adversarial patches on the clean image. 
Specifically, the adversarial example $\boldsymbol{x}^*$ with adversarial patches $\boldsymbol{P}^*$ can be defined as:
\begin{equation}
    \label{eq:problem-formulation}
    \boldsymbol{x}^* = (\boldsymbol{1}-\boldsymbol{M}_{\boldsymbol{P}^*}) \odot \boldsymbol{x} + \boldsymbol{M}_{\boldsymbol{P}^*} \odot \boldsymbol{P}^*,
\end{equation}
where $\boldsymbol{M}_{\boldsymbol{p}^*}$ (the pixel values of the foreground are 1, the rest are 0) and $\odot$ represent the mask of adversarial patches and Hadamard product, respectively. 

The previous approaches focus on optimizing all pixels of an adversarial patch by putting it on or outside objects, as shown in Fig. \ref{fig:patch_small_big_outside}. 
In comparison, our method designs the contextual background adversarial patch embedded with a specified shape to match the protected target, as shown in Fig. \ref{fig:pipeline}, which can fool various aerial detectors (CNN-based and Transformer-based, One-stage and Two-stage, Anchor-based and Anchor-free) when the targeted object is placed on the contextual adversarial background in the physical world.
In the following sections, we will systematically introduce how to elaborate the contextual adversarial background with aircraft as the protected objects, \ie adversarial aircraft, by our proposed framework CBA.

\begin{figure}[!t]
\centering
\includegraphics[width=0.98\linewidth]{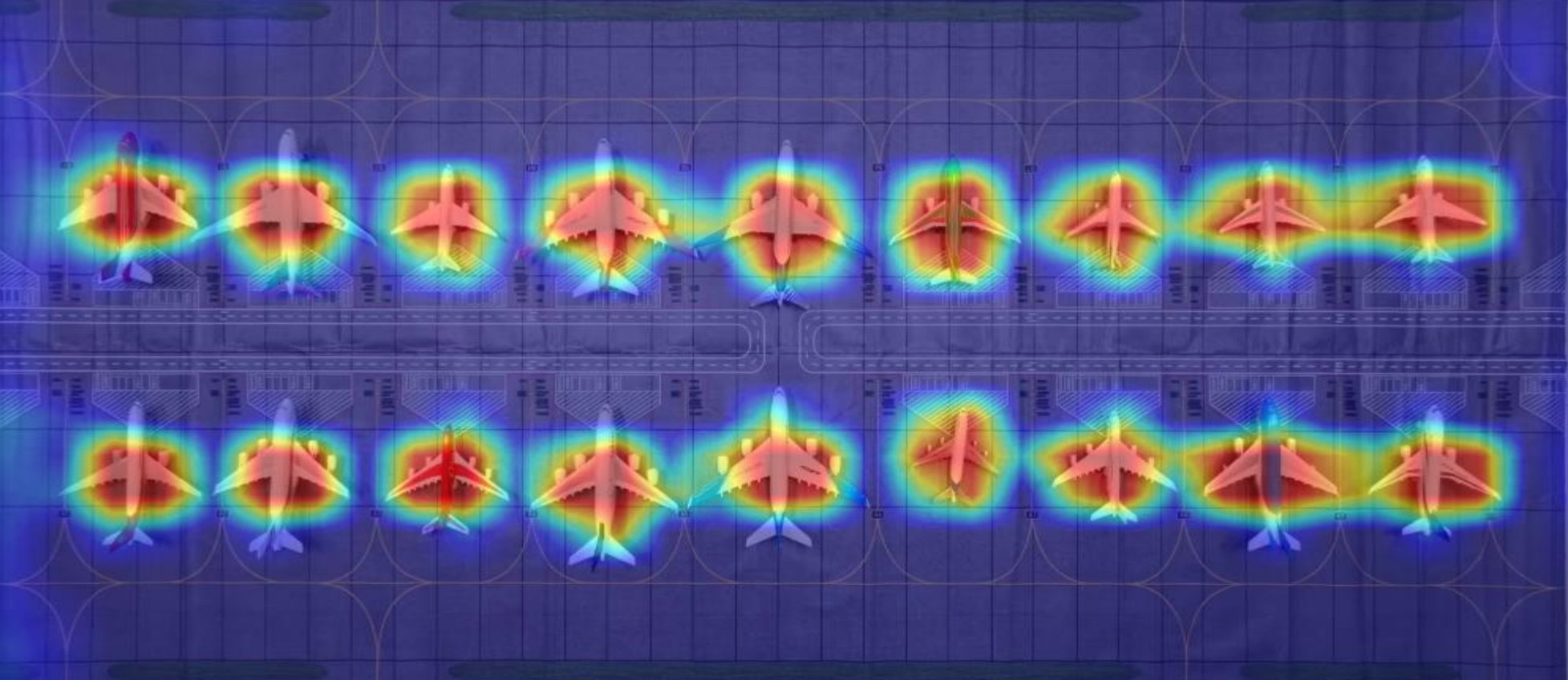}
\caption{
Visualization of the aerial detector's attention on the benign image.
}
\label{fig:cam_clean}
\end{figure}

\subsection{Adversarial Aircraft Elaboration}

Based on the previous work \cite{thys2019fooling,lian2022benchmarking,selvaraju2017grad}, the following observations can be obtained:
\begin{itemize}
    \item The bigger the adversarial patch size, the stronger the attack efficacy;
    \item The closer distance between the adversarial patch and the targeted object, the stronger the attack efficacy;
    \item According to the attention maps, as shown in Fig. \ref{fig:cam_clean}, the contextual background area plays a key role during detection.
\end{itemize}

Therefore, to achieve more robust attack efficacy,
we propose to perform contextual background attacks in the aerial detection task, by which we can generate adversarial patches as big as aircraft without taking up extra area. 
In addition, our method can make the adversarial patches as close as possible to the targeted objects, and no need to smear or obscure them.

Technically, we first take a picture of the aircraft model on a black backdrop to get the original patch $\boldsymbol{p}^0$.
Secondly, a saliency detector \cite{zhuge2022salient} is adopted to extract the saliency map of the aircraft $\boldsymbol{s}$.
Thirdly, the saliency map $\boldsymbol{s}$ is binarized as an aircraft mask $\boldsymbol{M}_{ac}$. 
Consequently, the background mask of the adversarial patch is defined as:
\begin{equation}
    \label{eq:mask_background}
    \boldsymbol{M}_{bg} = \boldsymbol{1} - \boldsymbol{M}_{ac}.
\end{equation}
Finally, we formulate the adversarial aircraft as follows:
\begin{equation}
    \label{eq:adversarial_aircraft}
    \boldsymbol{p}^{aa}_{i,j} = \boldsymbol{p}^0 \odot \boldsymbol{M}_{ac} + \boldsymbol{p}^*_{i,j} \odot \boldsymbol{M}_{bg},
\end{equation}
where $\boldsymbol{p}^*_{i,j}$ represents the optimized adversarial patch, and $i$, $j$ is the index of epochs and iterations during training, respectively. It can be found that the background pixels are used to update the adversary aircraft, as shown in Fig. \ref{fig:pipeline}. 

\subsection{Patch Adaption and Application}

Since our proposed CBA aims to generate an adversarial patch with solid attack efficacy in physical scenarios,
physical accommodations are adopted to simulate the real-scenario dynamics, including different noises, adaptive scales, random rotations, and varying lighting.
The above physical adaptations are bundled in the patch transformation function $PT$ similar with \cite{thys2019fooling,lian2022benchmarking}, and the augmented patches are shown in the top left part of Fig. \ref{fig:pipeline}.

Next, the elaborated adversarial patch must be pasted on the benign image with the proper size and location. 
Usually, the patch should be placed where it works in actual applications during the training process.
Thus, we try to place the adversarial patch in an aircraft shape (Extracted background area as shown in Fig. \ref{fig:pipeline}).
Technically, we adopt an oriented aerial detector \cite{xie2021oriented} to recognize aircraft and its orientation information. However, the extracted patch still cannot match the aircraft precisely, and severe occlusion exists due to the intra-class differences between aircraft and view angles.
Consequently, we propose a novel training strategy to overcome the above difficulties, putting the adversarial aircraft patches outside targets during the training, as the adversarial example shown in Fig. \ref{fig:pipeline}.
In this way, the aircraft both in and outside the patches can be detected, as shown in Fig. \ref{fig:pipeline}, which significantly contributes to strengthening the attack efficacy to hide targets both in and outside the adversarial patches.

Specifically, we place the adversarial patches outside targets at the proper distance and size, which is adaptive to the position and size of the targets according to the ground truth $\boldsymbol{y} = (x_1,y_1,x_2,y_2, class)$.
The coordination $\boldsymbol{l}_{\boldsymbol{p}^{aa}}$ and size $(w_{\boldsymbol{p}^{aa}}, h_{\boldsymbol{p}^{aa}})$ of the square adversarial patch are calculated similarly with \cite{lian2022benchmarking} as follows:
\begin{equation}
    \label{eq:patch_location}
    \boldsymbol{l}_{\boldsymbol{p}^{aa}} = (\frac{x_1+x_2}{2},\frac{y_1+y_2}{2} - \frac{y_2 - y_1}{r_d}),
\end{equation}
\begin{equation}
    \label{eq:patch_size}
    w_{\boldsymbol{p}^{aa}} =  h_{\boldsymbol{p}^{aa}} = \sqrt[2]{r_s \cdot w_{\boldsymbol{t}} \cdot h_{\boldsymbol{t}}},
\end{equation}
where $r_d$ and $r_s$ are coefficients for adaptively adjusting the patch distance and size. 
Then, we acquire the mask of the adversarial aircraft $\boldsymbol{M}_{\boldsymbol{P}^{aa}}$ and applied patches $\boldsymbol{P}^{aa}$ by putting ${PT(\boldsymbol{p}^{aa}})$ in proper size and location according to $\boldsymbol{l}_{\boldsymbol{p}^{aa}},w_{\boldsymbol{p}^{aa}},h_{\boldsymbol{p}^{aa}}$, which is formulated as $PA$:
\begin{equation}
    \label{eq:patch_adverarial_aircraft_mask}
    \left[\boldsymbol{P}^{aa}, \boldsymbol{M}_{\boldsymbol{P}^{aa}}\right] = PA({PT(\boldsymbol{p}^{aa}}),\boldsymbol{l}_{\boldsymbol{p}^{aa}},w_{\boldsymbol{p}^{aa}},h_{\boldsymbol{p}^{aa}}).
\end{equation}

Finally, we transform the initial formulation Eq. \eqref{eq:problem-formulation} as
\begin{equation}
    \label{eq:final_problem_formulation}
    \begin{aligned}
    \boldsymbol{x}^* = (\boldsymbol{1}-\boldsymbol{M}_{\boldsymbol{P}^{aa}}) \odot \boldsymbol{x} + \boldsymbol{M}_{\boldsymbol{P}^{aa}} \odot \boldsymbol{P}^{aa}.
    \end{aligned}
\end{equation}

\subsection{Loss Design}

In this paper, we aim to perform untargeted physical attacks, \ie to hide the specified targets from being detected in the physical scenarios. Therefore, the objectiveness function comprises adversarial objectiveness loss and smoothness constriction.

\textbf{Adversarial objectiveness}:
We thoroughly use the detected objects in and outside adversarial patches to optimize the patch, as shown in Fig. \ref{fig:pipeline}.
Specifically, all the objectiveness scores from detection results are taken into account, which is written as:
\begin{equation}
    \label{eq:Lobj}
    L_{obj} = E(\boldsymbol{r}) = \frac{1}{n} \sum\limits_{i=1}^n P_i(obj),
\end{equation}
where $n$ means the number of detected objects. The detection results $\boldsymbol{r}$ usually contain coordinates $(x_1,y_1,x_2,y_2)$, the objective score $P(obj)$, and the class probability such as $(P(aircraft),P(ship),...,P(bridge),P(harbor))$ of each object. $E(\boldsymbol{r})$ represents extracting objectiveness loss $L_{obj}$ from $\boldsymbol{r}$. The adversarial objectiveness loss gifts the adversarial patch with attack efficacy during the optimization.

\begin{algorithm}
\caption{Contextual Background Attack (CBA)}
\label{alg:A3}
\begin{algorithmic}[1] 
\REQUIRE Detector $D$, benign aerial image $\boldsymbol{x}$ and ground truth $\boldsymbol{y}$, original patch $\boldsymbol{p}^0$, the adversarial attack loss function $L$, the number of epochs $N_{epc}$ and the number of iterations of each epoch $N_{itr}$, hyperparameter $\alpha, \eta$.
\ENSURE Adversarial aircraft $\boldsymbol{p}^*$.
    \STATE $\boldsymbol{p}^{aa} = \boldsymbol{p}^0$;
    \STATE Extract aircraft saliency map $\boldsymbol{s}$ from $\boldsymbol{p}^0$;
    \STATE Binarize $\boldsymbol{s}$ to get the aircraft mask $\boldsymbol{M}_{ac}$;
    \STATE Patch's background mask: $\boldsymbol{M}_{bg} = \boldsymbol{1} - \boldsymbol{M}_{ac}$;
    \FOR{$i=0$ to $N_{epc}$}
        \FOR{$j=0$ to $N_{itr}$}
            \STATE $\boldsymbol{p}^{aa}_{i,j} = \boldsymbol{p}^0 \odot \boldsymbol{M}_{ac} + \boldsymbol{p}^*_{i,j} \odot \boldsymbol{M}_{bg}$;
            \STATE $\boldsymbol{p}^{aa}_{i,j} = PT({\boldsymbol{p}^{aa}_{i,j}},\boldsymbol{y})$;
            \STATE $\big[\boldsymbol{P}^{aa}_{i,j}, \boldsymbol{M}_{\boldsymbol{P}^{aa}_{i,j}}\big] = PA({\boldsymbol{p}^{aa}_{i,j}},\boldsymbol{l}_{\boldsymbol{p}^{aa}_{i,j}},w_{\boldsymbol{p}^{aa}_{i,j}},h_{\boldsymbol{p}^{aa}_{i,j}})$;
            \STATE $\boldsymbol{x}^* = (\boldsymbol{1}-\boldsymbol{M}_{\boldsymbol{P}^{aa}_{i,j}}) \odot \boldsymbol{x} + \boldsymbol{M}_{\boldsymbol{P}^{aa}_{i,j}} \odot \boldsymbol{P}^{aa}_{i,j}$;
            \STATE $\boldsymbol{r} = D(\boldsymbol{x}^*)$;
            \STATE $L_{obj} = E(\boldsymbol{r})$;
            \STATE $L = L_{obj} + \alpha \cdot L_{tv}$;
            \STATE $\boldsymbol{p}^*_{i,j+1} = \boldsymbol{p}^{aa}_{i,j} + \eta \cdot \nabla_{\boldsymbol{p}^{aa}_{i,j}}L$;
        \ENDFOR
    \ENDFOR
    \STATE $\boldsymbol{p}^{aa} = \boldsymbol{p}^{aa}_{N_{epc},N_{itr}}$;
    \STATE \textbf{return} $\boldsymbol{p}^{aa}$.
\end{algorithmic}
\end{algorithm}

\textbf{Smoothness constriction}:
For the physical attack, it is hard for detection systems to capture the gap between adjacent pixels. Hence, total variation \cite{sharif2016accessorize} is adopted to constraint the smoothness of the generated adversarial patch, which can be written as: 
\begin{equation}
    \label{eq:Ltv}
    L_{tv} = \sum\limits_{m,n} \sqrt{(\boldsymbol{p}^{aa}_{m+1,n}-\boldsymbol{p}^{aa}_{m,n})^2 + (\boldsymbol{p}^{aa}_{m,n+1}-\boldsymbol{p}^{aa}_{m,n})^2},
\end{equation}
where $\boldsymbol{p}^{aa}_{m,n}$ means the pixel value of the adversarial aircraft. Total variation is indispensable for its key role in maintaining the attack efficacy of the adversarial patch during the physical-digital transformation in physical attacks.

Finally, the overall objectiveness loss is as follows:
\begin{equation}
    \label{eq:total_loss}
    L = L_{obj} + \alpha \cdot L_{tv},
\end{equation}
where $\alpha$ is a balance parameter. The detailed discussion \wrt the selection of $\alpha$ will be presented in Sec. \ref{section4.4}. 

\subsection{Overall Training Procedures}

In this section, we choose aircraft as the targeted object to describe the overall training procedures of the proposed CBA as shown in Algorithm \ref{alg:A3}. 
The comprehensive explanation of the algorithm is given as follows:

1) Extract the saliency map $\boldsymbol{s}$ of the protected object, \ie the original patch $\boldsymbol{p}^0$, to mask the protected aircraft;

2) We adopt the extracted masks $\boldsymbol{M}_{ac}$ and $\boldsymbol{M}_{bg}$ to formulate adversarial aircraft $\boldsymbol{p}^{aa}$ as the contextual adversarial patch;

3) The adversarial aircraft are transformed by $PT$ to accommodate dynamic physical conditions and varying size targets, then these patches are pasted on the clean image in the appropriate position and adaptive size by $PA$;

4) The victim aerial detector takes the elaborated adversarial example $\boldsymbol{x}^*$ as input to make a prediction;

5) The objectiveness loss is extracted from detection results $\boldsymbol{r}$ by the function $E$;

6) We use the extracted objectiveness loss plus total variation loss to calculate the gradients concerning 
the contextual adversarial aircraft, which are adopted to optimize the pixel values of the contextual adversarial aircraft $\boldsymbol{p}^{aa}$;

7) Finally, repeat steps 2 to 6 until the end of training.






\section{Experiments}

\begin{figure*}
  \centering
  \begin{subfigure}{0.093\linewidth}
    \includegraphics[width=1\linewidth]{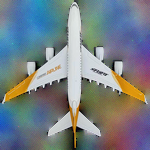}
    \captionsetup{font=tiny}
    \caption{YOLOv2}
  \end{subfigure}
  \begin{subfigure}{0.093\linewidth}
    \includegraphics[width=1\linewidth]{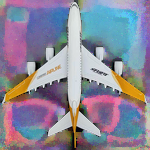}
    \captionsetup{font=tiny}
    \caption{YOLOv3}
  \end{subfigure}
  \begin{subfigure}{0.093\linewidth}
    \includegraphics[width=1\linewidth]{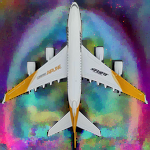}
    \captionsetup{font=tiny}
    \caption{YOLOv5n}
  \end{subfigure}
  \begin{subfigure}{0.093\linewidth}
    \includegraphics[width=1\linewidth]{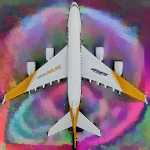}
    \captionsetup{font=tiny}
    \caption{YOLOv5s}
  \end{subfigure}
  \begin{subfigure}{0.093\linewidth}
    \includegraphics[width=1\linewidth]{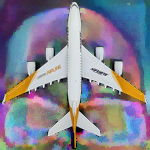}
    \captionsetup{font=tiny}
    \caption{YOLOv5m}
  \end{subfigure}
  \begin{subfigure}{0.093\linewidth}
    \includegraphics[width=1\linewidth]{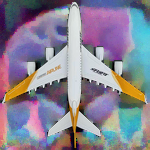}
    \captionsetup{font=tiny}
    \caption{YOLOv5l}
  \end{subfigure}
  \begin{subfigure}{0.093\linewidth}
    \includegraphics[width=1\linewidth]{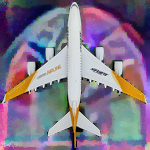}
    \captionsetup{font=tiny}
    \caption{YOLOv5x}
  \end{subfigure}
  \begin{subfigure}{0.093\linewidth}
    \includegraphics[width=1\linewidth]{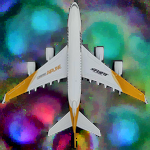}
    \captionsetup{font=tiny}
    \caption{SSD}
  \end{subfigure}
  \begin{subfigure}{0.093\linewidth}
    \includegraphics[width=1\linewidth]{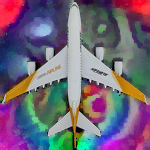}
    \captionsetup{font=tiny}
    \caption{Faster R-CNN}
  \end{subfigure}
  \begin{subfigure}{0.093\linewidth}
    \includegraphics[width=1\linewidth]{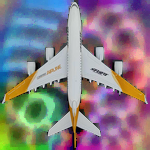}
    \captionsetup{font=tiny}
    \caption{Swin Transformer}
  \end{subfigure}
  \begin{subfigure}{0.093\linewidth}
    \includegraphics[width=1\linewidth]{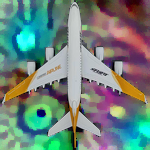}
    \captionsetup{font=tiny}
    \caption{Cascade R-CNN}
  \end{subfigure}
  \begin{subfigure}{0.093\linewidth}
    \includegraphics[width=1\linewidth]{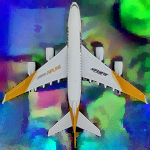}
    \captionsetup{font=tiny}
    \caption{RetinaNet}
  \end{subfigure}
  \begin{subfigure}{0.093\linewidth}
    \includegraphics[width=1\linewidth]{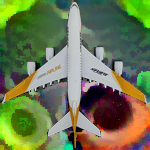}
    \captionsetup{font=tiny}
    \caption{Mask R-CNN}
  \end{subfigure}
  \begin{subfigure}{0.093\linewidth}
    \includegraphics[width=1\linewidth]{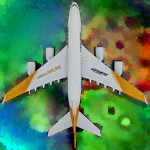}
    \captionsetup{font=tiny}
    \caption{FoveaBox}
  \end{subfigure}
  \begin{subfigure}{0.093\linewidth}
    \includegraphics[width=1\linewidth]{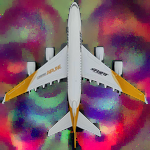}
    \captionsetup{font=tiny}
    \caption{FreeAnchor}
  \end{subfigure}
  \begin{subfigure}{0.093\linewidth}
    \includegraphics[width=1\linewidth]{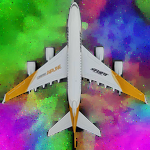}
    \captionsetup{font=tiny}
    \caption{FSAF}
  \end{subfigure}
  \begin{subfigure}{0.093\linewidth}
    \includegraphics[width=1\linewidth]{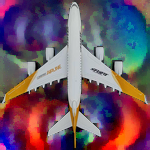}
    \captionsetup{font=tiny}
    \caption{RepPoints}
  \end{subfigure}
  \begin{subfigure}{0.093\linewidth}
    \includegraphics[width=1\linewidth]{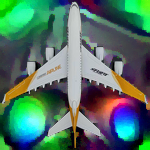}
    \captionsetup{font=tiny}
    \caption{TOOD}
  \end{subfigure}
  \begin{subfigure}{0.093\linewidth}
    \includegraphics[width=1\linewidth]{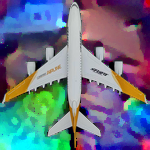}
    \captionsetup{font=tiny}
    \caption{ATSS}
  \end{subfigure}
  \begin{subfigure}{0.093\linewidth}
    \includegraphics[width=1\linewidth]{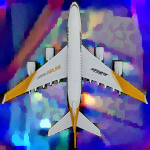}
    \captionsetup{font=tiny}
    \caption{VarifocalNet}
  \end{subfigure}
  \caption{The contextual adversarial patches elaborated by the proposed CBA.}
  \label{fig:adversarial_aircraft}
\end{figure*}

\begin{figure*}[!t]
\centering
\includegraphics*[width=1\linewidth]{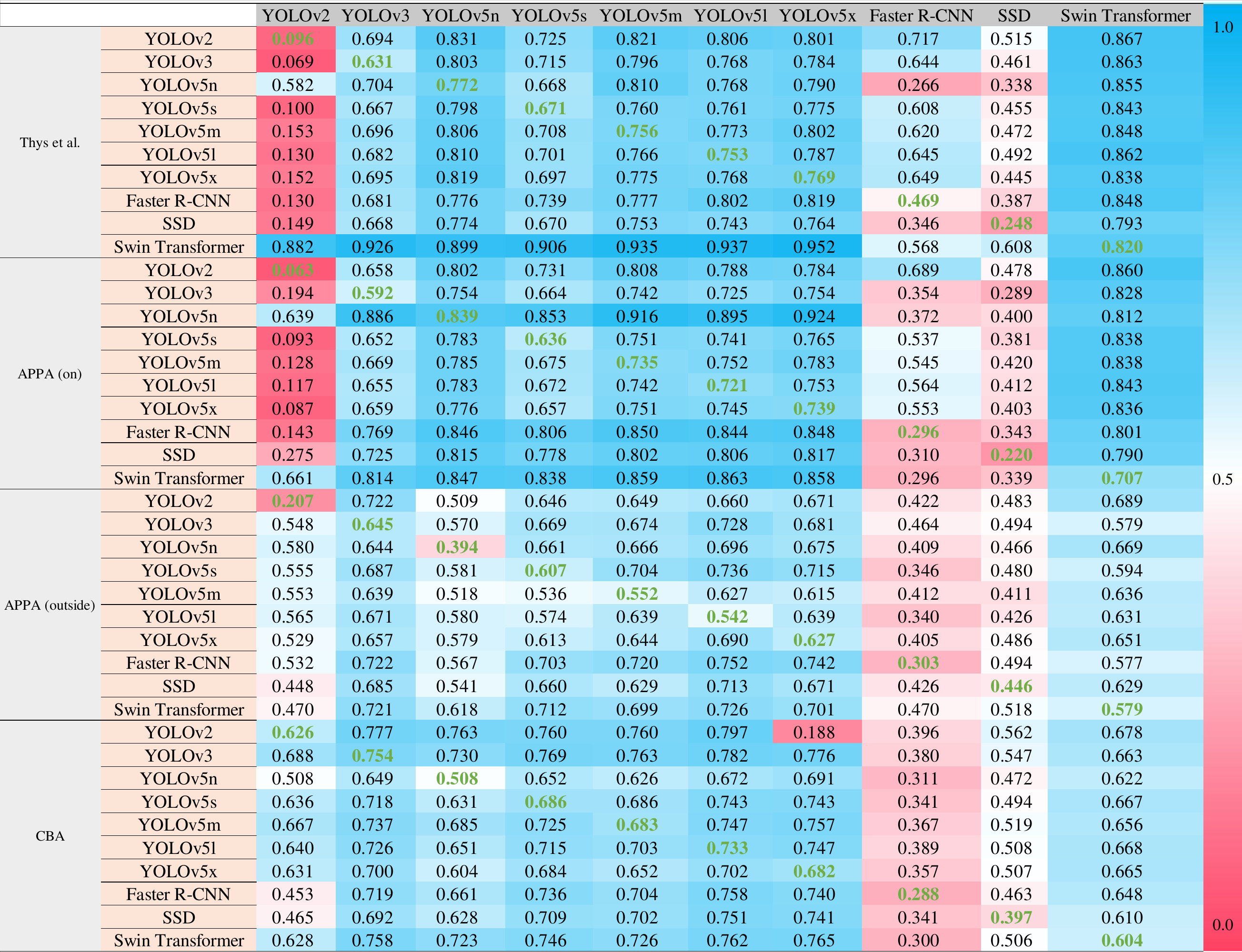}
\caption{
Digital attack results comparison.
Notes: 
1) white-box attacks are highlighted in green; the rest are black-box attacks; 2) `on' and `outside' mean putting patches on or outside targets in APPA; 3) the orange and dark grey areas represent detectors for training adversarial patches and tests, respectively, and light grey means attack methods;
4) numbers are color-filled according to their values, which means that the redder the color, the stronger the attack performance; the bluer the color, the worse the attack performance; the redder and smoother the color of each row, the better the attack transferability.
}
\label{fig:table_digital_results}
\end{figure*}

\begin{figure*}[!t]
\centering
\includegraphics*[width=1\linewidth]{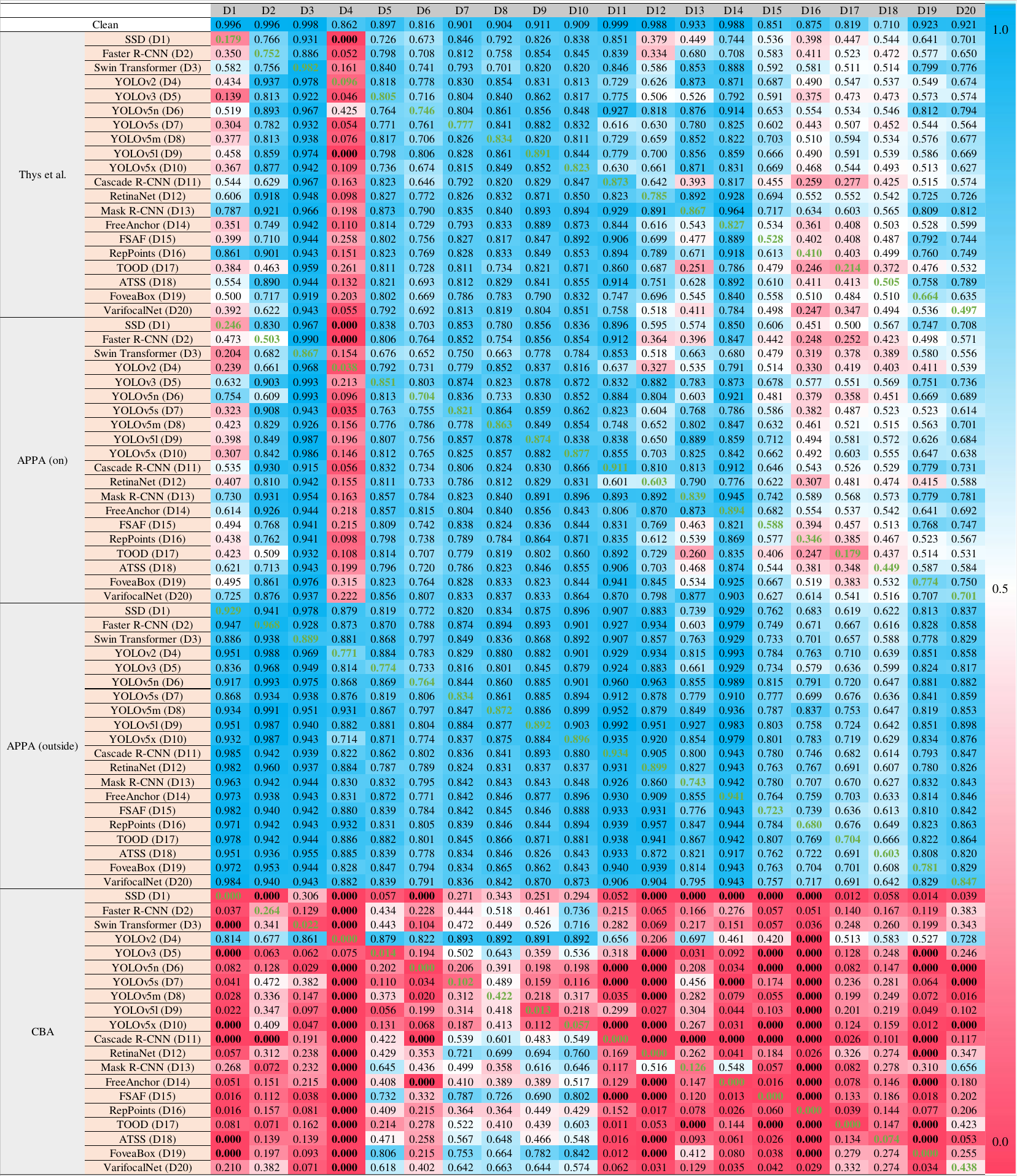}
\caption{
Physical attack results in comparison. 
Notes: 
1) white-box attacks are highlighted in green; the rest are black-box attacks; 2) `on' and `outside' mean putting patches on or outside targets in APPA; 3) the orange and dark grey areas represent detectors for training adversarial patches and tests, respectively, and light grey means attack methods;
4) numbers are color-filled according to their values, which means that the redder the color, the stronger the attack performance; the bluer the color, the worse the attack performance; the redder and smoother the color of each row, the better the attack transferability.
}
\label{fig:table_physical_results}
\end{figure*}

\begin{table*}[t!]
\caption{\\\small{DETAILED RESULTS OF WHITE-BOX ATTACKS.}}
\label{table_white_box_attack_results_comparison}
\centering
\setlength{\tabcolsep}{0.92mm}
\begin{threeparttable}
\begin{tabular*}{\hsize}{ccccccccccccccccccccc}
\hline\hline
\diaghead{\theadfont Diag ColumnmnHead}{Detectors}{Targets} &--- &T1 &T2 &T3 &T4 &T5 &T6 &T7 &T8 &T9 &T10 &T11 &T12 &T13 &T14 &T15 &T16 &T17 &T18 &Average 
\\ \hline\hline
\multirow{4}{*}{SSD\cite{liu2016ssd}} & APPA (on)        & 0.00          & 0.00          & 0.00          & 0.00          & 0.00          & 0.00          & 0.00          & 0.45          & 0.00          & 0.00          & 0.22          & 0.40          & 0.22          & 0.94          & 0.40          & 0.84          & 0.31          & 0.65          & 0.246          \\
                     & APPA (outside) & 0.95          & 1.00          & 0.97          & 0.98          & 0.96          & 0.72          & 0.98          & 0.99          & 0.96          & 0.96          & 0.99          & 0.50          & 0.93          & 1.00          & 0.86          & 0.98          & 1.00          & 0.99          & 0.929          \\
                     & Thys \etal      & 0.00 & 0.00          & 0.00          & 0.00          & 0.00          & 0.00          & 0.00          & 0.72          & 0.00          & 0.00          & 0.00          & 0.24          & 0.00          & 0.41          & 0.00          & 0.83          & 0.21          & 0.82          & 0.179          \\
                     & Ours             & \textbf{0.00} & \textbf{0.00} & \textbf{0.00} & \textbf{0.00} & \textbf{0.00} & \textbf{0.00} & \textbf{0.00} & \textbf{0.00} & \textbf{0.00} & \textbf{0.00} & \textbf{0.00} & \textbf{0.00} & \textbf{0.00} & \textbf{0.00} & \textbf{0.00} & \textbf{0.00} & \textbf{0.00} & \textbf{0.00} & \textbf{0.000}
 \\ \hline
 \multirow{4}{*}{Faster R-CNN\cite{ren2015faster}} & APPA (on)        & 0.00          & 0.90          & 0.00          & \textbf{0.00} & \textbf{0.00} & 0.00          & 0.99          & 0.99          & 0.99          & \textbf{0.00} & \textbf{0.66} & 1.00          & 0.55          & 0.72          & 0.27          & 1.00          & 0.00          & 0.99          & 0.503          \\
                              & APPA (outside) & 1.00          & 1.00          & 0.99          & 0.99          & 0.99          & 0.81          & 1.00          & 1.00          & 0.99          & 1.00          & 1.00          & 1.00          & 1.00          & 1.00          & 0.66          & 1.00          & 1.00          & 1.00          & 0.968          \\
                              & Thys \etal      & \textbf{0.00} & 0.99          & 0.97          & 0.21          & 0.37          & 0.00          & 1.00          & 1.00          & 0.98          & 0.76          & 0.95          & 1.00          & 0.98          & \textbf{0.36} & 0.96          & 1.00          & 1.00          & 1.00          & 0.752          \\
                              & Ours             & 0.27          & \textbf{0.33} & \textbf{0.00} & 0.93          & 0.99          & \textbf{0.00} & \textbf{0.00} & \textbf{0.00} & \textbf{0.00} & 0.32          & 0.98          & \textbf{0.00} & \textbf{0.00} & 0.93          & \textbf{0.00} & \textbf{0.00} & \textbf{0.00} & \textbf{0.00} & \textbf{0.264}
\\ \hline
\multirow{4}{*}{Swin Transformer\cite{liu2021swin}} & APPA (on)      & 0.98          & 1.00          & 0.93          & 0.25          & 1.00          & 0.74          & 0.99          & 1.00          & 1.00          & 0.84          & 0.94          & 0.97          & 0.97          & 1.00          & 0.00          & 1.00          & 1.00          & 1.00          & 0.867          \\
                                  & APPA (outside) & 1.00          & 1.00          & 1.00          & 1.00          & 1.00          & 0.00          & 1.00          & 1.00          & 1.00          & 1.00          & 1.00          & 1.00          & 1.00          & 1.00          & 0.00          & 1.00          & 1.00          & 1.00          & 0.889          \\
                                  & Thys \etal    & 0.97          & 1.00          & 1.00          & 0.95          & 1.00          & 0.86          & 1.00          & 1.00          & 1.00          & 1.00          & 1.00          & 1.00          & 1.00          & 1.00          & 0.91          & 0.99          & 1.00          & 1.00          & 0.982          \\
                                  & Ours           & \textbf{0.00} & \textbf{0.40} & \textbf{0.00} & \textbf{0.00} & \textbf{0.00} & \textbf{0.00} & \textbf{0.00} & \textbf{0.00} & \textbf{0.00} & \textbf{0.00} & \textbf{0.00} & \textbf{0.00} & \textbf{0.00} & \textbf{0.00} & \textbf{0.00} & \textbf{0.00} & \textbf{0.00} & \textbf{0.00} & \textbf{0.022}
\\ \hline
\multirow{4}{*}{YOLOv2\cite{redmon2017yolo9000}} & APPA (on)      & 0.00          & 0.00          & 0.00          & 0.00          & 0.00          & 0.00          & 0.00          & 0.00          & 0.00          & 0.00          & 0.00          & 0.69          & 0.00          & 0.00          & 0.00          & 0.00          & 0.00          & 0.00          & 0.038          \\
                        & APPA (outside) & 0.00          & 1.00          & 1.00          & 0.00          & 1.00          & 0.00          & 1.00          & 1.00          & 1.00          & 0.00          & 1.00          & 1.00          & 0.96          & 1.00          & 0.99          & 1.00          & 0.93          & 0.99          & 0.771          \\
                        & Thys \etal    & 0.00          & 0.00          & 0.00          & 0.00          & 0.00          & 0.00          & 0.00          & 0.00          & 0.00          & 0.00          & 0.00          & 0.99          & 0.00          & 0.00          & 0.00          & 0.73          & 0.00          & 0.00          & 0.096          \\
                        & Ours           & \textbf{0.00} & \textbf{0.00} & \textbf{0.00} & \textbf{0.00} & \textbf{0.00} & \textbf{0.00} & \textbf{0.00} & \textbf{0.00} & \textbf{0.00} & \textbf{0.00} & \textbf{0.00} & \textbf{0.00} & \textbf{0.00} & \textbf{0.00} & \textbf{0.00} & \textbf{0.00} & \textbf{0.00} & \textbf{0.00} & \textbf{0.000}
\\ \hline
\multirow{4}{*}{YOLOv3\cite{redmon2018yolov3}} & APPA (on)        & 0.88          & 0.87          & 0.84          & 0.88          & 0.91          & 0.37          & 0.87          & 0.87          & 0.90          & 0.83          & 0.89          & 0.87          & 0.86          & 0.90          & 0.86          & 0.88          & 0.92          & 0.91          & 0.851          \\
                        & APPA (outside) & 0.90          & 0.85          & 0.84          & 0.70          & 0.88          & 0.00          & 0.22          & 0.85          & 0.80          & 0.85          & 0.89          & 0.88          & 0.85          & 0.88          & 0.87          & 0.88          & 0.90          & 0.89          & 0.774          \\
                        & Thys \etal      & 0.86          & 0.88          & 0.79          & 0.86          & 0.90          & 0.43          & 0.80          & 0.89          & 0.87          & 0.83          & 0.82          & 0.87          & 0.36          & 0.89          & 0.76          & 0.89          & 0.88          & 0.91          & 0.805          \\
                        & Ours             & \textbf{0.00} & \textbf{0.00} & \textbf{0.00} & \textbf{0.00} & \textbf{0.00} & \textbf{0.00} & \textbf{0.00} & \textbf{0.00} & \textbf{0.00} & \textbf{0.00} & \textbf{0.00} & \textbf{0.00} & \textbf{0.25} & \textbf{0.00} & \textbf{0.00} & \textbf{0.00} & \textbf{0.00} & \textbf{0.00} & \textbf{0.014}
\\ \hline
\multirow{4}{*}{YOLOv5n\cite{jocher2020yolov5}} & APPA (on)        & 0.00          & 0.81          & 0.62          & 0.62          & 0.88          & 0.00          & 0.82          & 0.85          & 0.85          & 0.82          & 0.87          & 0.84          & 0.77          & 0.76          & 0.61          & 0.83          & 0.86          & 0.86          & 0.704          \\
                         & APPA   (outside) & 0.84          & 0.88          & 0.79          & 0.84          & 0.86          & 0.00          & 0.80          & 0.85          & 0.86          & 0.76          & 0.90          & 0.77          & 0.87          & 0.86          & 0.33          & 0.80          & 0.87          & 0.88          & 0.764          \\
                         & Thys \etal      & 0.82          & 0.77          & 0.75          & 0.50          & 0.83          & 0.00          & 0.84          & 0.84          & 0.87          & 0.73          & 0.80          & 0.86          & 0.82          & 0.82          & 0.60          & 0.83          & 0.87          & 0.87          & 0.746          \\
                         & Ours             & \textbf{0.00} & \textbf{0.00} & \textbf{0.00} & \textbf{0.00} & \textbf{0.00} & \textbf{0.00} & \textbf{0.00} & \textbf{0.00} & \textbf{0.00} & \textbf{0.00} & \textbf{0.00} & \textbf{0.00} & \textbf{0.00} & \textbf{0.00} & \textbf{0.00} & \textbf{0.00} & \textbf{0.00} & \textbf{0.00} & \textbf{0.000}
\\ \hline
\multirow{4}{*}{Cascade R-CNN \cite{cai2019cascade}} & APPA (on)        & 1.00          & 0.99          & 1.00          & 0.96          & 0.99          & 0.00          & 1.00          & 1.00          & 1.00          & 0.62          & 1.00          & 1.00          & 0.98          & 1.00          & 0.85          & 1.00          & 1.00          & 1.00          & 0.911          \\
                               & APPA   (outside) & 1.00          & 1.00          & 0.99          & 0.95          & 1.00          & 0.00          & 0.99          & 0.99          & 1.00          & 1.00          & 1.00          & 1.00          & 0.99          & 1.00          & 0.92          & 0.98          & 1.00          & 1.00          & 0.934          \\
                               & Thys \etal      & 1.00          & 1.00          & 1.00          & 0.89          & 1.00          & 0.00          & 0.99          & 1.00          & 1.00          & 0.85          & 1.00          & 1.00          & 0.99          & 1.00          & 0.00          & 0.99          & 1.00          & 1.00          & 0.873          \\
                               & Ours             & \textbf{0.00} & \textbf{0.00} & \textbf{0.00} & \textbf{0.00} & \textbf{0.00} & \textbf{0.00} & \textbf{0.00} & \textbf{0.00} & \textbf{0.00} & \textbf{0.00} & \textbf{0.00} & \textbf{0.00} & \textbf{0.00} & \textbf{0.00} & \textbf{0.00} & \textbf{0.00} & \textbf{0.00} & \textbf{0.00} & \textbf{0.000}
\\ \hline
\multirow{4}{*}{RetinaNet \cite{lin2017focal}} & APPA (on)        & 0.44          & 0.77          & 0.86          & 0.00          & 0.00          & 0.00          & 0.52          & 0.95          & 0.95          & 0.44          & 0.62          & 0.96          & 0.59          & 0.71          & 0.32          & 0.89          & 0.99          & 0.84          & 0.603          \\
                           & APPA   (outside) & 0.93          & 0.95          & 0.86          & 0.80          & 0.92          & 0.57          & 0.93          & 0.96          & 0.97          & 0.80          & 0.99          & 0.94          & 1.00          & 0.99          & 0.82          & 0.79          & 0.99          & 0.97          & 0.899          \\
                           & Thys \etal      & 0.40          & 0.86          & 0.97          & 0.91          & 0.00          & 0.00          & 0.87          & 0.92          & 0.91          & 0.77          & 0.99          & 0.99          & 0.95          & 0.98          & 0.93          & 0.95          & 0.89          & 0.84          & 0.785          \\
                           & Ours             & \textbf{0.00} & \textbf{0.00} & \textbf{0.00} & \textbf{0.00} & \textbf{0.00} & \textbf{0.00} & \textbf{0.00} & \textbf{0.00} & \textbf{0.00} & \textbf{0.00} & \textbf{0.00} & \textbf{0.00} & \textbf{0.00} & \textbf{0.00} & \textbf{0.00} & \textbf{0.00} & \textbf{0.00} & \textbf{0.00} & \textbf{0.000}
\\ \hline
\multirow{4}{*}{Mask R-CNN \cite{he2017mask}} & APPA (on)        & 0.73          & 0.94          & 0.21          & 0.90          & 0.71          & 0.43          & 0.94          & 0.98          & 0.99          & 0.90          & 0.82          & 0.98          & 0.70          & 0.96          & 0.95          & 0.99          & 0.99          & 0.98          & 0.839          \\
                            & APPA   (outside) & 0.73          & \textbf{0.51} & 0.61          & 0.73          & 0.86          & 0.00          & 0.53          & 0.27          & 0.91          & 0.92          & 0.98          & 0.94          & 0.96          & 0.97          & 0.93          & 0.86          & 0.67          & 0.99          & 0.743          \\
                            & Thys \etal      & 0.71          & 0.97          & 0.97          & 0.86          & 0.92          & 0.45          & 0.95          & 0.99          & 0.99          & 0.89          & 0.74          & 0.99          & 0.33          & 0.97          & 0.96          & 0.98          & 0.94          & 0.99          & 0.867          \\
                            & Ours             & \textbf{0.28} & 0.53 & \textbf{0.00} & \textbf{0.42} & \textbf{0.48} & \textbf{0.00} & \textbf{0.00} & \textbf{0.00} & \textbf{0.00} & \textbf{0.21} & \textbf{0.00} & \textbf{0.00} & \textbf{0.00} & \textbf{0.00} & \textbf{0.00} & \textbf{0.00} & \textbf{0.00} & \textbf{0.35} & \textbf{0.126}
\\ \hline
\multirow{4}{*}{FreeAnchor \cite{zhang2019freeanchor}} & APPA (on)        & 1.00          & 0.91          & 1.00          & 0.82          & 0.94          & 0.00          & 1.00          & 1.00          & 1.00          & 1.00          & 1.00          & 1.00          & 1.00          & 1.00          & 0.43          & 0.99          & 1.00          & 1.00          & 0.894          \\
                            & APPA   (outside) & 1.00          & 1.00          & 1.00          & 1.00          & 1.00          & 0.00          & 1.00          & 1.00          & 1.00          & 1.00          & 1.00          & 1.00          & 1.00          & 1.00          & 0.93          & 1.00          & 1.00          & 1.00          & 0.941          \\
                            & Thys \etal      & 0.22          & 0.89          & 1.00          & 0.47          & 0.91          & 0.00          & 1.00          & 1.00          & 1.00          & 0.97          & 1.00          & 1.00          & 0.98          & 1.00          & 0.49          & 0.99          & 0.99          & 0.97          & 0.827          \\
                            & Ours             & \textbf{0.00} & \textbf{0.00} & \textbf{0.00} & \textbf{0.00} & \textbf{0.00} & \textbf{0.00} & \textbf{0.00} & \textbf{0.00} & \textbf{0.00} & \textbf{0.00} & \textbf{0.00} & \textbf{0.00} & \textbf{0.00} & \textbf{0.00} & \textbf{0.00} & \textbf{0.00} & \textbf{0.00} & \textbf{0.00} & \textbf{0.000}
\\ \hline
\multirow{4}{*}{FSAF \cite{zhu2019feature}} & APPA (on)        & 0.35          & 0.51          & 0.77          & 0.21          & 0.31          & 0.00          & 0.76          & 0.86          & 0.76          & 0.49          & 0.51          & 0.78          & 0.68          & 0.60          & 0.67          & 0.84          & 0.61          & 0.88          & 0.588          \\
                      & APPA   (outside) & 0.86          & 0.81          & 0.77          & 0.61          & 0.87          & 0.38          & 0.49          & 0.69          & 0.78          & 0.86          & 0.83          & 0.84          & 0.82          & 0.77          & 0.30          & 0.76          & 0.78          & 0.80          & 0.723          \\
                      & Thys \etal      & 0.32          & 0.46          & 0.72          & 0.43          & 0.47          & 0.00          & 0.63          & 0.68          & 0.71          & 0.28          & 0.56          & 0.83          & 0.62          & 0.42          & 0.46          & 0.74          & 0.57          & 0.61          & 0.528          \\
                      & Ours             & \textbf{0.00} & \textbf{0.00} & \textbf{0.00} & \textbf{0.00} & \textbf{0.00} & \textbf{0.00} & \textbf{0.00} & \textbf{0.00} & \textbf{0.00} & \textbf{0.00} & \textbf{0.00} & \textbf{0.00} & \textbf{0.00} & \textbf{0.00} & \textbf{0.00} & \textbf{0.00} & \textbf{0.00} & \textbf{0.00} & \textbf{0.000}
\\ \hline
\multirow{4}{*}{RepPoints \cite{yang2019reppoints}} & APPA (on)        & 0.00          & 0.00          & 0.50          & 0.00          & 0.35          & 0.00          & 0.61          & 0.88          & 0.35          & 0.00          & 0.38          & 0.90          & 0.38        & 0.34          & 0.46          & 0.63          & 0.00          & 0.45          & 0.346          \\
                           & APPA   (outside) & 0.56          & 0.81          & 0.73          & 0.57          & 0.53          & 0.00          & 0.54          & 0.71          & 0.78          & 0.75          & 0.89          & 0.84          & 0.71          & 0.63          & 0.73          & 0.85          & 0.72          & 0.89          & 0.680          \\
                           & Thys \etal      & 0.00          & 0.23          & 0.57          & 0.23          & 0.22          & 0.00          & 0.23          & 0.87          & 0.49          & 0.24          & 0.50          & 0.89          & 0.43          & 0.46          & 0.21          & 0.77          & 0.52          & 0.52          & 0.410          \\
                           & Ours             & \textbf{0.00} & \textbf{0.00} & \textbf{0.00} & \textbf{0.00} & \textbf{0.00} & \textbf{0.00} & \textbf{0.00} & \textbf{0.00} & \textbf{0.00} & \textbf{0.00} & \textbf{0.00} & \textbf{0.00} & \textbf{0.00} & \textbf{0.00} & \textbf{0.00} & \textbf{0.00} & \textbf{0.00} & \textbf{0.00} & \textbf{0.000}
\\ \hline
\multirow{4}{*}{TOOD \cite{feng2021tood}} & APPA (on)        & 0.00          & 0.00          & 0.00          & 0.00          & 0.30          & 0.00          & 0.36          & 0.44          & 0.00          & 0.00          & 0.00          & 0.85          & 0.44          & 0.46          & 0.00          & 0.00          & 0.00          & 0.38          & 0.179          \\
                      & APPA   (outside) & 0.71          & 0.78          & 0.84          & 0.70          & 0.85          & 0.28          & 0.51          & 0.82          & 0.63          & 0.42          & 0.57          & 0.87          & 0.59          & 0.72          & 0.87          & 0.81          & 0.83          & 0.87          & 0.704          \\
                      & Thys \etal      & 0.00          & 0.20          & 0.22          & 0.00          & 0.00          & 0.00          & 0.00          & 0.30          & 0.43          & 0.00          & 0.00          & 0.69          & 0.38          & 0.24          & 0.59          & 0.21          & 0.20          & 0.40          & 0.214          \\
                      & Ours             & \textbf{0.00} & \textbf{0.00} & \textbf{0.00} & \textbf{0.00} & \textbf{0.00} & \textbf{0.00} & \textbf{0.00} & \textbf{0.00} & \textbf{0.00} & \textbf{0.00} & \textbf{0.00} & \textbf{0.00} & \textbf{0.00} & \textbf{0.00} & \textbf{0.00} & \textbf{0.00} & \textbf{0.00} & \textbf{0.00} & \textbf{0.000}
\\ \hline
\multirow{4}{*}{ATSS \cite{zhang2020bridging}} & APPA (on)        & 0.29          & 0.57          & 0.60          & 0.38          & 0.59          & 0.00          & 0.61          & 0.70          & 0.66          & 0.00          & 0.22          & 0.67          & 0.37          & 0.43          & 0.49          & 0.60          & 0.32          & 0.59          & 0.449          \\
                      & APPA   (outside) & 0.62          & 0.56          & 0.73          & 0.53          & 0.56          & 0.00          & 0.72          & 0.75          & 0.76          & 0.47          & 0.53          & 0.74          & 0.51          & 0.57          & 0.64          & 0.75          & 0.68          & 0.74          & 0.603          \\
                      & Thys \etal      & 0.45          & 0.67          & 0.74          & 0.54          & 0.56          & 0.00          & 0.64          & 0.68          & 0.73          & 0.34          & 0.40          & 0.63          & \textbf{0.24} & 0.57          & 0.35          & 0.55          & 0.49          & 0.51          & 0.505          \\
                      & Ours             & \textbf{0.00} & \textbf{0.26} & \textbf{0.00} & \textbf{0.00} & \textbf{0.28} & \textbf{0.00} & \textbf{0.00} & \textbf{0.00} & \textbf{0.00} & \textbf{0.00} & \textbf{0.00} & \textbf{0.00} & 0.29          & \textbf{0.00} & \textbf{0.00} & \textbf{0.00} & \textbf{0.22} & \textbf{0.28} & \textbf{0.074}
\\ \hline
\multirow{4}{*}{FoveaBox \cite{kong2020foveabox}} & APPA (on)        & 0.86          & 0.95          & 0.83          & 0.89          & 0.84          & 0.00          & 0.78          & 0.82          & 0.73          & 0.56          & 0.93          & 0.86          & 0.76          & 0.94          & 0.89          & 0.81          & 0.58          & 0.91          & 0.774          \\
                          & APPA   (outside) & 0.82          & 0.95          & 0.78          & 0.81          & 0.90          & 0.00          & 0.49          & 0.85          & 0.83          & 0.85          & 0.89          & 0.80          & 0.88          & 0.91          & 0.82          & 0.86          & 0.73          & 0.89          & 0.781          \\
                          & Thys \etal      & 0.70          & 0.92          & 0.92          & 0.74          & 0.87          & 0.00          & 0.81          & 0.88          & 0.85          & 0.00          & 0.47          & 0.94          & 0.00          & 0.71          & 0.53          & 0.82          & 0.93          & 0.87          & 0.664          \\
                          & Ours             & \textbf{0.00} & \textbf{0.00} & \textbf{0.00} & \textbf{0.00} & \textbf{0.00} & \textbf{0.00} & \textbf{0.00} & \textbf{0.00} & \textbf{0.00} & \textbf{0.00} & \textbf{0.00} & \textbf{0.00} & \textbf{0.00} & \textbf{0.00} & \textbf{0.00} & \textbf{0.00} & \textbf{0.00} & \textbf{0.00} & \textbf{0.000}
\\ \hline
\multirow{4}{*}{VarifocalNet \cite{zhang2021varifocalnet}} & APPA (on)        & \textbf{0.00} & 0.76          & 0.89          & 0.43          & 0.78          & 0.44          & 0.89          & 0.92          & 0.94          & \textbf{0.22} & 0.61          & 0.92          & 0.59          & 0.81          & 0.86          & 0.90          & 0.74          & 0.92          & 0.701          \\
                              & APPA   (outside) & 0.79          & 0.87          & 0.91          & 0.82          & 0.84          & 0.77          & 0.93          & 0.88          & 0.92          & 0.67          & 0.78          & 0.89          & 0.83          & 0.90          & 0.83          & 0.88          & 0.85          & 0.88          & 0.847          \\
                              & Thys \etal      & 0.23          & \textbf{0.58} & 0.89          & \textbf{0.22} & \textbf{0.25} & 0.00          & 0.23          & 0.90          & 0.83          & 0.30          & \textbf{0.00} & 0.94          & \textbf{0.33} & \textbf{0.53} & 0.63          & 0.73          & 0.49          & 0.87          & 0.497          \\
                              & Ours             & 0.66          & 0.87          & \textbf{0.59} & 0.46          & 0.41          & \textbf{0.00} & \textbf{0.00} & \textbf{0.24} & \textbf{0.42} & 0.62          & 0.71          & \textbf{0.42} & 0.52          & 0.67          & \textbf{0.00} & \textbf{0.42} & \textbf{0.46} & \textbf{0.41} & \textbf{0.438}         
\\                 
\hline\hline
\end{tabular*}
    \begin{tablenotes}
        \footnotesize      
        \item Strongest attack results are highlighted in \textbf{bold}.
        \item "on" and "outside" represent patches on and outside targets, respectively.
    \end{tablenotes}
\end{threeparttable}
\end{table*}

\begin{table*}[t!]
\caption{\\\small{DETAILED RESULTS OF BLACK-BOX ATTACKS.}}
\label{table_black_box_attack_results_comparison}
\centering
\setlength{\tabcolsep}{0.92mm}
\begin{threeparttable}
\begin{tabular*}{\hsize}{ccccccccccccccccccccc}
\hline\hline
\diaghead{\theadfont Diag ColumnmnHead}{Patches}{Targets} &--- &T1 &T2 &T3 &T4 &T5 &T6 &T7 &T8 &T9 &T10 &T11 &T12 &T13 &T14 &T15 &T16 &T17 &T18 &Average 
\\ \hline\hline
\multirow{4}{*}{SSD\cite{liu2016ssd}} & APPA (on)        & 0.00          & 0.00          & 0.90          & 0.99          & 0.82          & 0.43          & 0.79          & 0.98          & 0.93          & 0.83          & 0.99          & 0.98          & 0.76          & 0.99          & 0.50          & 0.96          & 0.77          & 0.95          & 0.754          \\
                     & APPA   (outside) & 0.91          & 0.99          & 0.63          & 0.94          & 0.99          & 0.48          & 0.94          & 0.91          & 0.95          & 0.97          & 1.00          & 0.97          & 0.94          & 1.00          & 0.94          & 0.95          & 1.00          & 1.00          & 0.917          \\
                     & Thys \etal      & 0.00          & 0.00          & 0.34          & \textbf{0.00} & 0.00          & \textbf{0.00} & 0.96          & 0.93          & 0.76          & 0.00          & 0.67          & 0.98          & 0.44          & 0.98          & 0.86          & 0.94          & 0.53          & 0.95          & 0.519          \\
                     & Ours             & \textbf{0.00} & \textbf{0.00} & \textbf{0.00} & 0.37          & \textbf{0.00} & 0.31          & \textbf{0.00} & \textbf{0.00} & \textbf{0.00} & \textbf{0.00} & \textbf{0.00} & \textbf{0.00} & \textbf{0.00} & \textbf{0.80} & \textbf{0.00} & \textbf{0.00} & \textbf{0.00} & \textbf{0.00} & \textbf{0.082}
 \\ \hline
 \multirow{4}{*}{Faster R-CNN\cite{ren2015faster}} & APPA (on)        & 0.00          & 0.77          & 0.00          & \textbf{0.00} & 0.82          & 0.00          & 0.99          & 1.00          & 0.98          & \textbf{0.44} & 0.92          & 1.00          & 0.94          & \textbf{0.25} & 0.33          & 0.99          & 0.54          & 0.99          & 0.609          \\
                              & APPA   (outside) & 1.00          & 1.00          & 1.00          & 1.00          & 1.00          & 0.91          & 0.99          & 1.00          & 1.00          & 1.00          & 1.00          & 1.00          & 1.00          & 0.99          & 0.98          & 1.00          & 1.00          & 1.00          & 0.993          \\
                              & Thys \etal      & 0.99          & 0.94          & 0.99          & 0.89          & 0.78          & 0.00          & 1.00          & 1.00          & 1.00          & 0.95          & 0.99          & 1.00          & 0.96          & 0.63          & 0.98          & 1.00          & 0.99          & 0.99          & 0.893          \\
                              & Ours             & \textbf{0.00} & \textbf{0.24} & \textbf{0.00} & 0.51          & \textbf{0.27} & \textbf{0.00} & \textbf{0.00} & \textbf{0.00} & \textbf{0.00} & 0.80          & \textbf{0.00} & \textbf{0.00} & \textbf{0.00} & 0.49          & \textbf{0.00} & \textbf{0.00} & \textbf{0.00} & \textbf{0.00} & \textbf{0.128}
\\ \hline
\multirow{4}{*}{Swin Transformer\cite{liu2021swin}} & APPA (on)        & 0.98          & 1.00          & 1.00          & 1.00          & 1.00          & 0.95          & 1.00          & 1.00          & 1.00          & 0.95          & 0.99          & 1.00          & 1.00          & 1.00          & 1.00          & 1.00          & 1.00          & 1.00          & 0.993          \\
                                  & APPA   (outside) & 1.00          & 1.00          & 1.00          & 1.00          & 1.00          & 0.56          & 1.00          & 1.00          & 1.00          & 1.00          & 1.00          & 1.00          & 1.00          & 1.00          & 0.99          & 1.00          & 1.00          & 1.00          & 0.975          \\
                                  & Thys \etal      & 1.00          & 1.00          & 1.00          & 0.99          & 1.00          & 0.42          & 1.00          & 1.00          & 1.00          & 1.00          & 1.00          & 1.00          & 1.00          & 1.00          & 0.99          & 1.00          & 1.00          & 1.00          & 0.967          \\
                                  & Ours             & \textbf{0.00} & \textbf{0.33} & \textbf{0.00} & \textbf{0.00} & \textbf{0.00} & \textbf{0.00} & \textbf{0.00} & \textbf{0.00} & \textbf{0.00} & \textbf{0.00} & \textbf{0.00} & \textbf{0.00} & \textbf{0.00} & \textbf{0.20} & \textbf{0.00} & \textbf{0.00} & \textbf{0.00} & \textbf{0.00} & \textbf{0.029}
\\ \hline
\multirow{4}{*}{YOLOv2\cite{redmon2017yolo9000}} & APPA (on)        & 0.00          & 0.00          & 0.00          & 0.00          & 0.00          & 0.00          & 0.00          & 0.00          & 0.00          & 0.00          & 0.00          & 0.73          & 0.00          & 0.99          & 0.00          & 0.00          & 0.00          & 0.00          & 0.096          \\
                        & APPA   (outside) & 1.00          & 1.00          & 1.00          & 1.00          & 1.00          & 0.00          & 0.00          & 1.00          & 1.00          & 0.99          & 1.00          & 0.99          & 0.98          & 1.00          & 0.82          & 0.98          & 0.88          & 0.99          & 0.868          \\
                        & Thys \etal      & 0.00          & 0.00          & 1.00          & 0.00          & 0.00          & 0.00          & 0.99          & 0.99          & 1.00          & 0.00          & 0.00          & 0.97          & 0.00          & 0.98          & 0.00          & 0.97          & 0.00          & 0.75          & 0.425          \\
                        & Ours             & \textbf{0.00} & \textbf{0.00} & \textbf{0.00} & \textbf{0.00} & \textbf{0.00} & \textbf{0.00} & \textbf{0.00} & \textbf{0.00} & \textbf{0.00} & \textbf{0.00} & \textbf{0.00} & \textbf{0.00} & \textbf{0.00} & \textbf{0.00} & \textbf{0.00} & \textbf{0.00} & \textbf{0.00} & \textbf{0.00} & \textbf{0.000}
\\ \hline
\multirow{4}{*}{YOLOv3\cite{redmon2018yolov3}} & APPA (on)        & 0.88          & 0.87          & 0.72          & 0.88          & 0.90          & 0.00          & 0.87          & 0.88          & 0.89          & 0.74          & 0.87          & 0.86          & 0.84          & 0.91          & 0.80          & 0.90          & 0.91          & 0.91          & 0.813          \\
                        & APPA   (outside) & 0.87          & 0.89          & 0.83          & 0.82          & 0.90          & 0.71          & 0.84          & 0.87          & 0.88          & 0.86          & 0.91          & 0.91          & 0.88          & 0.91          & 0.88          & 0.89          & 0.89          & 0.90          & 0.869          \\
                        & Thys \etal      & 0.72          & \textbf{0.50}          & 0.78          & 0.85          & 0.90          & 0.00          & 0.86          & 0.89          & 0.88          & 0.82          & 0.66          & 0.89          & 0.56          & 0.89          & 0.82          & 0.90          & 0.91          & 0.92          & 0.764          \\
                        & Ours             & \textbf{0.00} & 0.67 & \textbf{0.00} & \textbf{0.74} & \textbf{0.87} & \textbf{0.00} & \textbf{0.00} & \textbf{0.00} & \textbf{0.00} & \textbf{0.00} & \textbf{0.00} & \textbf{0.00} & \textbf{0.48} & \textbf{0.88} & \textbf{0.00} & \textbf{0.00} & \textbf{0.00} & \textbf{0.00} & \textbf{0.202}
\\ \hline
\multirow{4}{*}{YOLOv5l\cite{jocher2020yolov5}} & APPA (on)        & 0.90          & 0.91          & 0.89          & 0.83          & 0.91          & 0.00          & 0.89          & 0.90          & 0.91          & 0.67          & 0.90          & 0.90          & 0.89          & 0.92          & 0.82          & 0.89          & 0.91          & 0.90          & 0.830          \\
                         & APPA   (outside) & 0.90          & 0.92          & 0.90          & 0.89          & 0.89          & 0.75          & 0.87          & 0.88          & 0.89          & 0.82          & 0.93          & 0.91          & 0.93          & 0.92          & 0.84          & 0.89          & 0.91          & 0.89          & 0.885          \\
                         & Thys \etal      & 0.85          & 0.89          & 0.89          & 0.88          & 0.89          & 0.22          & 0.90          & 0.92          & 0.89          & 0.87          & 0.90          & 0.90          & 0.90          & 0.91          & 0.88          & 0.90          & 0.91          & 0.90          & 0.856          \\
                         & Ours             & \textbf{0.00} & \textbf{0.89} & \textbf{0.00} & \textbf{0.52} & \textbf{0.82} & \textbf{0.00} & \textbf{0.00} & \textbf{0.00} & \textbf{0.00} & \textbf{0.00} & \textbf{0.00} & \textbf{0.00} & \textbf{0.00} & \textbf{0.88} & \textbf{0.00} & \textbf{0.00} & \textbf{0.45} & \textbf{0.00} & \textbf{0.198}
\\ \hline
\multirow{4}{*}{Cascade R-CNN \cite{cai2019cascade}} & APPA (on)        & 0.98          & 1.00          & 1.00          & 0.99          & 1.00          & 0.97          & 1.00          & 1.00          & 1.00          & 0.00          & 1.00          & 1.00          & 1.00          & 1.00          & 0.00          & 0.99          & 0.99          & 1.00          & 0.884          \\
                               & APPA   (outside) & 1.00          & 1.00          & 0.99          & 0.95          & 1.00          & 0.99          & 1.00          & 1.00          & 0.99          & 1.00          & 1.00          & 1.00          & 1.00          & 1.00          & 0.36          & 1.00          & 1.00          & 1.00          & 0.960          \\
                               & Thys \etal      & 0.98          & 1.00          & 1.00          & 1.00          & 1.00          & 0.00          & 1.00          & 1.00          & 1.00          & 1.00          & 1.00          & 1.00          & 1.00          & 1.00          & 0.71          & 1.00          & 1.00          & 1.00          & 0.927          \\
                               & Ours             & \textbf{0.00} & \textbf{0.00} & \textbf{0.00} & \textbf{0.00} & \textbf{0.00} & \textbf{0.00} & \textbf{0.00} & \textbf{0.00} & \textbf{0.00} & \textbf{0.00} & \textbf{0.00} & \textbf{0.00} & \textbf{0.00} & \textbf{0.00} & \textbf{0.00} & \textbf{0.00} & \textbf{0.00} & \textbf{0.00} & \textbf{0.000}
\\ \hline
\multirow{4}{*}{RetinaNet \cite{lin2017focal}} & APPA (on)        & 0.00          & 0.83          & 0.97          & 0.95          & 0.98          & 0.49          & 0.93          & 0.93          & 0.97          & 0.00          & 0.93          & 0.99          & 0.85          & 0.98          & 0.80          & 0.94          & 0.98          & 0.95          & 0.804          \\
                           & APPA   (outside) & 0.98          & 1.00          & 0.91          & 0.98          & 0.97          & 0.85          & 0.97          & 0.97          & 0.98          & 0.99          & 1.00          & 0.96          & 0.96          & 0.99          & 0.93          & 0.93          & 0.98          & 0.98          & 0.963          \\
                           & Thys \etal      & 0.28          & 0.89          & 0.96          & 0.77          & 0.59          & 0.00          & 0.96          & 0.97          & 0.96          & 0.95          & 0.83          & 0.99          & 0.78          & 0.98          & 0.90          & 0.98          & 0.95          & 0.99          & 0.818          \\
                           & Ours             & \textbf{0.00} & \textbf{0.00} & \textbf{0.00} & \textbf{0.00} & \textbf{0.00} & \textbf{0.00} & \textbf{0.00} & \textbf{0.00} & \textbf{0.00} & \textbf{0.00} & \textbf{0.00} & \textbf{0.00} & \textbf{0.00} & \textbf{0.00} & \textbf{0.00} & \textbf{0.00} & \textbf{0.00} & \textbf{0.00} & \textbf{0.000}
\\ \hline
\multirow{4}{*}{Mask R-CNN \cite{he2017mask}} & APPA (on)        & 0.55          & 0.80          & 0.23          & 0.79          & 0.40          & 0.00          & 0.86          & 0.97          & 0.92          & \textbf{0.39} & 0.37          & 0.99          & 0.57          & \textbf{0.50} & 0.57          & 0.97          & 0.00          & 0.97          & 0.603          \\
                            & APPA   (outside) & \textbf{0.48} & 0.99          & 0.84          & 0.97          & 0.98          & 0.76          & 0.74          & 0.87          & 0.93          & 0.99          & 0.98          & 0.54          & 0.98          & 0.98          & 0.70          & 0.98          & 0.69          & 0.99          & 0.855          \\
                            & Thys \etal      & 0.98          & 0.98          & 0.94          & 0.91          & 0.56          & 0.00          & 0.96          & 0.98          & 0.96          & 0.97          & 0.97          & 0.99          & 0.90          & 0.94          & 0.92          & 0.98          & 0.85          & 0.98          & 0.876          \\
                            & Ours             & 0.55          & \textbf{0.44} & \textbf{0.00} & \textbf{0.72} & \textbf{0.40} & \textbf{0.00} & \textbf{0.00} & \textbf{0.00} & \textbf{0.00} & 0.85          & \textbf{0.00} & \textbf{0.00} & \textbf{0.00} & 0.78          & \textbf{0.00} & \textbf{0.00} & \textbf{0.00} & \textbf{0.00} & \textbf{0.208}
\\ \hline
\multirow{4}{*}{FreeAnchor \cite{zhang2019freeanchor}} & APPA (on)        & 0.98          & 1.00          & 1.00          & 1.00          & 1.00          & 0.00          & 1.00          & 1.00          & 1.00          & 0.85          & 1.00          & 1.00          & 0.96          & 1.00          & 0.88          & 1.00          & 0.92          & 0.99          & 0.921          \\
                            & APPA   (outside) & 1.00          & 1.00          & 1.00          & 1.00          & 1.00          & 0.87          & 1.00          & 1.00          & 1.00          & 1.00          & 1.00          & 1.00          & 1.00          & 1.00          & 0.94          & 1.00          & 1.00          & 1.00          & 0.989          \\
                            & Thys \etal      & 1.00          & 0.99          & 1.00          & 0.94          & 0.73          & 0.00          & 1.00          & 1.00          & 1.00          & 0.99          & 1.00          & 1.00          & 0.85          & 0.99          & 0.99          & 1.00          & 0.98          & 1.00          & 0.914          \\
                            & Ours             & \textbf{0.00} & \textbf{0.00} & \textbf{0.00} & \textbf{0.40} & \textbf{0.21} & \textbf{0.00} & \textbf{0.00} & \textbf{0.00} & \textbf{0.00} & \textbf{0.00} & \textbf{0.00} & \textbf{0.00} & \textbf{0.00} & \textbf{0.00} & \textbf{0.00} & \textbf{0.00} & \textbf{0.00} & \textbf{0.00} & \textbf{0.034}
\\ \hline
\multirow{4}{*}{FSAF \cite{zhu2019feature}} & APPA (on)        & 0.00          & 0.56          & 0.61          & 0.00          & 0.43          & 0.00          & 0.76          & 0.84          & 0.69          & 0.00          & 0.40          & 0.81          & 0.60          & 0.32          & 0.48          & 0.74          & 0.65          & 0.76          & 0.481          \\
                      & APPA   (outside) & 0.86          & 0.85          & 0.67          & 0.76          & 0.85          & 0.80          & 0.82          & 0.85          & 0.73          & 0.86          & 0.92          & 0.86          & 0.86          & 0.74          & 0.62          & 0.88          & 0.88          & 0.86          & 0.815          \\
                      & Thys \etal      & 0.82          & 0.70          & 0.72          & 0.46          & 0.43          & 0.00          & 0.67          & 0.83          & 0.71          & 0.63          & 0.69          & 0.93          & 0.75          & 0.51          & 0.64          & 0.82          & 0.64          & 0.80          & 0.653          \\
                      & Ours             & \textbf{0.00} & \textbf{0.00} & \textbf{0.00} & \textbf{0.00} & \textbf{0.00} & \textbf{0.00} & \textbf{0.00} & \textbf{0.00} & \textbf{0.00} & \textbf{0.00} & \textbf{0.00} & \textbf{0.00} & \textbf{0.00} & \textbf{0.00} & \textbf{0.00} & \textbf{0.00} & \textbf{0.00} & \textbf{0.00} & \textbf{0.000}
\\ \hline
\multirow{4}{*}{RepPoints \cite{yang2019reppoints}} & APPA (on)        & 0.00          & 0.00          & 0.64          & 0.00          & 0.35          & 0.00          & 0.79          & 0.90          & 0.58          & 0.00          & 0.27          & 0.87          & 0.31          & 0.00          & 0.32          & 0.73          & 0.26          & 0.80          & 0.379          \\
                           & APPA   (outside) & 0.87          & 0.90          & 0.84          & 0.86          & 0.87          & 0.61          & 0.85          & 0.86          & 0.79          & 0.87          & 0.90          & 0.83          & 0.89          & 0.89          & 0.44          & 0.89          & 0.24          & 0.83          & 0.791          \\
                           & Thys \etal      & 0.00          & 0.38          & 0.67          & 0.31          & 0.68          & 0.00          & 0.77          & 0.86          & 0.75          & 0.45          & 0.75          & 0.94          & 0.50          & 0.43          & 0.65          & 0.86          & 0.26          & 0.71          & 0.554          \\
                           & Ours             & \textbf{0.00} & \textbf{0.00} & \textbf{0.00} & \textbf{0.00} & \textbf{0.00} & \textbf{0.00} & \textbf{0.00} & \textbf{0.00} & \textbf{0.00} & \textbf{0.00} & \textbf{0.00} & \textbf{0.00} & \textbf{0.00} & \textbf{0.00} & \textbf{0.00} & \textbf{0.00} & \textbf{0.00} & \textbf{0.00} & \textbf{0.000}
\\ \hline
\multirow{4}{*}{TOOD \cite{feng2021tood}} & APPA (on)        & 0.00          & 0.39          & 0.24          & 0.42          & 0.38          & 0.00          & 0.60          & 0.50          & 0.41          & \textbf{0.00} & 0.49          & 0.79          & 0.44          & 0.36          & 0.35          & 0.36          & 0.27          & 0.45          & 0.358          \\
                      & APPA   (outside) & 0.74          & 0.88          & 0.78          & 0.73          & 0.70          & 0.70          & 0.77          & 0.79          & 0.81          & 0.78          & 0.68          & 0.68          & 0.44          & 0.67          & 0.76          & 0.79          & 0.50          & 0.76          & 0.720          \\
                      & Thys \etal      & 0.57          & 0.69          & 0.73          & 0.66          & 0.56          & 0.00          & 0.68          & 0.60          & 0.39          & 0.34          & 0.53          & 0.81          & 0.36          & 0.41          & 0.63          & 0.79          & 0.40          & 0.47          & 0.534          \\
                      & Ours             & \textbf{0.00} & \textbf{0.25} & \textbf{0.00} & \textbf{0.30} & \textbf{0.00} & \textbf{0.00} & \textbf{0.00} & \textbf{0.00} & \textbf{0.00} & 0.34          & \textbf{0.00} & \textbf{0.00} & \textbf{0.22} & \textbf{0.37} & \textbf{0.00} & \textbf{0.00} & \textbf{0.00} & \textbf{0.00} & \textbf{0.082}
\\ \hline
\multirow{4}{*}{ATSS \cite{zhang2020bridging}} & APPA (on)        & 0.35          & 0.67          & 0.63          & 0.37          & 0.42          & 0.00          & 0.61          & 0.71          & 0.62          & 0.00          & 0.25          & 0.69          & 0.39          & 0.64          & 0.29          & 0.56          & 0.41          & 0.51          & 0.451          \\
                      & APPA   (outside) & 0.61          & 0.76          & 0.72          & 0.58          & 0.57          & 0.71          & 0.70          & 0.75          & 0.70          & 0.58          & 0.54          & 0.72          & 0.53          & 0.66          & 0.61          & 0.73          & 0.56          & 0.62          & 0.647          \\
                      & Thys \etal      & 0.41          & 0.60          & 0.67          & 0.54          & 0.56          & 0.00          & 0.63          & 0.67          & 0.65          & 0.49          & 0.42          & 0.76          & 0.44          & 0.62          & 0.59          & 0.67          & 0.50          & 0.61          & 0.546          \\
                      & Ours             & \textbf{0.00} & \textbf{0.00} & \textbf{0.42} & \textbf{0.00} & \textbf{0.00} & \textbf{0.00} & \textbf{0.30} & \textbf{0.42} & \textbf{0.00} & \textbf{0.00} & \textbf{0.00} & \textbf{0.46} & \textbf{0.36} & \textbf{0.00} & \textbf{0.26} & \textbf{0.43} & \textbf{0.00} & \textbf{0.00} & \textbf{0.147}
\\ \hline
\multirow{4}{*}{FoveaBox \cite{kong2020foveabox}} & APPA (on)        & 0.38          & 0.93          & 0.84          & 0.54          & 0.90          & 0.00          & 0.89          & 0.89          & 0.75          & 0.00          & 0.80          & 0.92          & 0.28          & 0.75          & 0.48          & 0.91          & 0.83          & 0.95          & 0.669          \\
                          & APPA   (outside) & 0.93          & 0.94          & 0.81          & 0.95          & 0.92          & 0.86          & 0.83          & 0.83          & 0.84          & 0.94          & 0.95          & 0.82          & 0.93          & 0.93          & 0.72          & 0.84          & 0.92          & 0.90          & 0.881          \\
                          & Thys \etal      & 0.92          & 0.92          & 0.91          & 0.93          & 0.95          & 0.00          & 0.89          & 0.88          & 0.81          & 0.82          & 0.81          & 0.93          & 0.83          & 0.85          & 0.53          & 0.88          & 0.87          & 0.89          & 0.812          \\
                          & Ours             & \textbf{0.00} & \textbf{0.00} & \textbf{0.00} & \textbf{0.00} & \textbf{0.00} & \textbf{0.00} & \textbf{0.00} & \textbf{0.00} & \textbf{0.00} & \textbf{0.00} & \textbf{0.00} & \textbf{0.00} & \textbf{0.00} & \textbf{0.00} & \textbf{0.00} & \textbf{0.00} & \textbf{0.00} & \textbf{0.00} & \textbf{0.000}
\\ \hline
\multirow{4}{*}{VarifocalNet \cite{zhang2021varifocalnet}} & APPA (on)        & 0.55          & 0.84          & 0.83          & 0.65          & 0.73          & 0.24          & 0.89          & 0.90          & 0.88          & 0.30          & 0.34          & 0.90          & 0.81          & 0.68          & 0.58          & 0.82          & 0.56          & 0.90          & 0.689          \\
                              & APPA   (outside) & 0.88          & 0.93          & 0.89          & 0.86          & 0.85          & 0.90          & 0.91          & 0.91          & 0.91          & 0.89          & 0.88          & 0.93          & 0.88          & 0.87          & 0.87          & 0.90          & 0.77          & 0.85          & 0.882          \\
                              & Thys \etal      & 0.76          & 0.85          & 0.89          & 0.83          & 0.88          & 0.00          & 0.90          & 0.92          & 0.91          & 0.79          & 0.73          & 0.94          & 0.72          & 0.66          & 0.91          & 0.92          & 0.81          & 0.87          & 0.794          \\
                              & Ours             & \textbf{0.00} & \textbf{0.00} & \textbf{0.00} & \textbf{0.00} & \textbf{0.00} & \textbf{0.00} & \textbf{0.00} & \textbf{0.00} & \textbf{0.00} & \textbf{0.00} & \textbf{0.00} & \textbf{0.00} & \textbf{0.00} & \textbf{0.00} & \textbf{0.00} & \textbf{0.00} & \textbf{0.00} & \textbf{0.00} & \textbf{0.000}      
\\                 
\hline\hline
\end{tabular*}
    \begin{tablenotes}
        \footnotesize      
        \item Strongest attack results are highlighted in \textbf{bold}.
        \item "on" and "outside" represent patches on and outside targets, respectively.
        \item The proxy model is YOLOv5n in the black-box setting.
    \end{tablenotes}
\end{threeparttable}
\end{table*}

\begin{figure*}
  \centering
  \begin{subfigure}{0.24\linewidth}
    \includegraphics[width=1\linewidth]{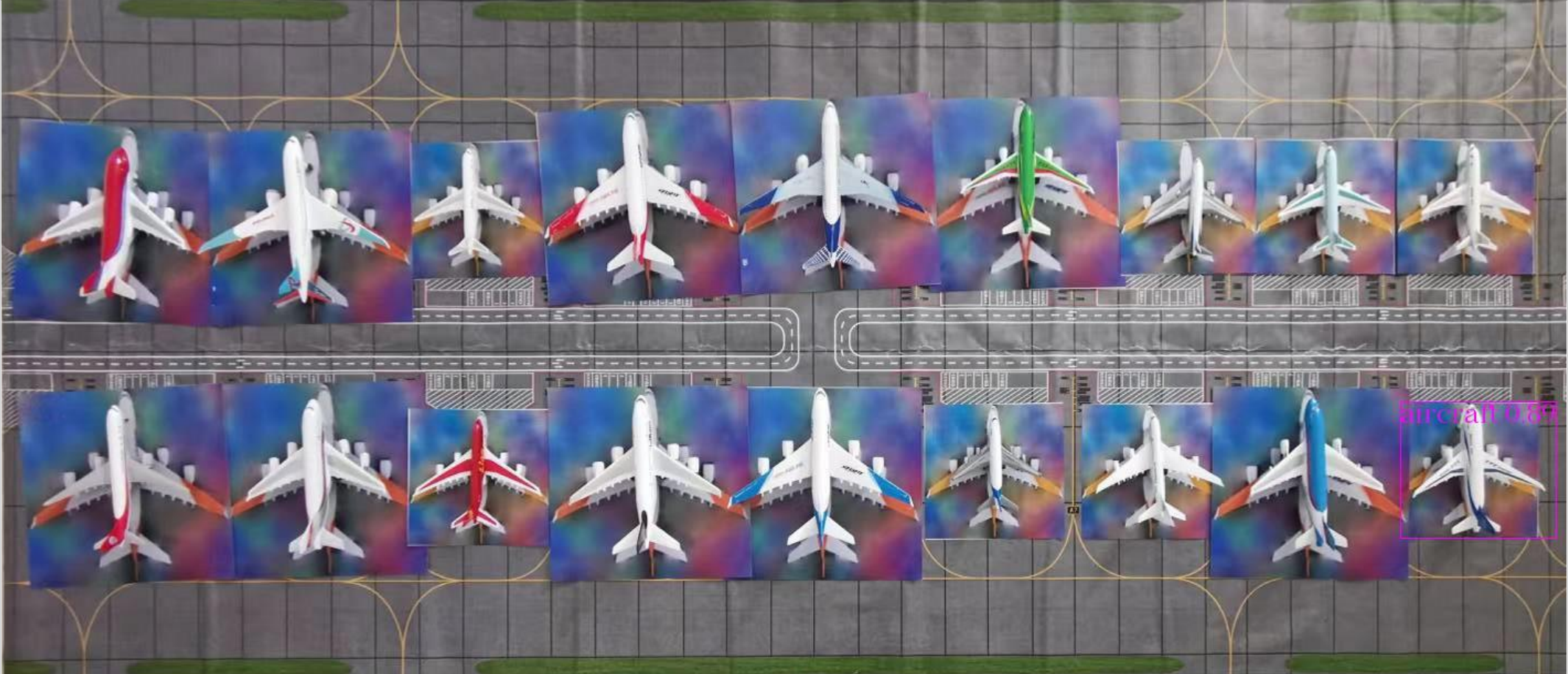}
    \caption{YOLOv2}
  \end{subfigure}
  \begin{subfigure}{0.24\linewidth}
    \includegraphics[width=1\linewidth]{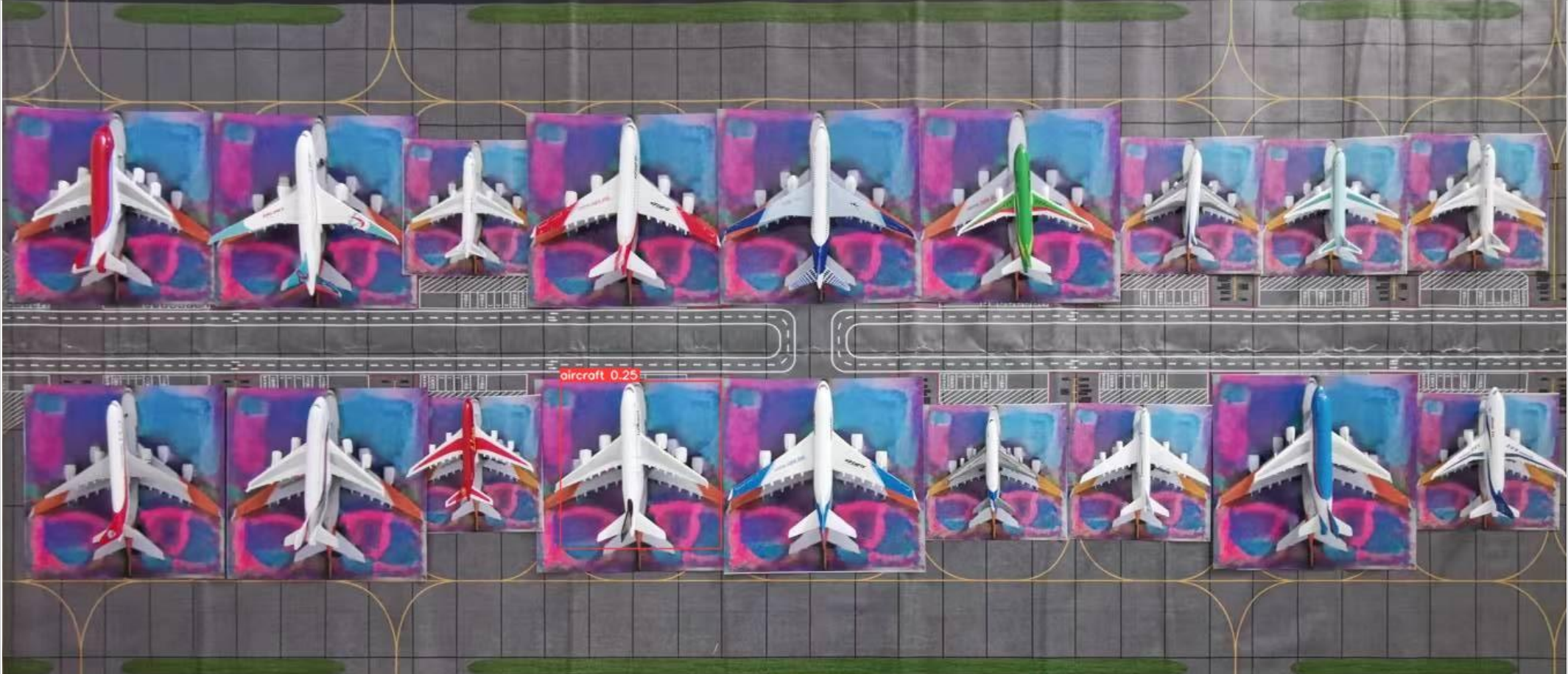}
    \caption{YOLOv3}
  \end{subfigure}
  \begin{subfigure}{0.24\linewidth}
    \includegraphics[width=1\linewidth]{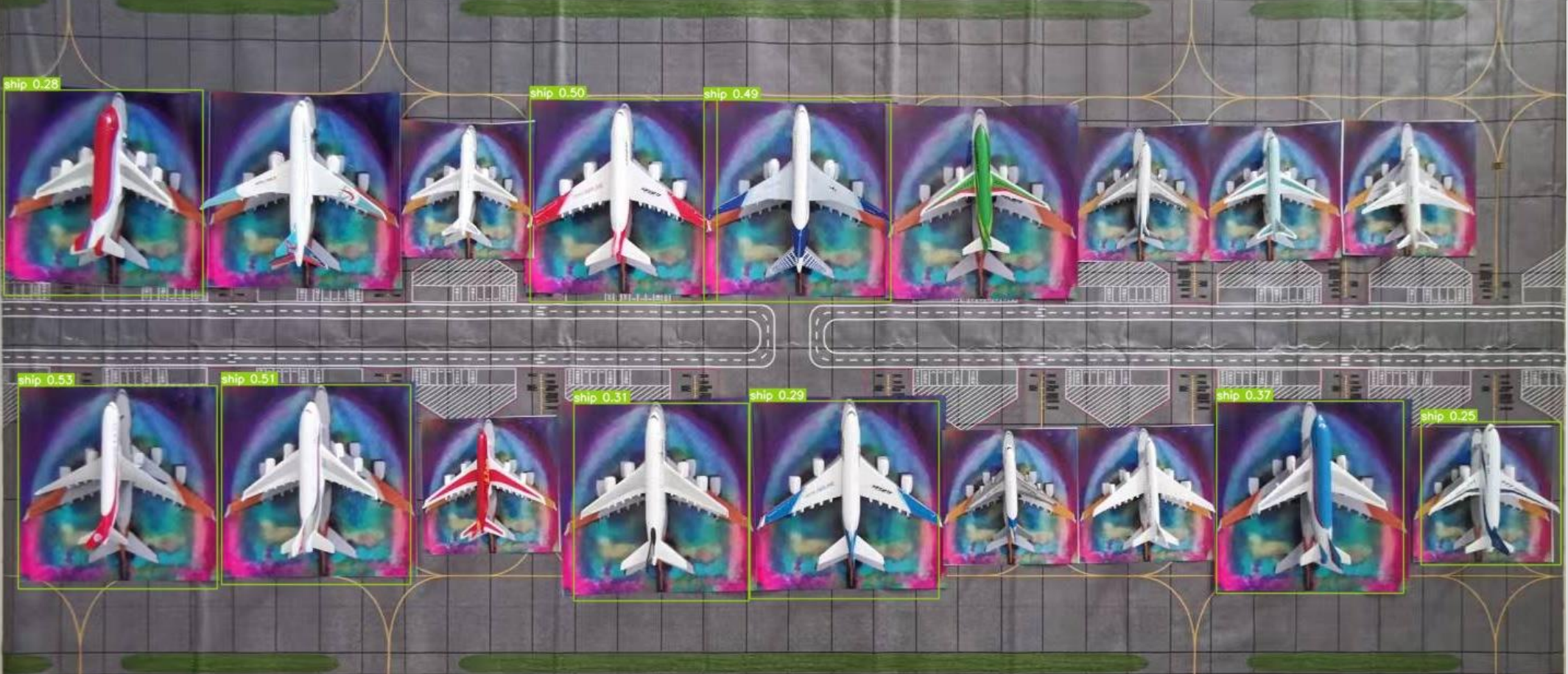}
    \caption{YOLOv5n}
  \end{subfigure}
  \begin{subfigure}{0.24\linewidth}
    \includegraphics[width=1\linewidth]{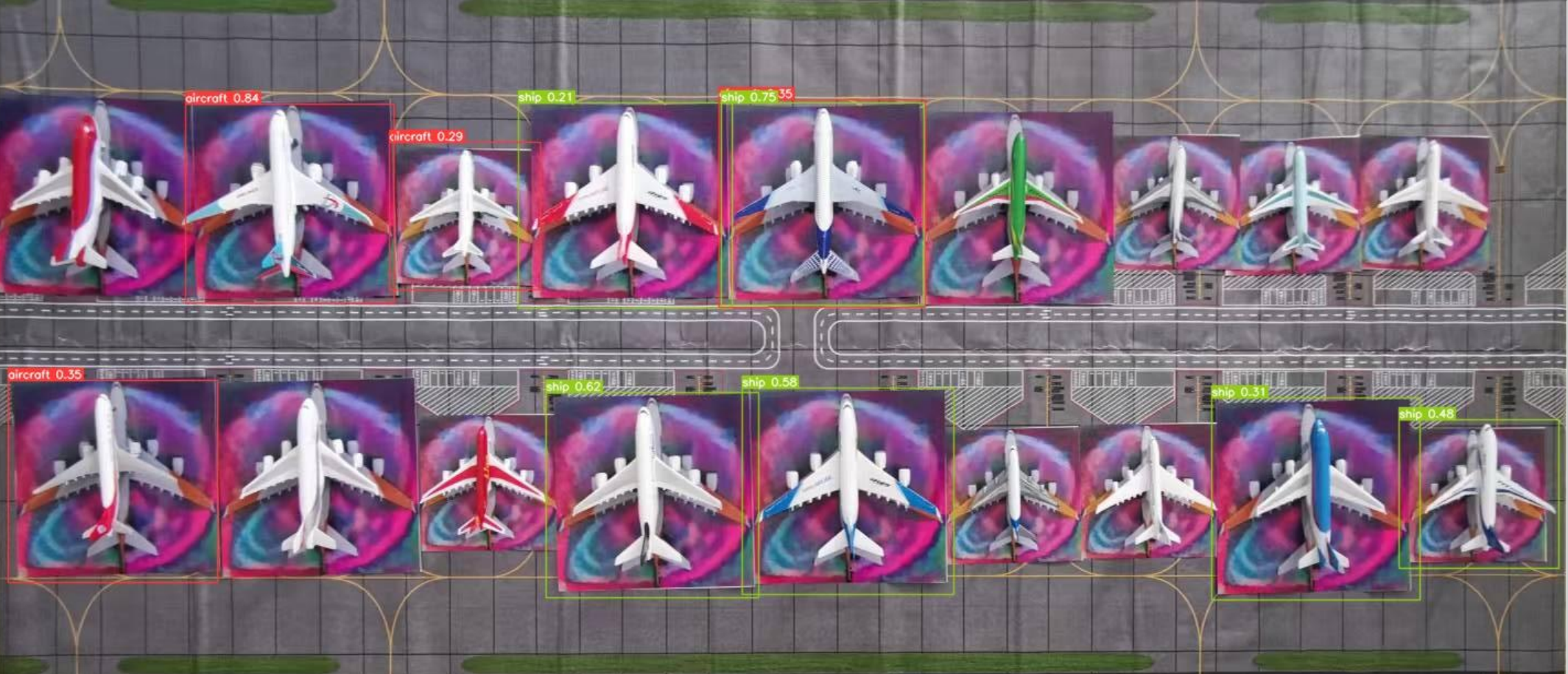}
    \caption{YOLOv5s}
  \end{subfigure}
  \begin{subfigure}{0.24\linewidth}
    \includegraphics[width=1\linewidth]{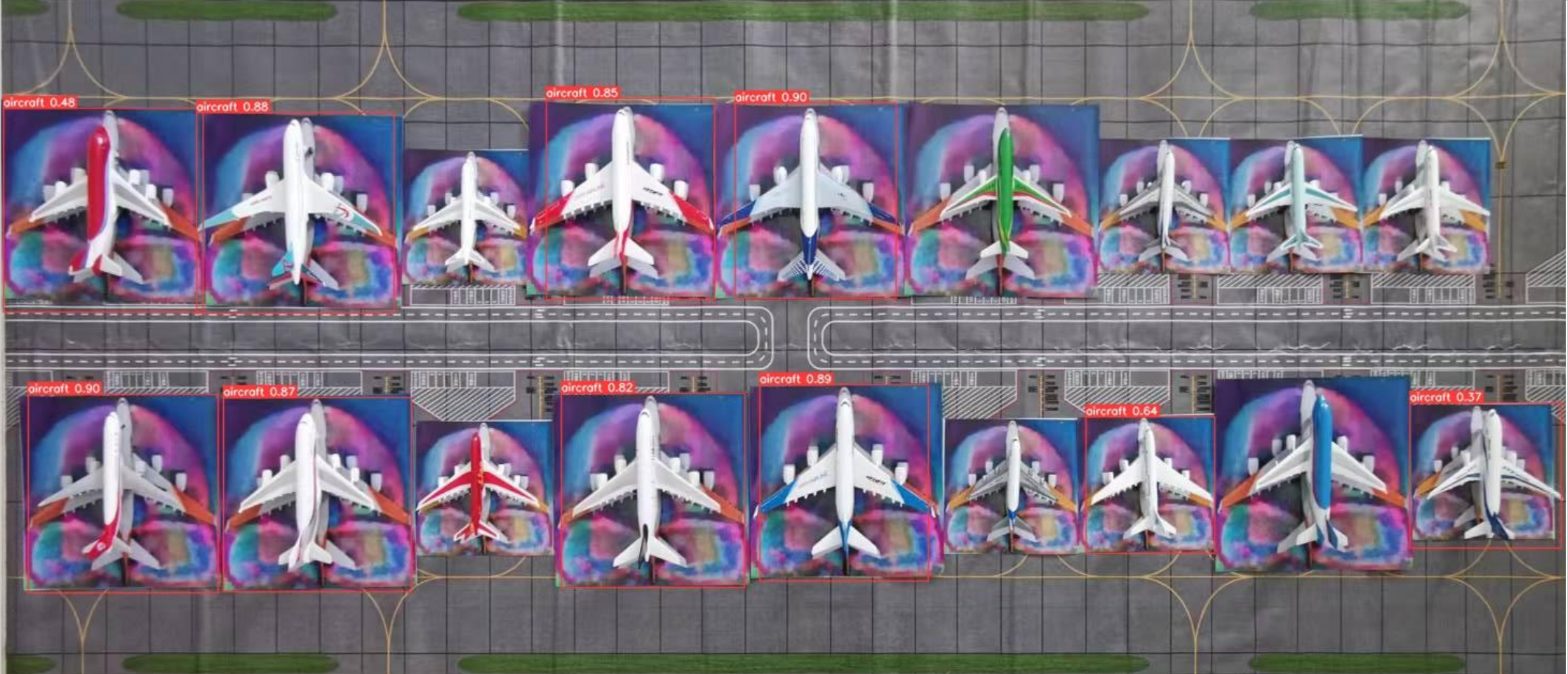}
    \caption{YOLOv5m}
  \end{subfigure}
  \begin{subfigure}{0.24\linewidth}
    \includegraphics[width=1\linewidth]{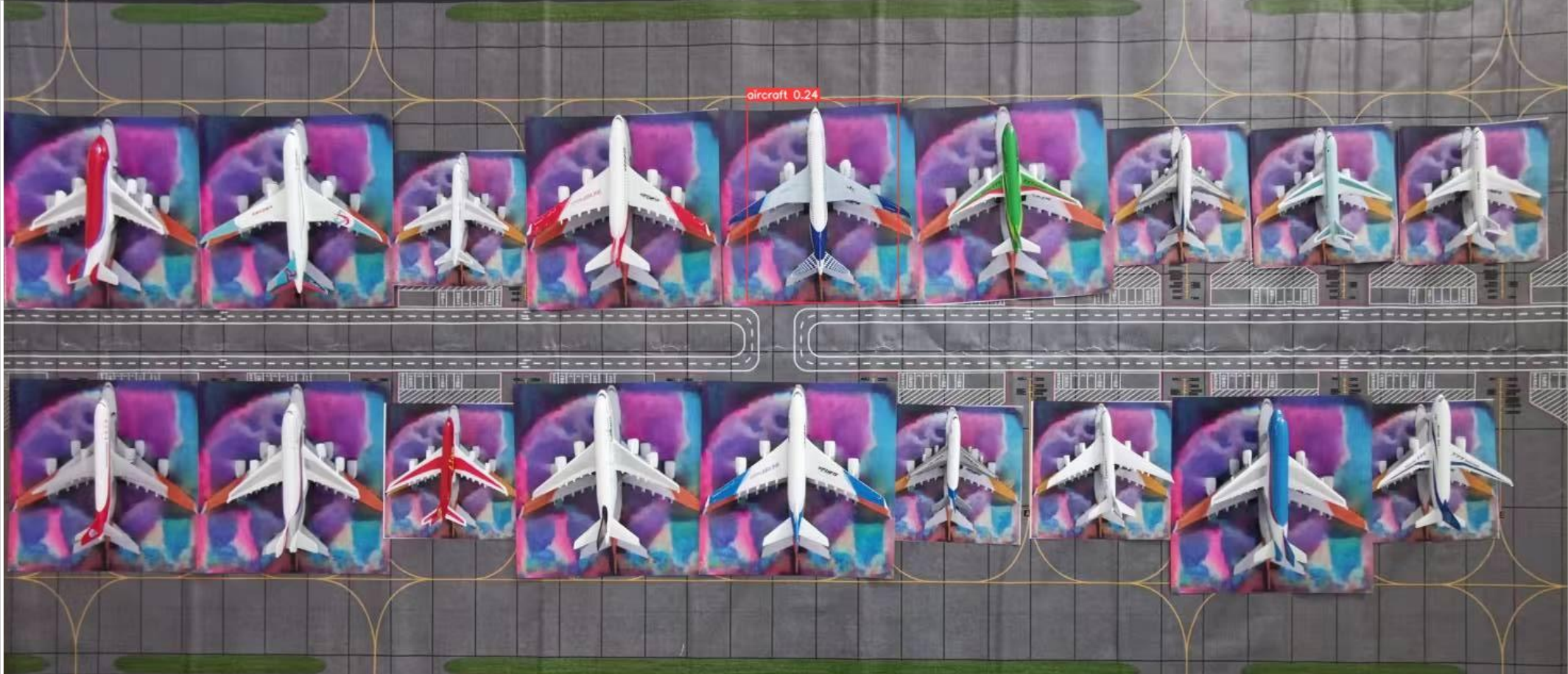}
    \caption{YOLOv5l}
  \end{subfigure}
  \begin{subfigure}{0.24\linewidth}
    \includegraphics[width=1\linewidth]{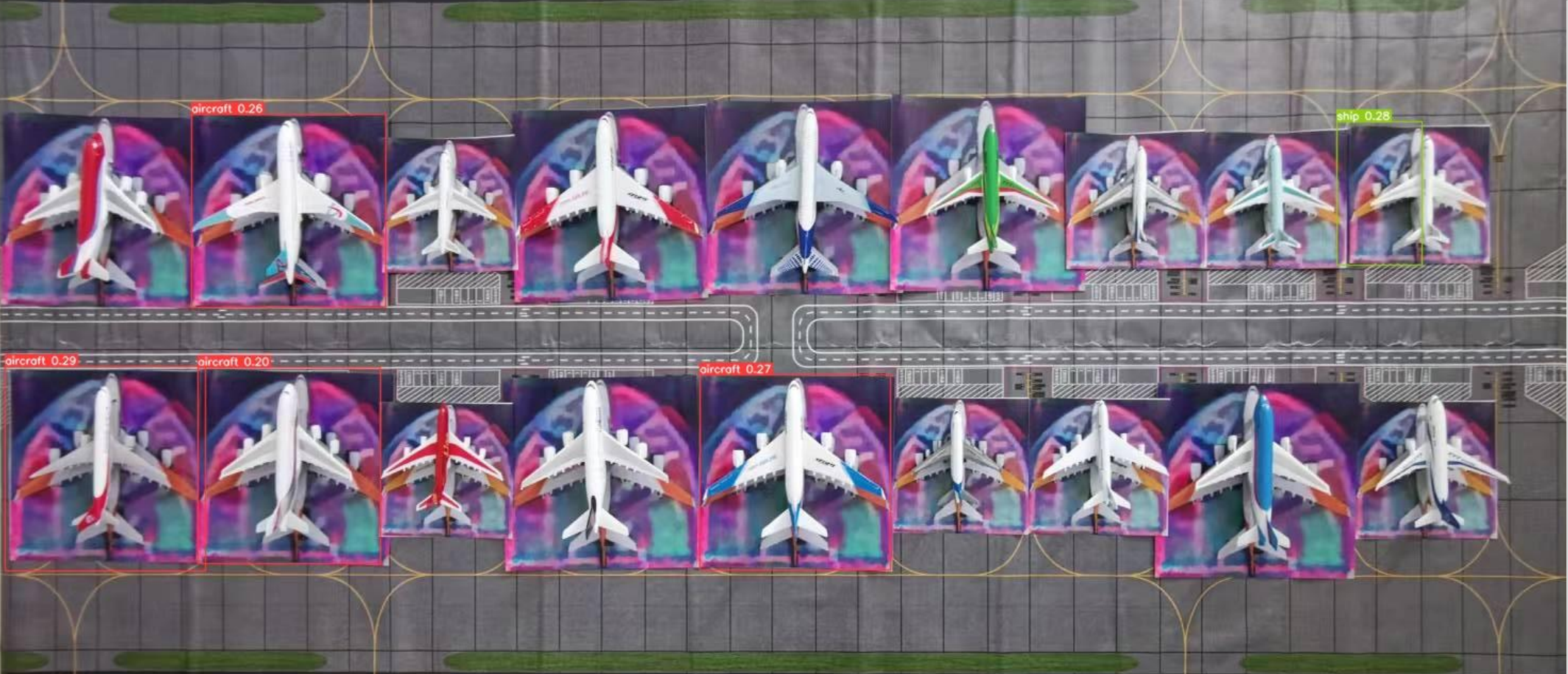}
    \caption{YOLOv5x}
  \end{subfigure}
  \begin{subfigure}{0.24\linewidth}
    \includegraphics[width=1\linewidth]{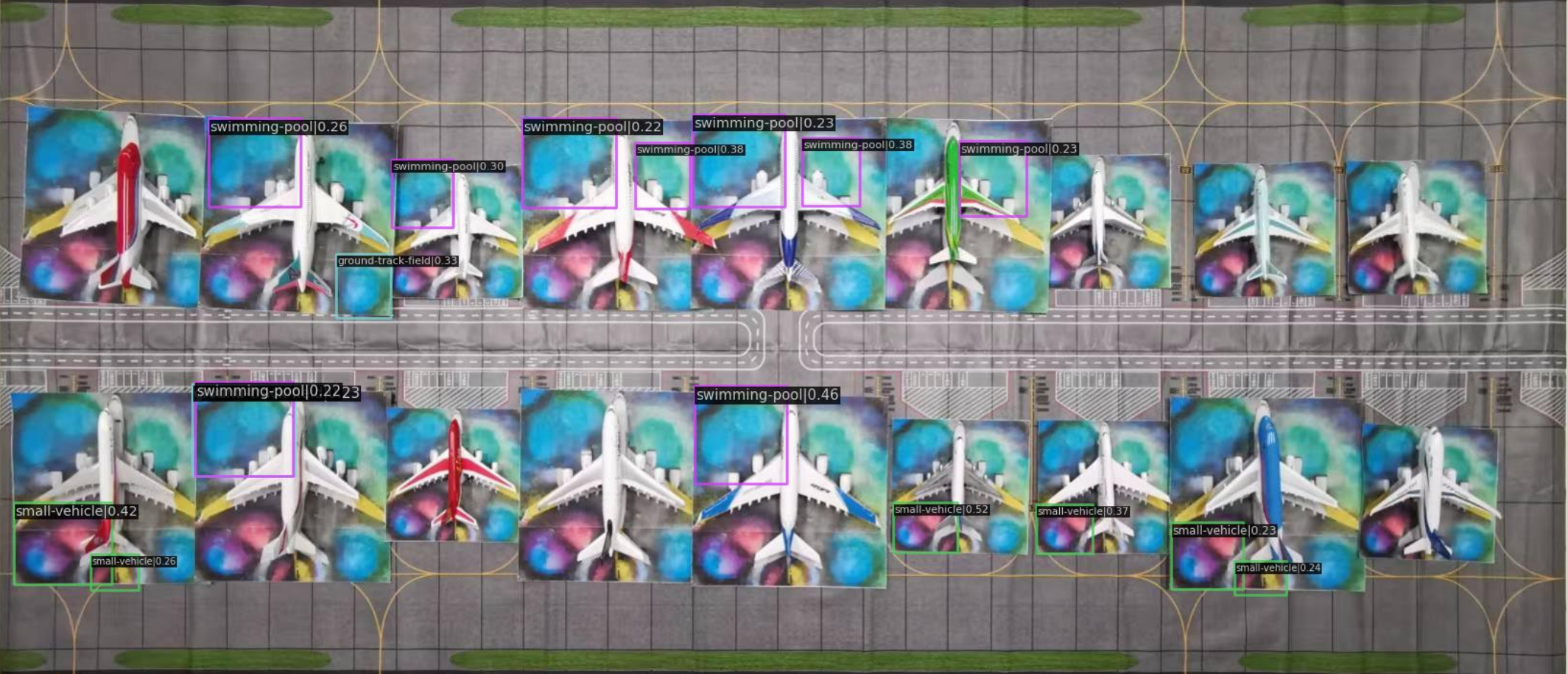}
    \caption{SSD}
  \end{subfigure}
  \begin{subfigure}{0.24\linewidth}
    \includegraphics[width=1\linewidth]{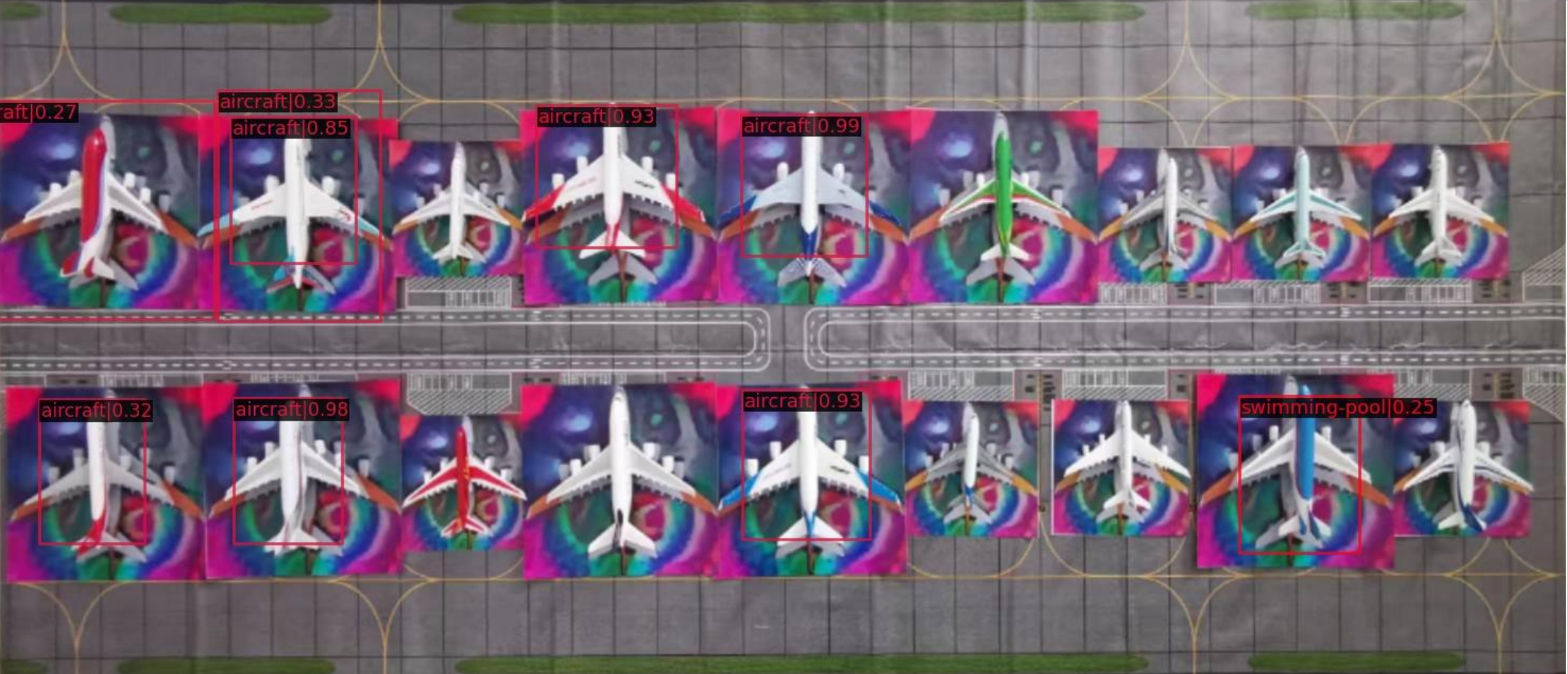}
    \caption{Faster R-CNN}
  \end{subfigure}
  \begin{subfigure}{0.24\linewidth}
    \includegraphics[width=1\linewidth]{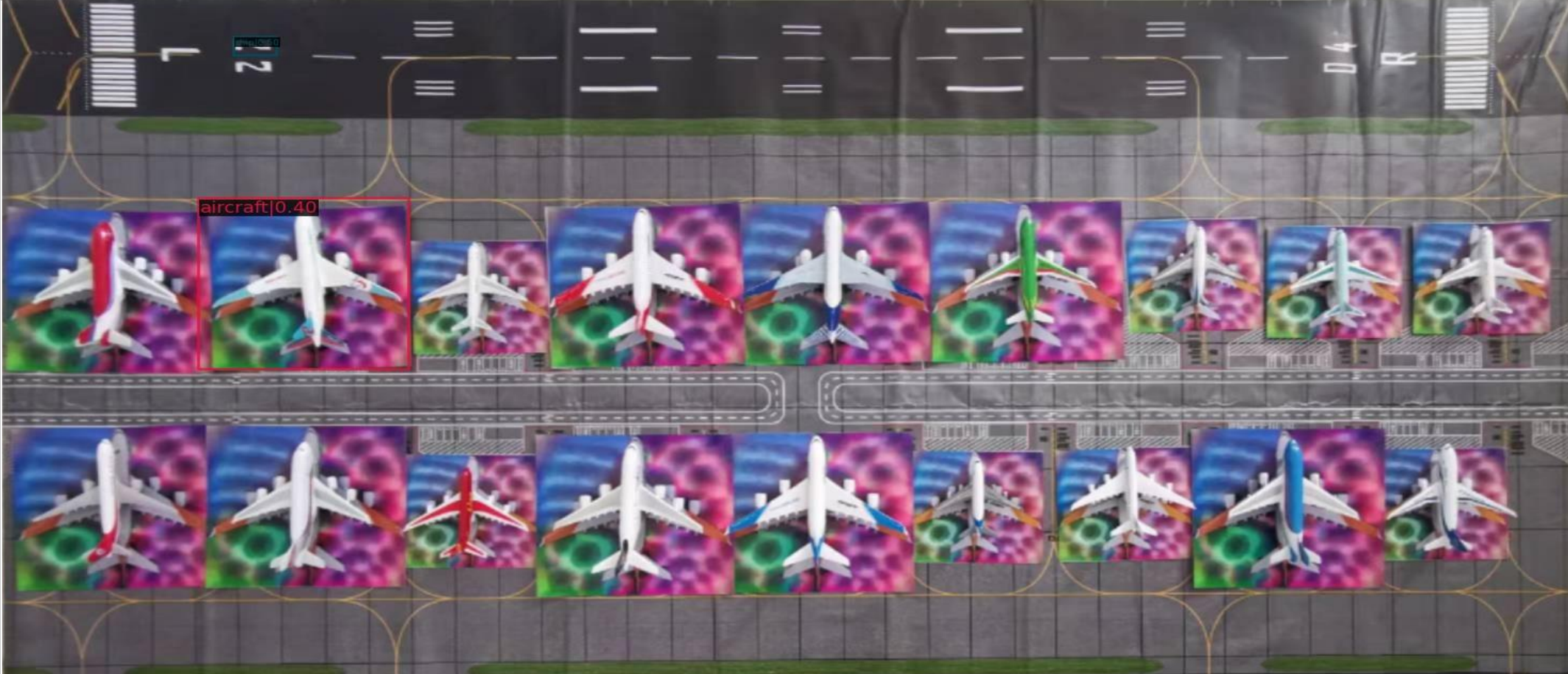}
    \caption{Swin Transformer}
  \end{subfigure}
  \begin{subfigure}{0.24\linewidth}
    \includegraphics[width=1\linewidth]{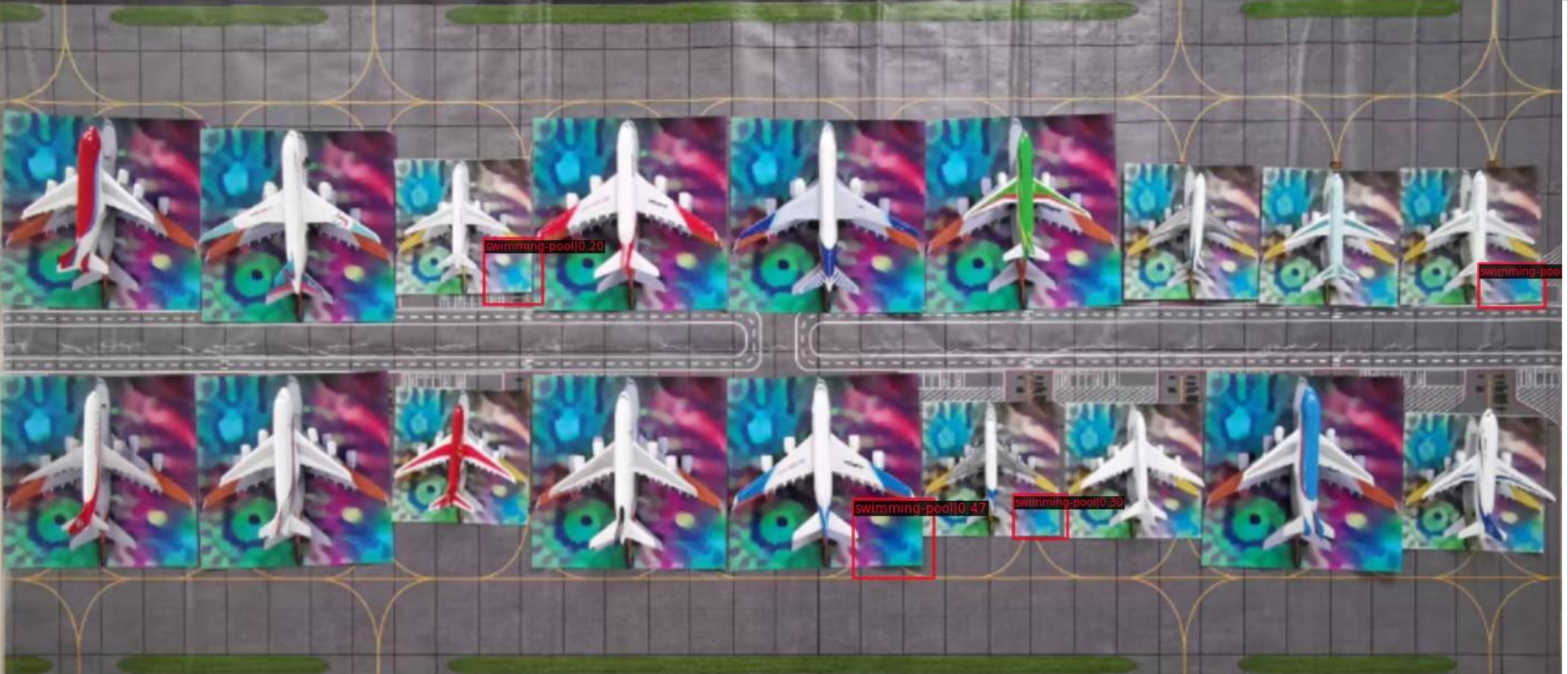}
    \caption{Cascade R-CNN}
  \end{subfigure}
  \begin{subfigure}{0.24\linewidth}
    \includegraphics[width=1\linewidth]{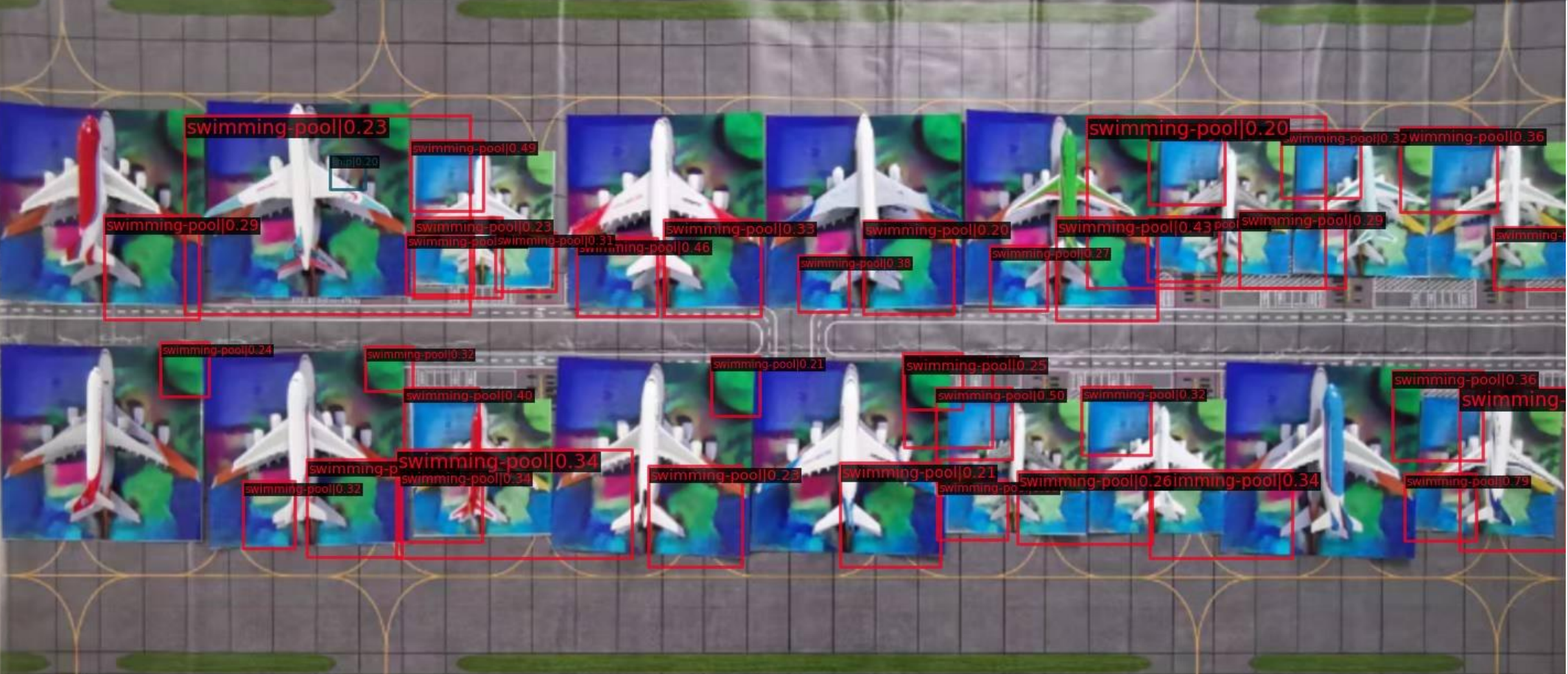}
    \caption{RetinaNet}
  \end{subfigure}
  \begin{subfigure}{0.24\linewidth}
    \includegraphics[width=1\linewidth]{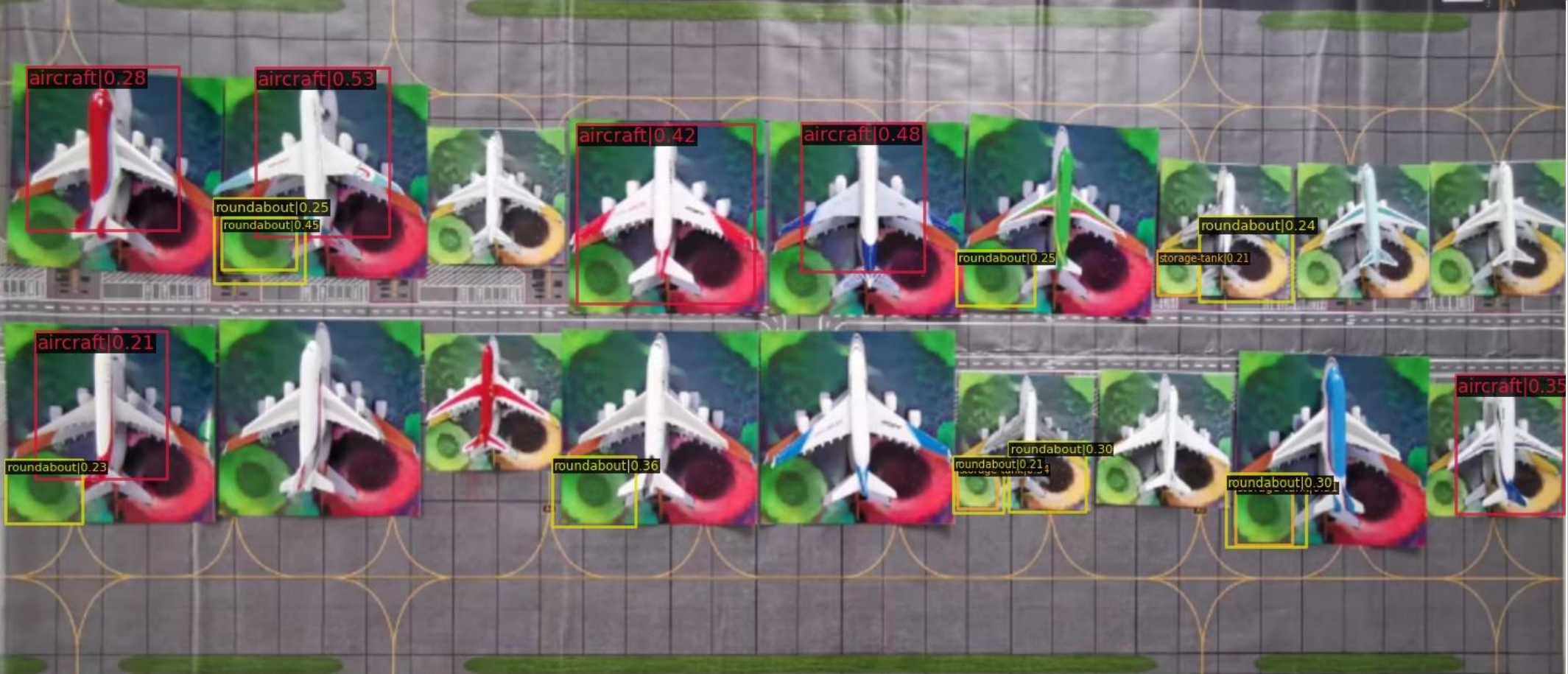}
    \caption{Mask R-CNN}
  \end{subfigure}
  \begin{subfigure}{0.24\linewidth}
    \includegraphics[width=1\linewidth]{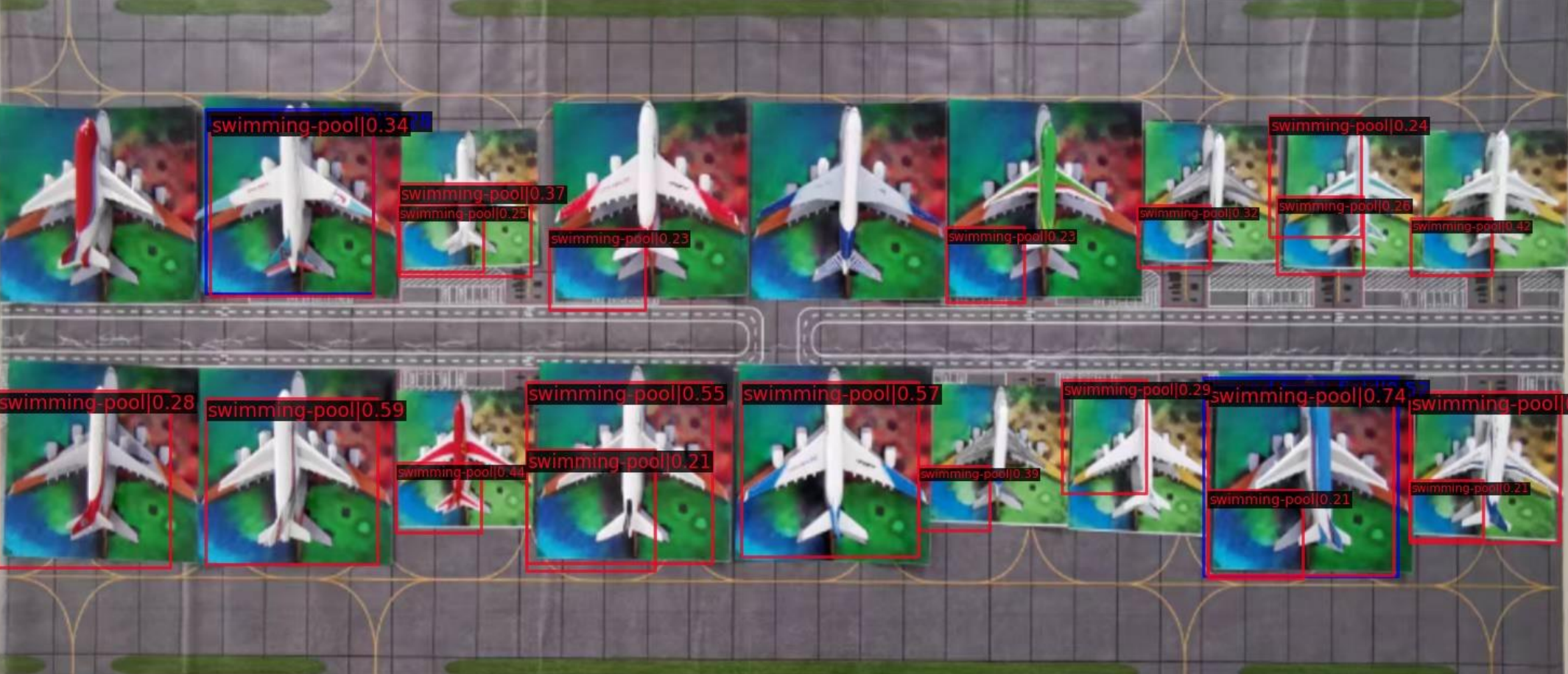}
    \caption{FoveaBox}
  \end{subfigure}
  \begin{subfigure}{0.24\linewidth}
    \includegraphics[width=1\linewidth]{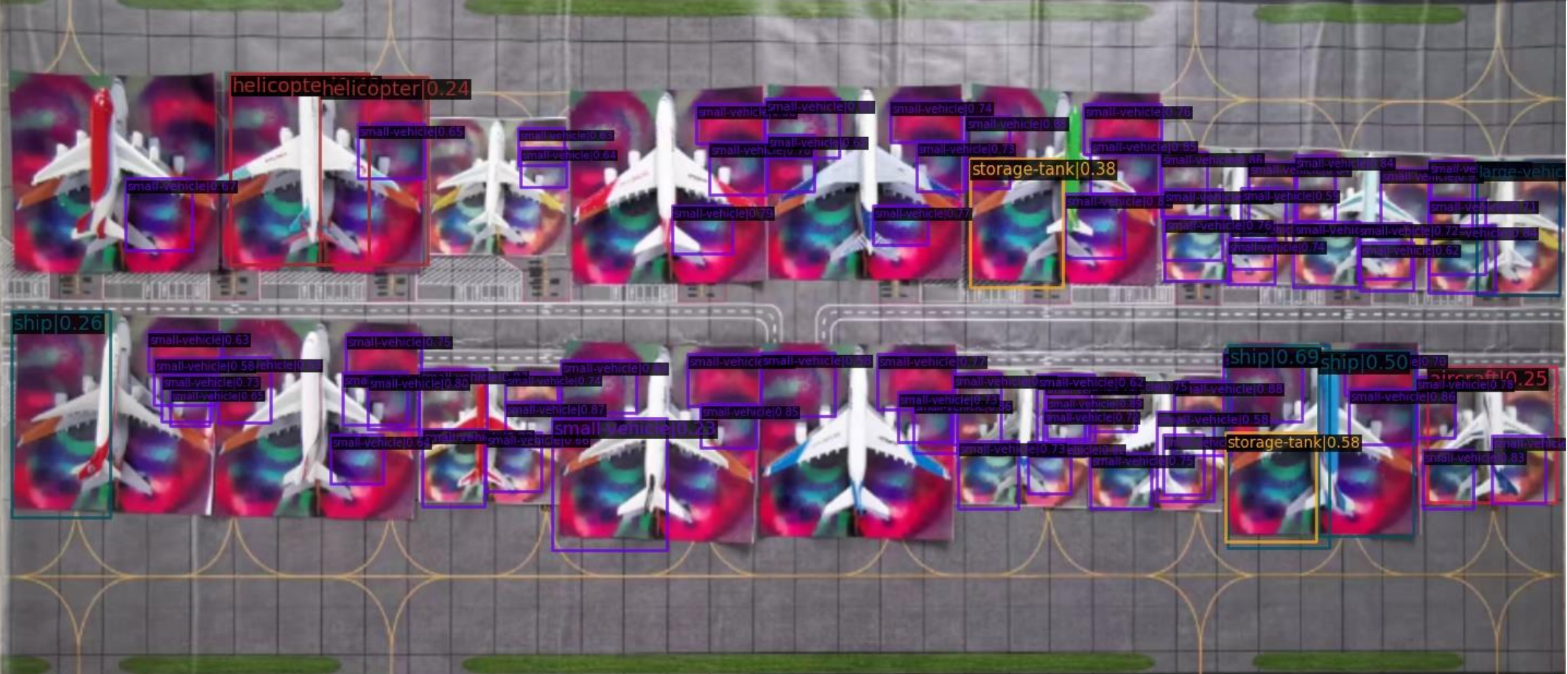}
    \caption{FreeAnchor}
  \end{subfigure}
  \begin{subfigure}{0.24\linewidth}
    \includegraphics[width=1\linewidth]{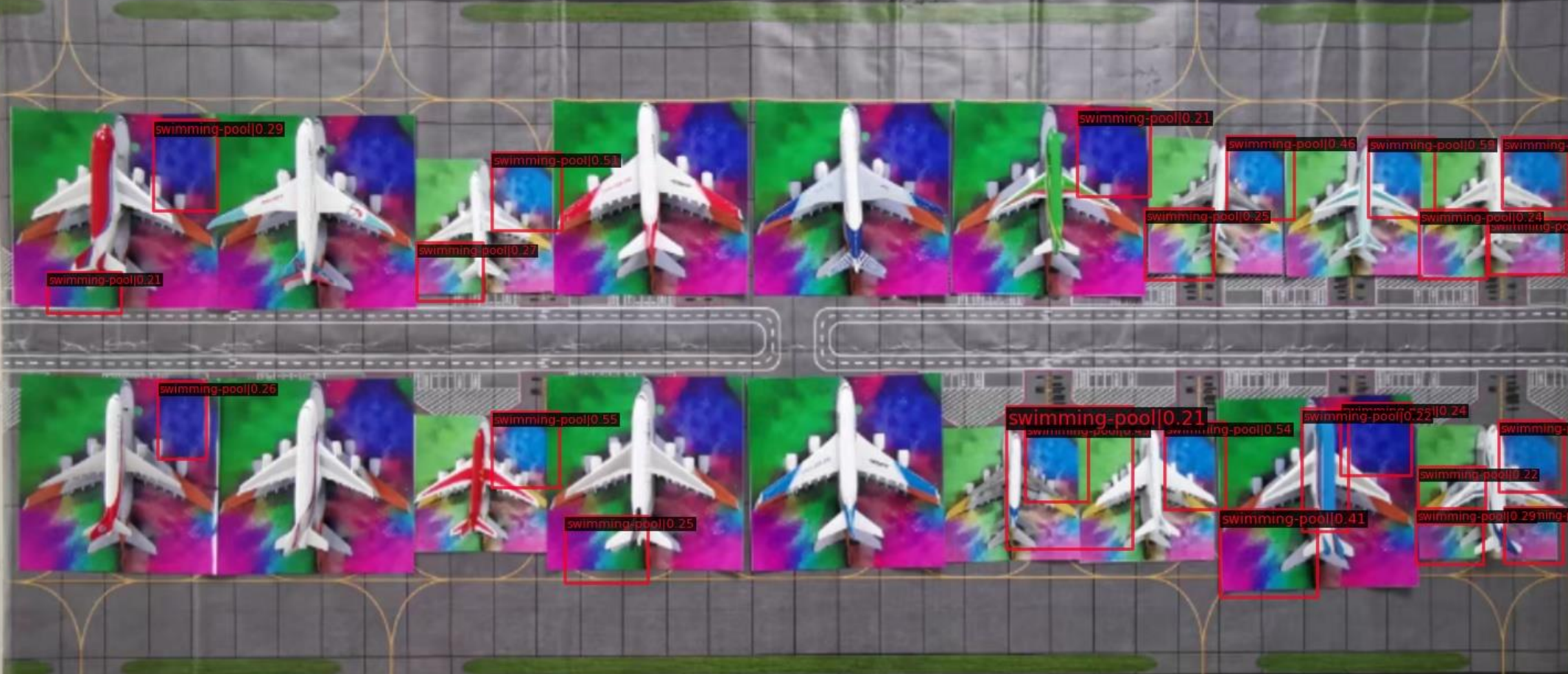}
    \caption{FSAF}
  \end{subfigure}
  \begin{subfigure}{0.24\linewidth}
    \includegraphics[width=1\linewidth]{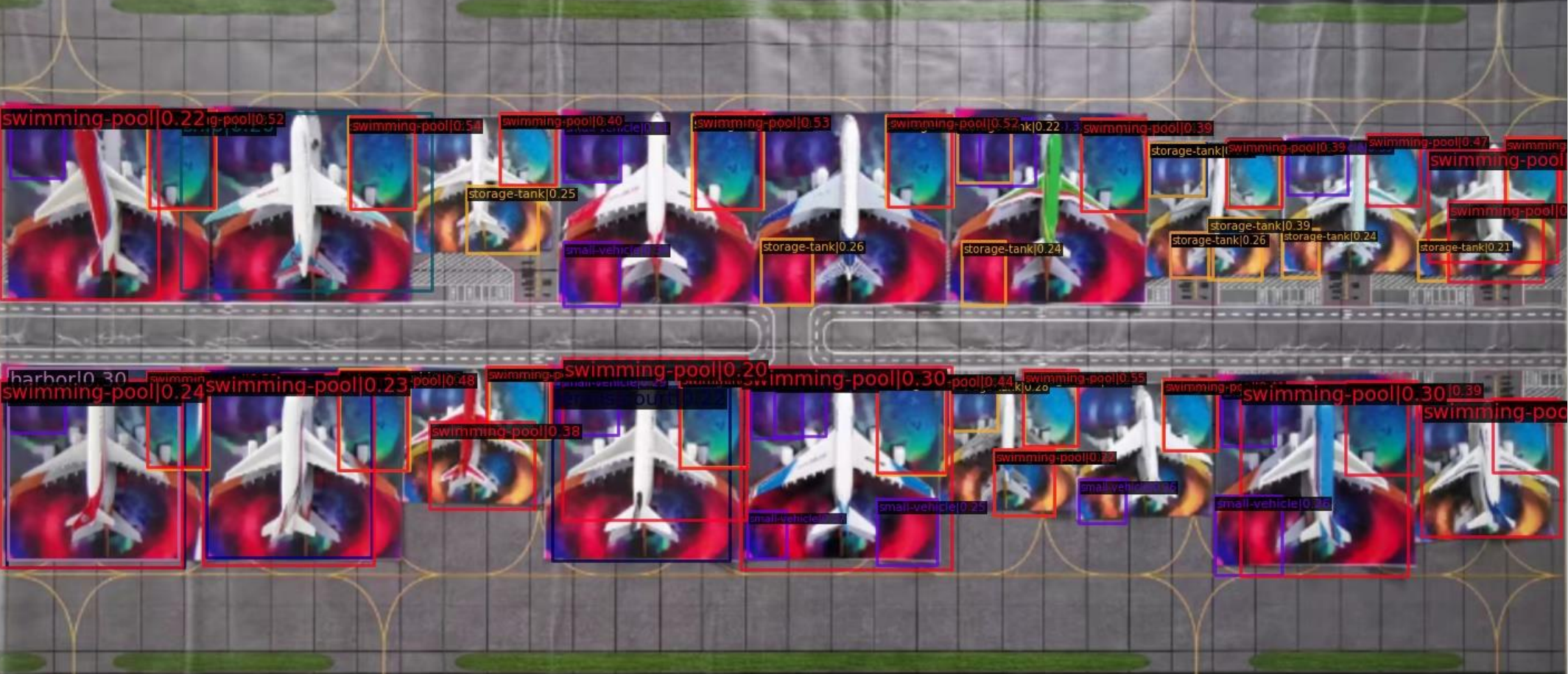}
    \caption{RepPoints}
  \end{subfigure}
  \begin{subfigure}{0.24\linewidth}
    \includegraphics[width=1\linewidth]{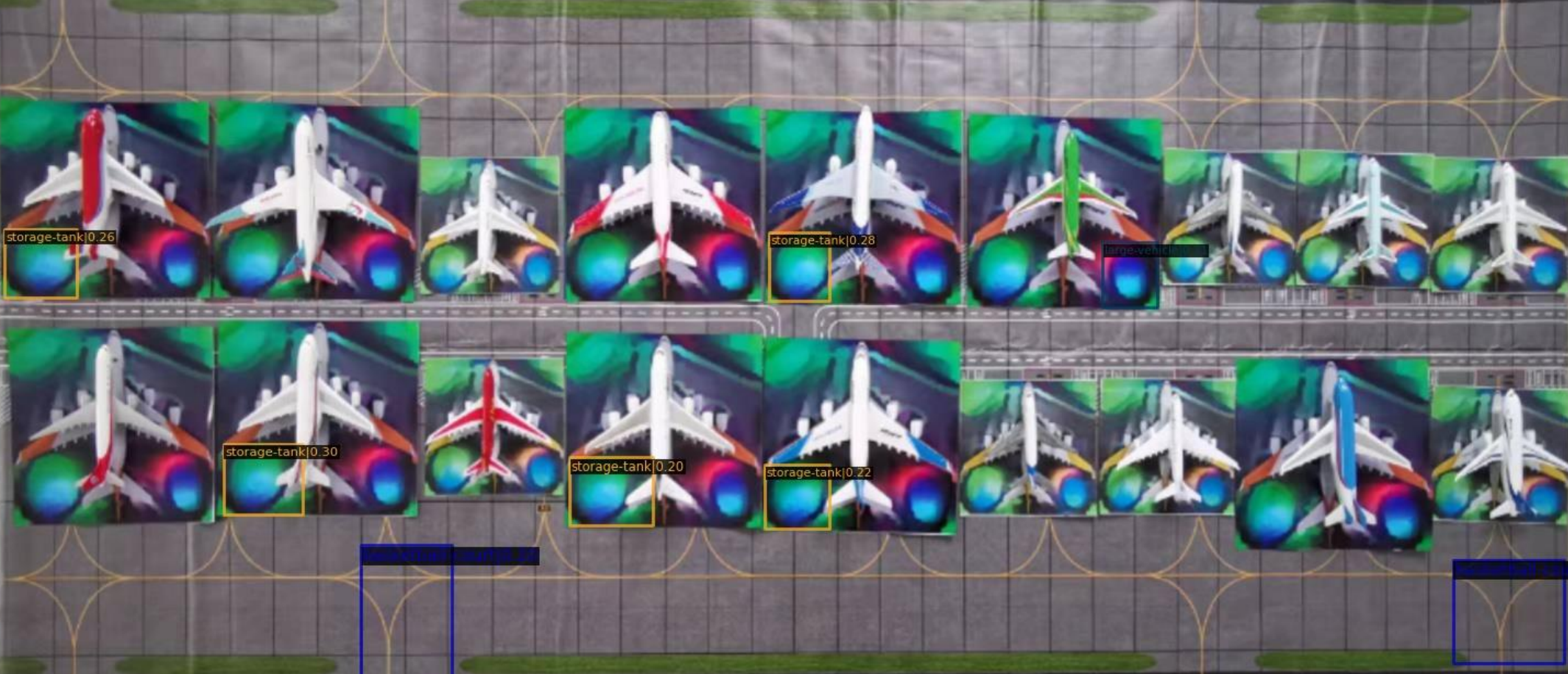}
    \caption{TOOD}
  \end{subfigure}
  \begin{subfigure}{0.24\linewidth}
    \includegraphics[width=1\linewidth]{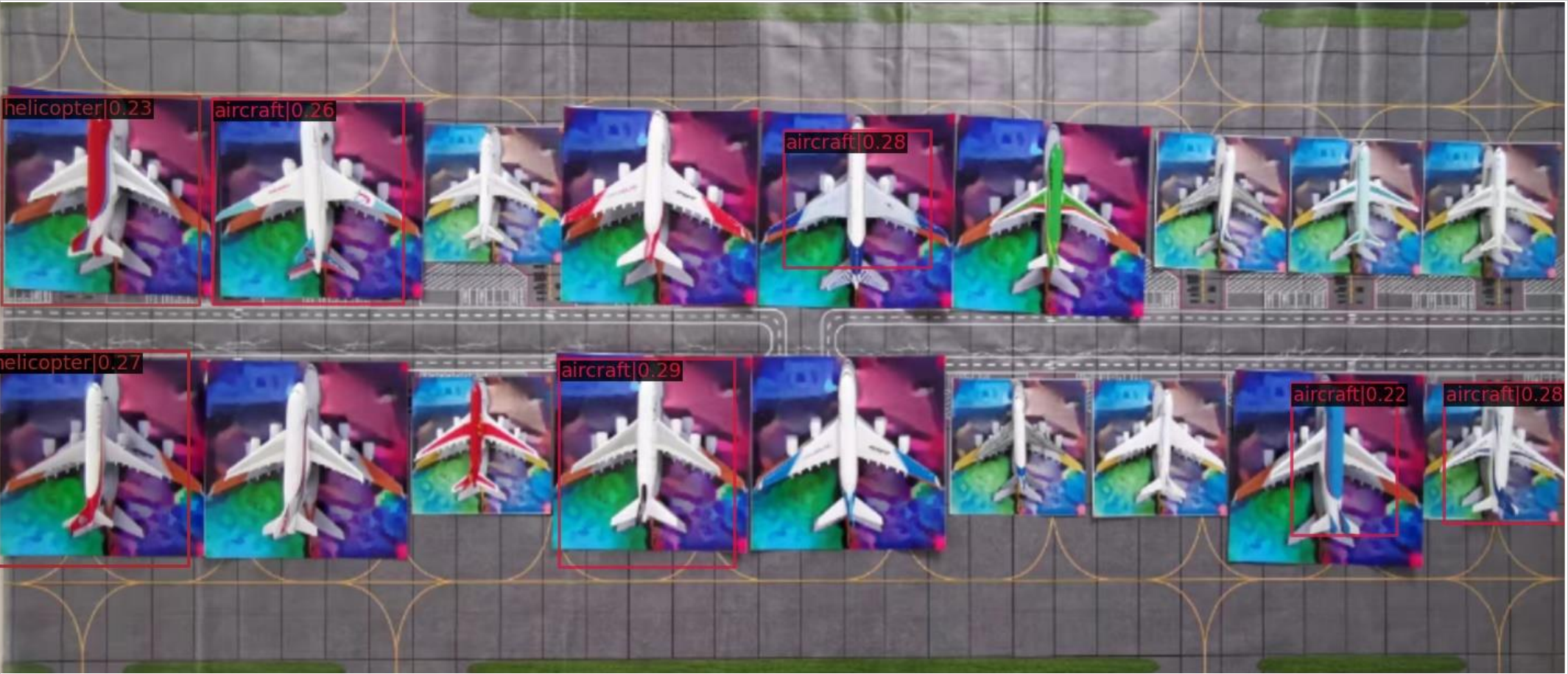}
    \caption{ATSS}
  \end{subfigure}
  \begin{subfigure}{0.24\linewidth}
    \includegraphics[width=1\linewidth]{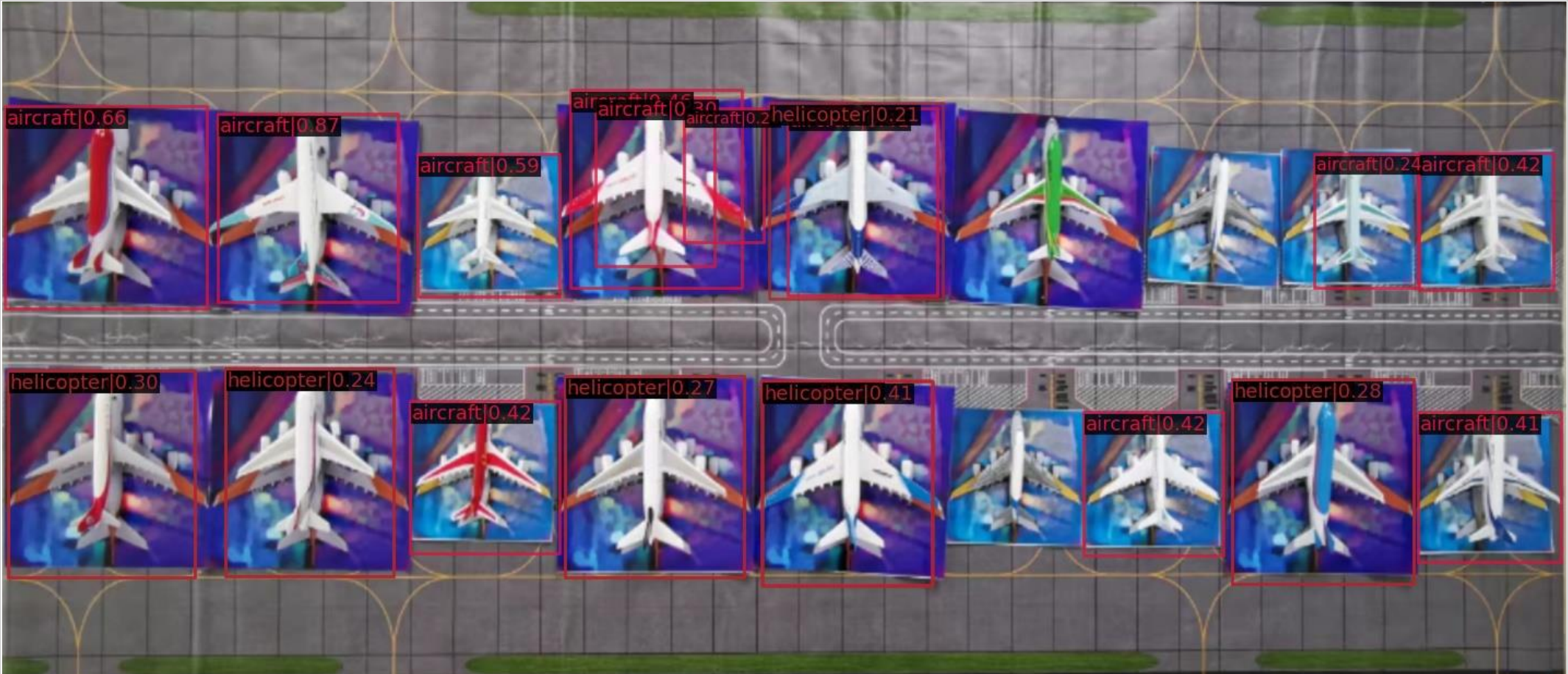}
    \caption{VarifocalNet}
  \end{subfigure}
  \caption{Qualitative results of white-box attacks against twenty different aerial detectors in the physical world.}
  \label{fig:white_box_physical_attack_qualitative_results}
\end{figure*}

\begin{figure*}
  \centering
  \begin{subfigure}{0.24\linewidth}
    \includegraphics[width=1\linewidth]{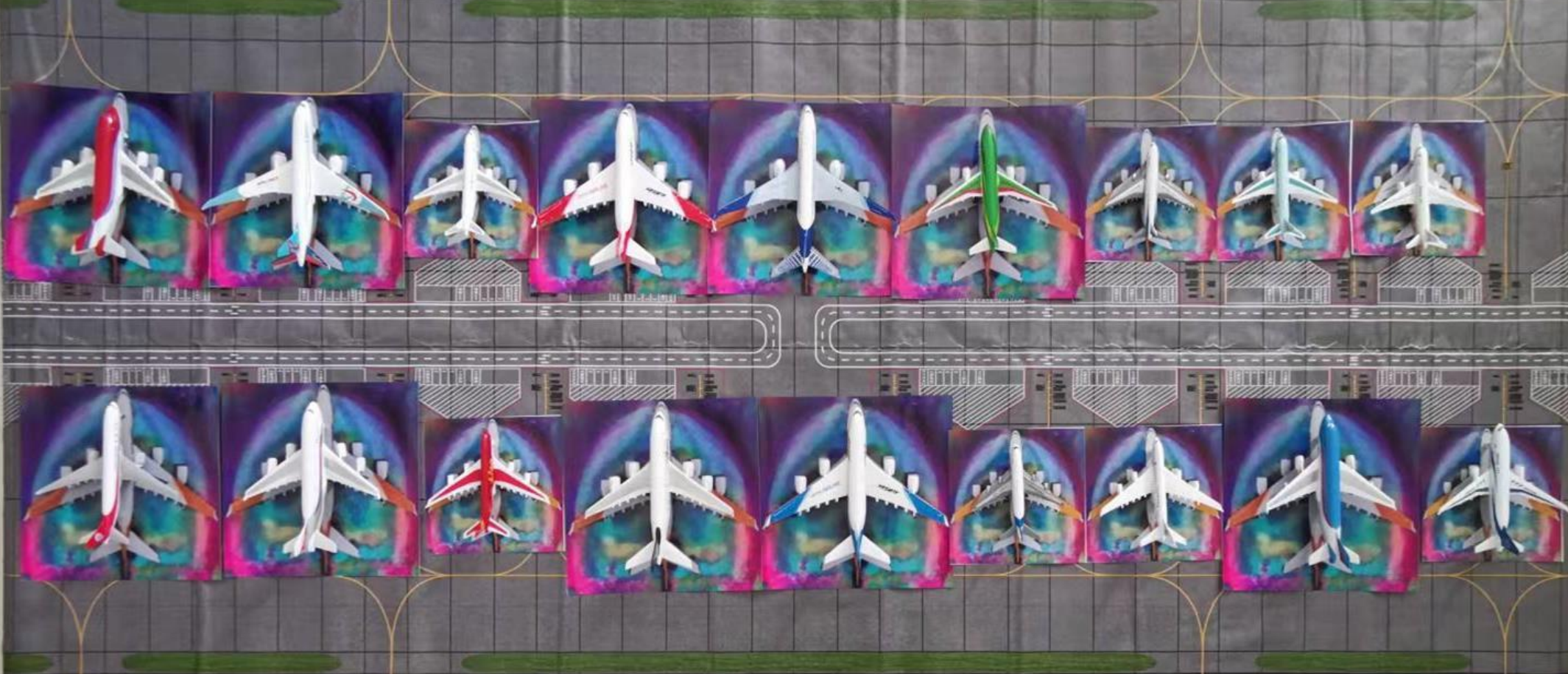}
    \caption{YOLOv2}
  \end{subfigure}
  \begin{subfigure}{0.24\linewidth}
    \includegraphics[width=1\linewidth]{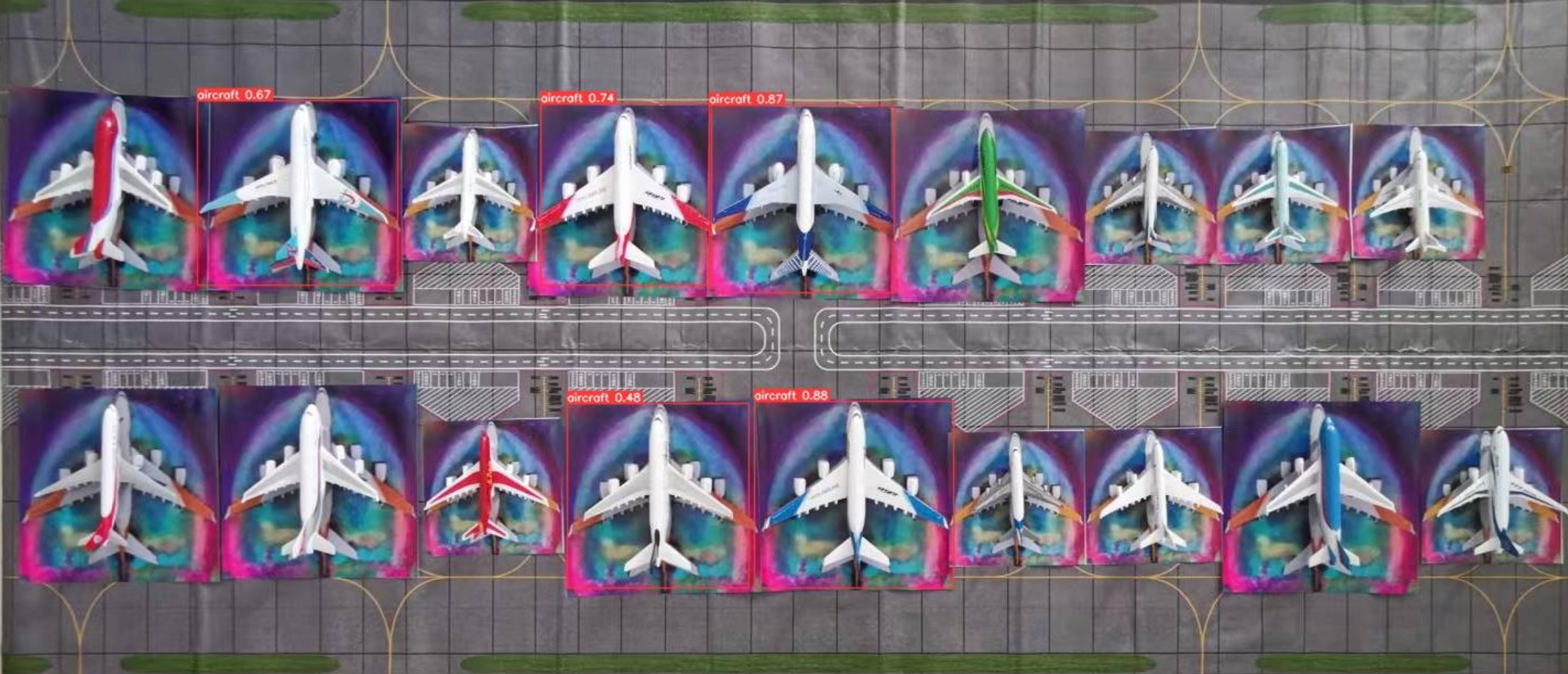}
    \caption{YOLOv3}
  \end{subfigure}
  \begin{subfigure}{0.24\linewidth}
    \includegraphics[width=1\linewidth]{yolov5n_A3_yolov5n.pdf}
    \caption{YOLOv5n}
  \end{subfigure}
  \begin{subfigure}{0.24\linewidth}
    \includegraphics[width=1\linewidth]{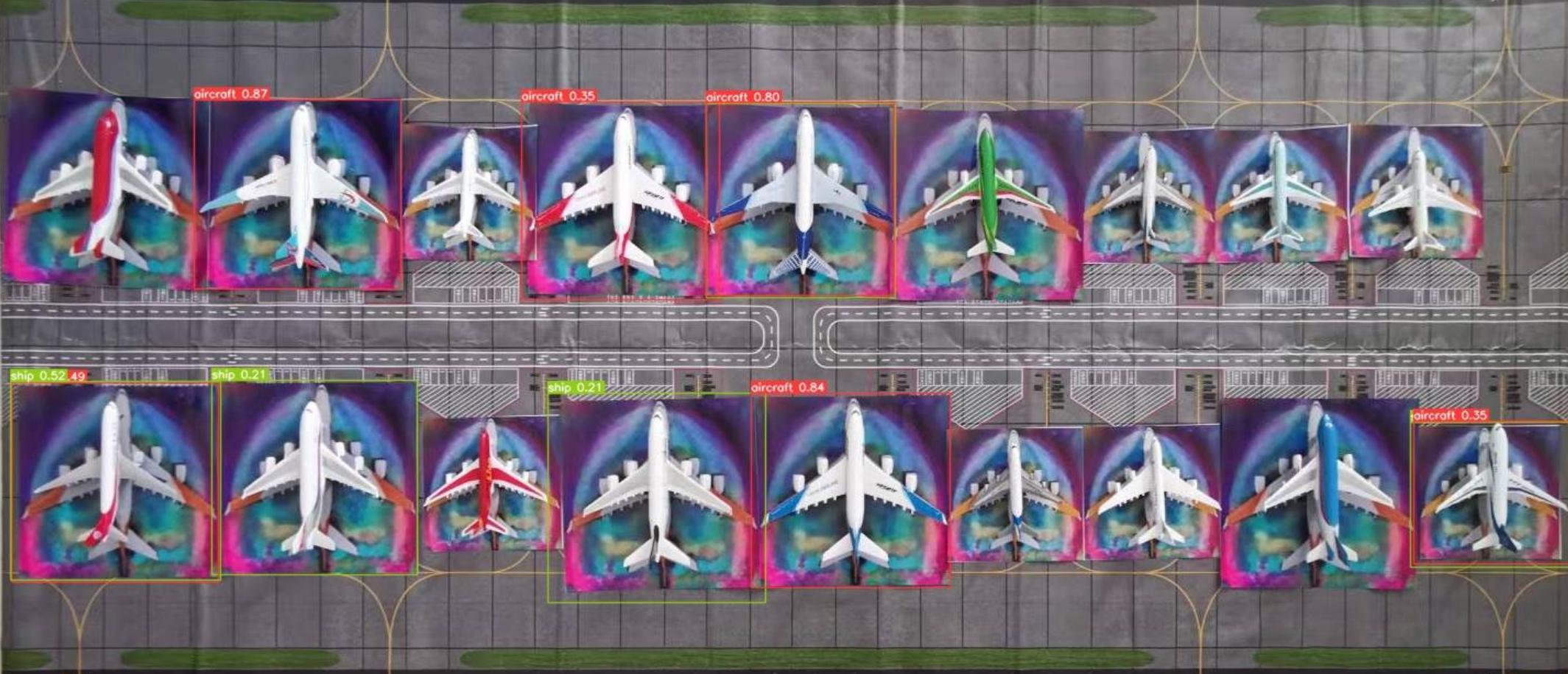}
    \caption{YOLOv5s}
  \end{subfigure}
  \begin{subfigure}{0.24\linewidth}
    \includegraphics[width=1\linewidth]{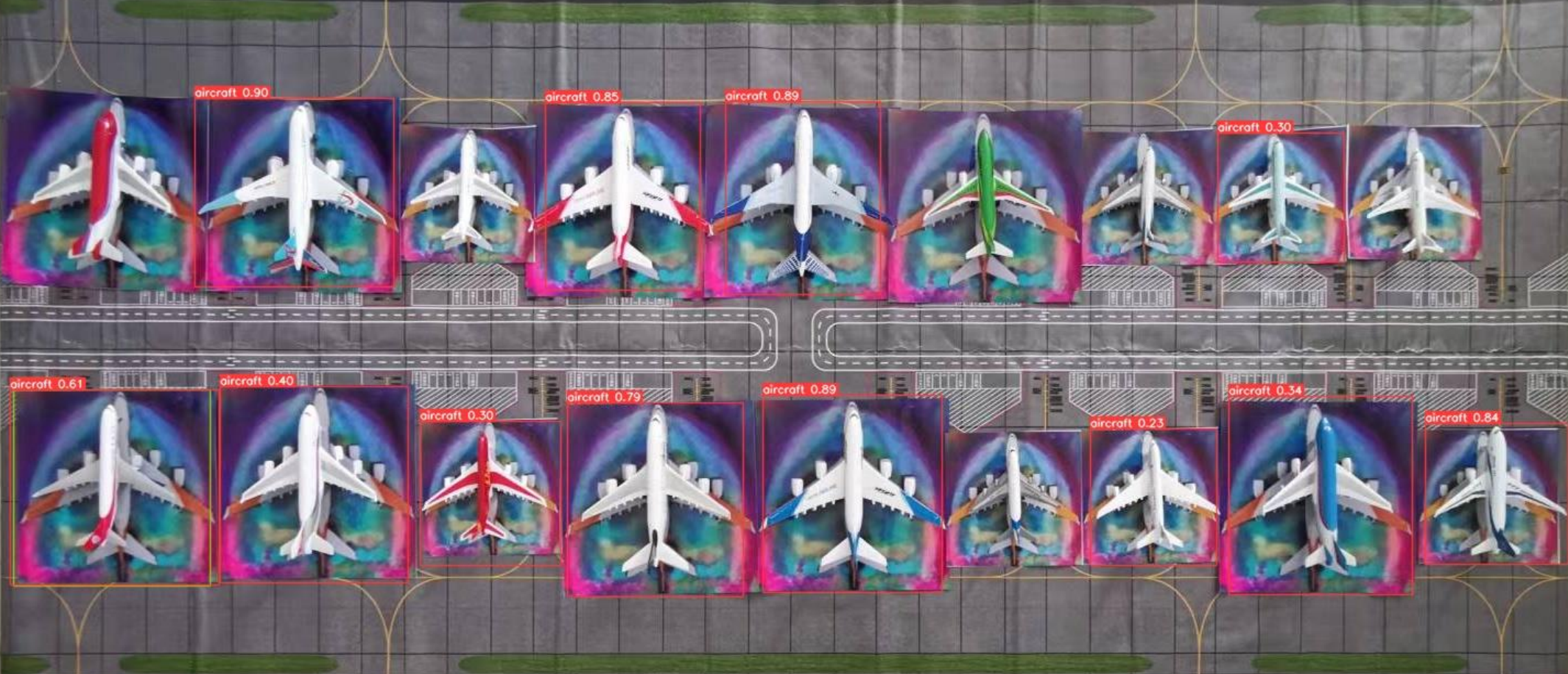}
    \caption{YOLOv5m}
  \end{subfigure}
  \begin{subfigure}{0.24\linewidth}
    \includegraphics[width=1\linewidth]{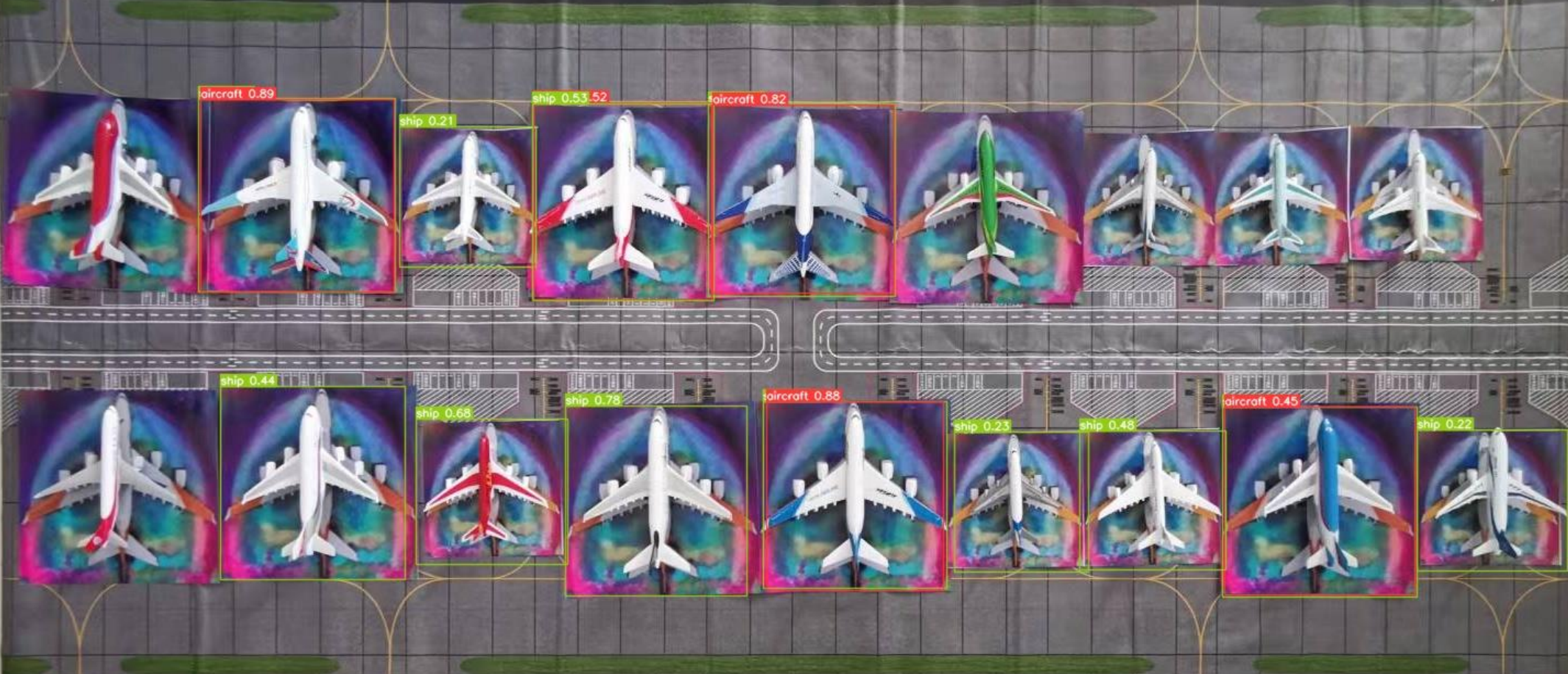}
    \caption{YOLOv5l}
  \end{subfigure}
  \begin{subfigure}{0.24\linewidth}
    \includegraphics[width=1\linewidth]{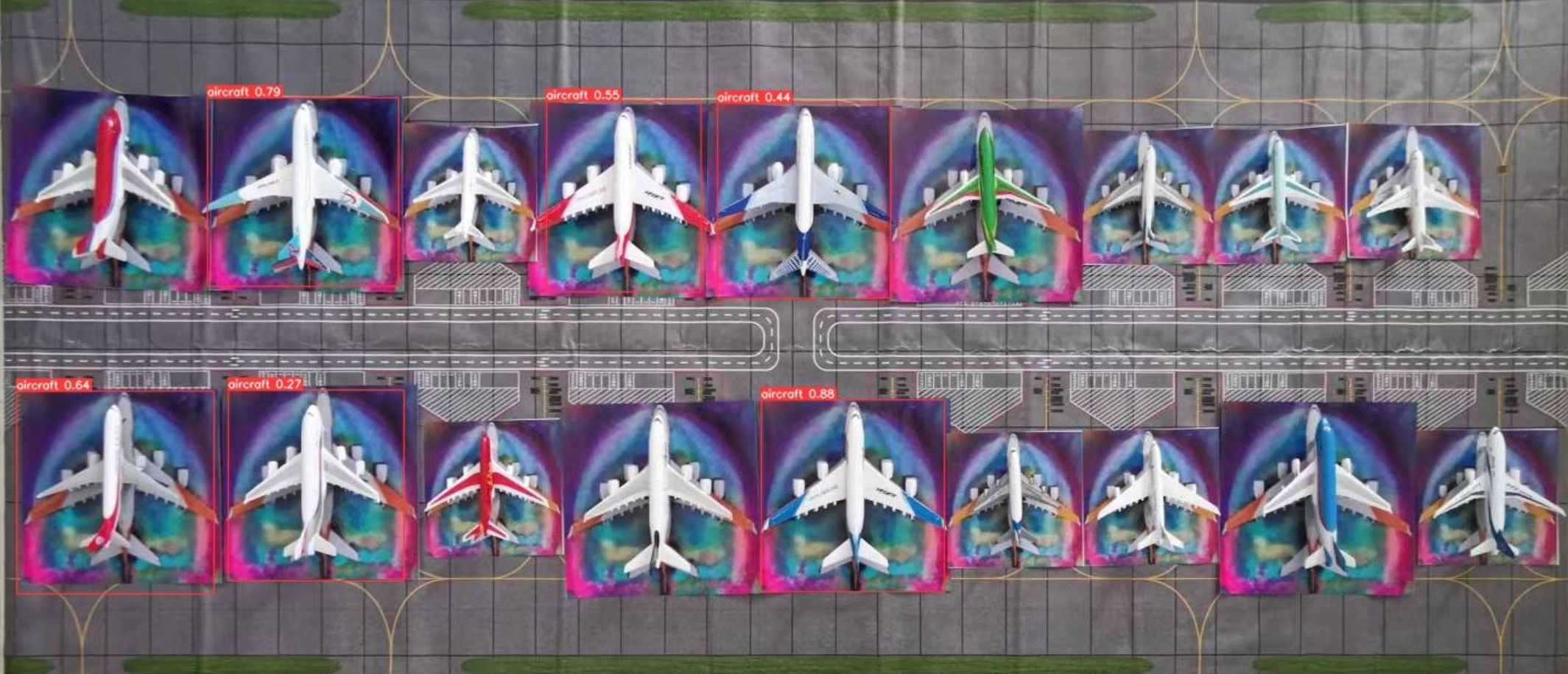}
    \caption{YOLOv5x}
  \end{subfigure}
  \begin{subfigure}{0.24\linewidth}
    \includegraphics[width=1\linewidth]{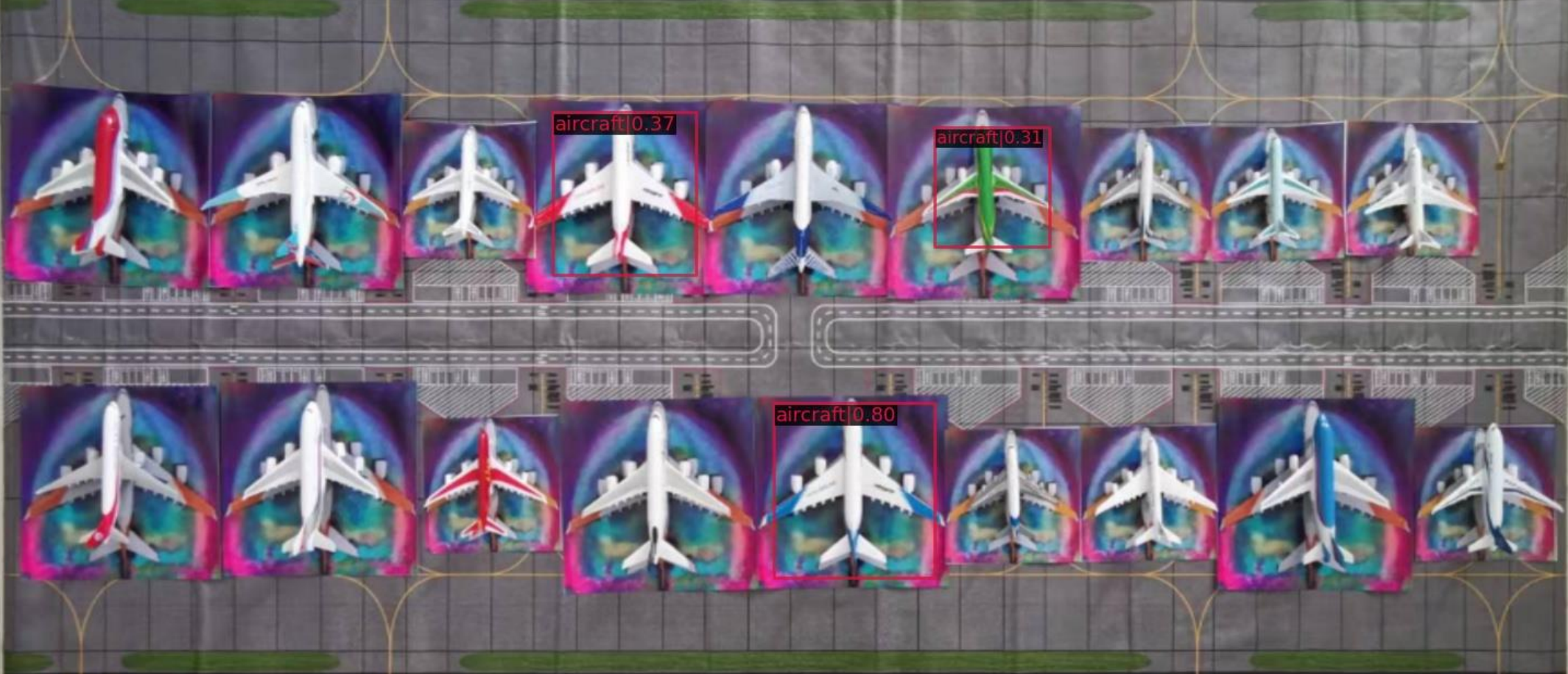}
    \caption{SSD}
  \end{subfigure}
  \begin{subfigure}{0.24\linewidth}
    \includegraphics[width=1\linewidth]{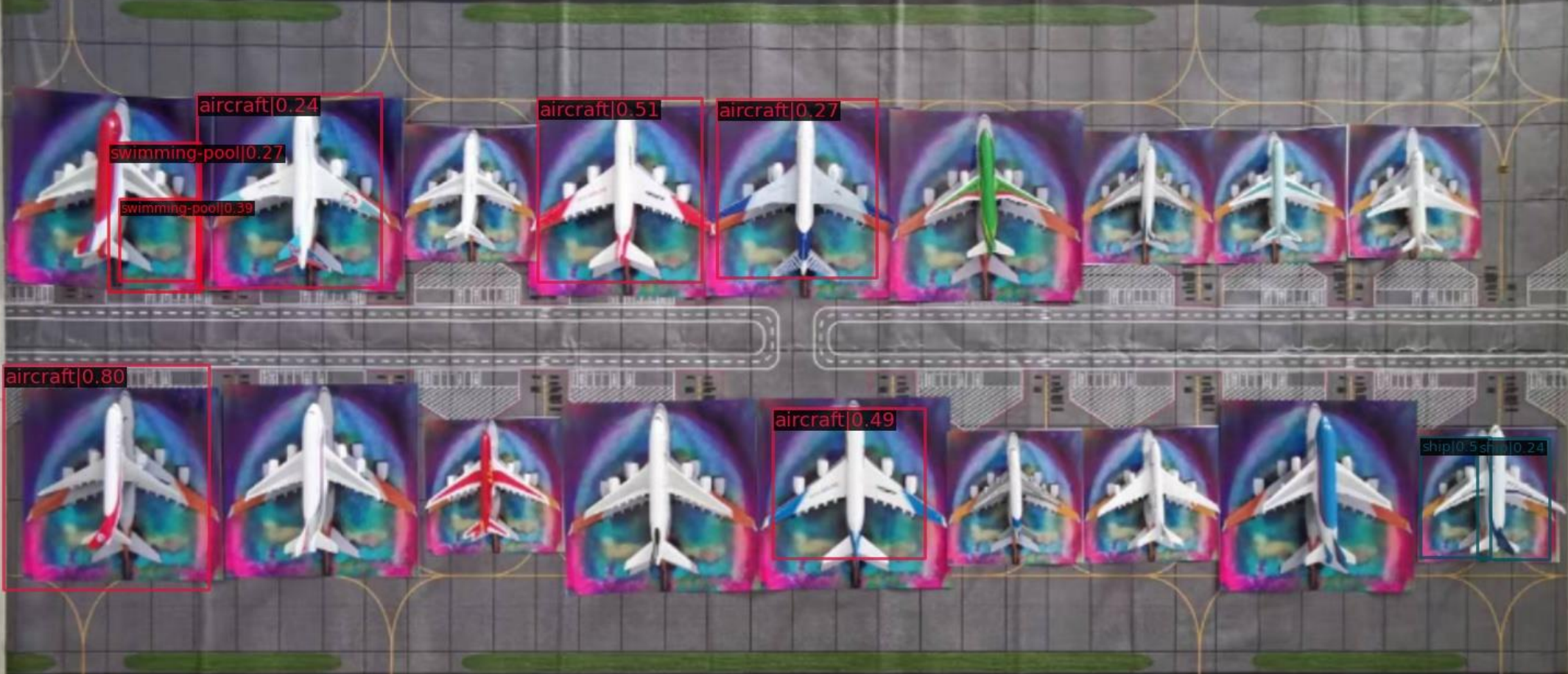}
    \caption{Faster R-CNN}
  \end{subfigure}
  \begin{subfigure}{0.24\linewidth}
    \includegraphics[width=1\linewidth]{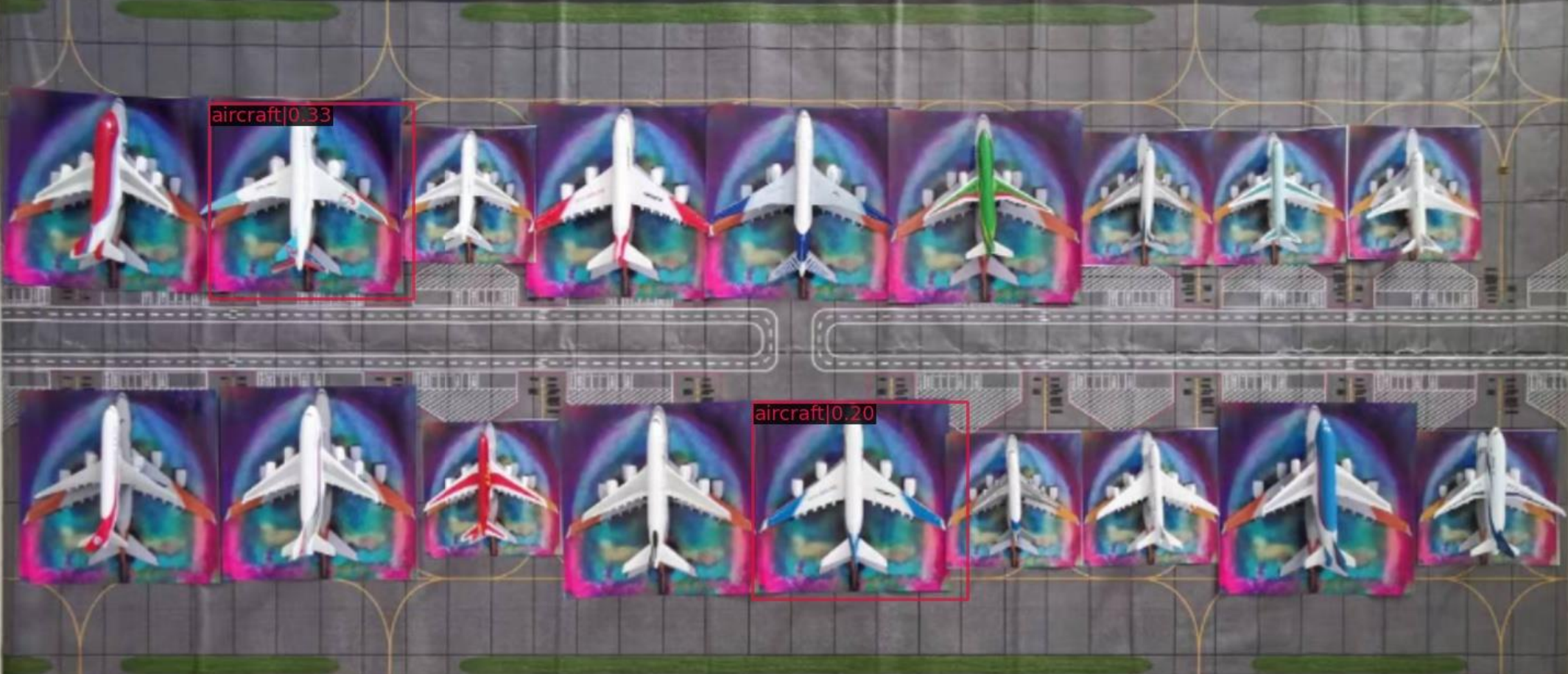}
    \caption{Swin Transformer}
  \end{subfigure}
  \begin{subfigure}{0.24\linewidth}
    \includegraphics[width=1\linewidth]{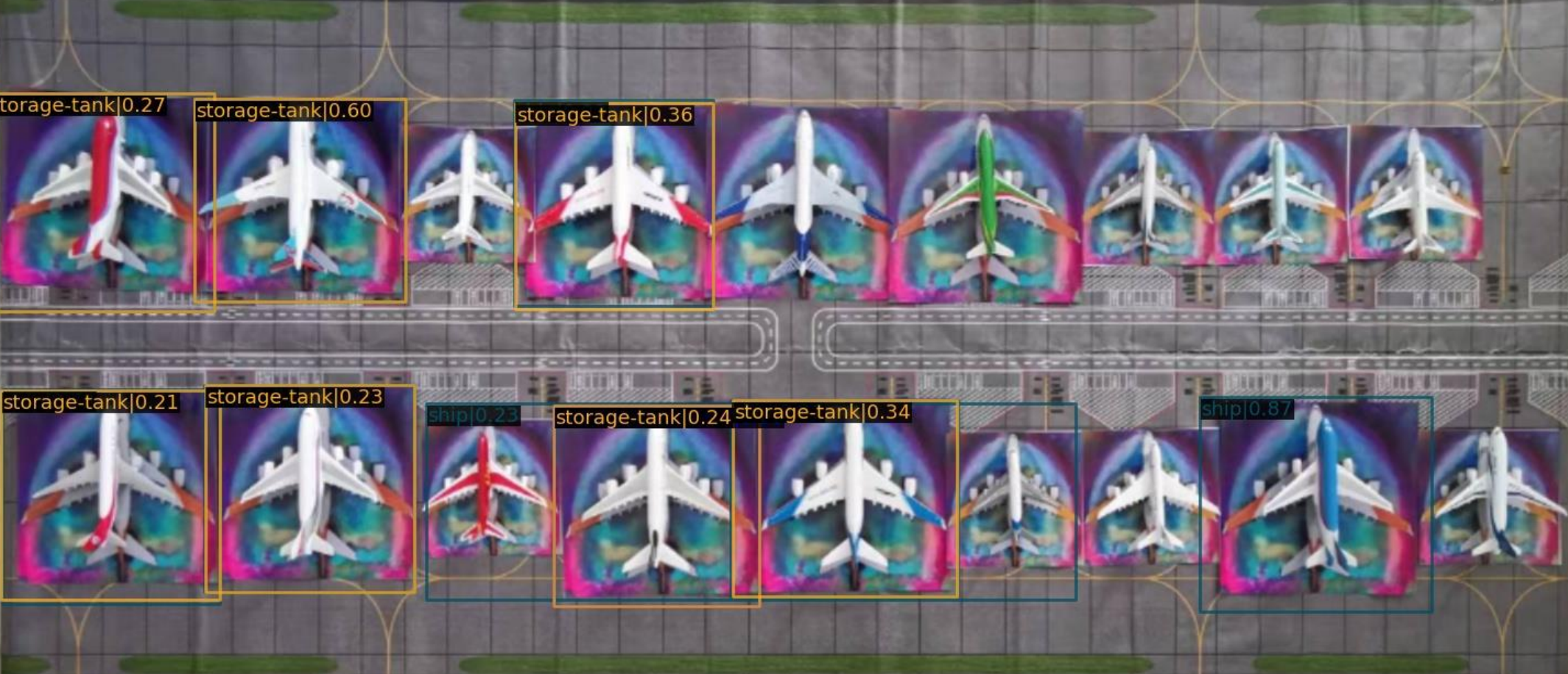}
    \caption{Cascade R-CNN}
  \end{subfigure}
  \begin{subfigure}{0.24\linewidth}
    \includegraphics[width=1\linewidth]{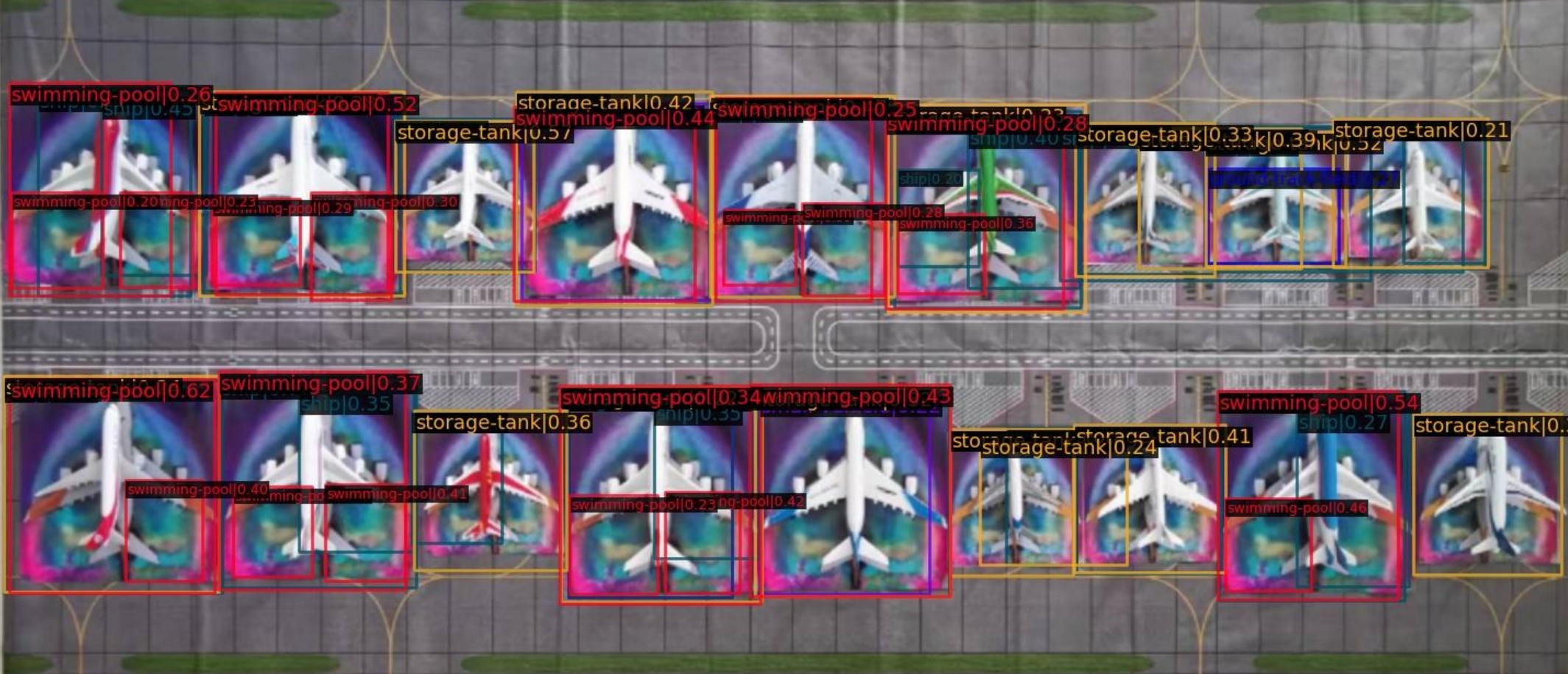}
    \caption{RetinaNet}
  \end{subfigure}
  \begin{subfigure}{0.24\linewidth}
    \includegraphics[width=1\linewidth]{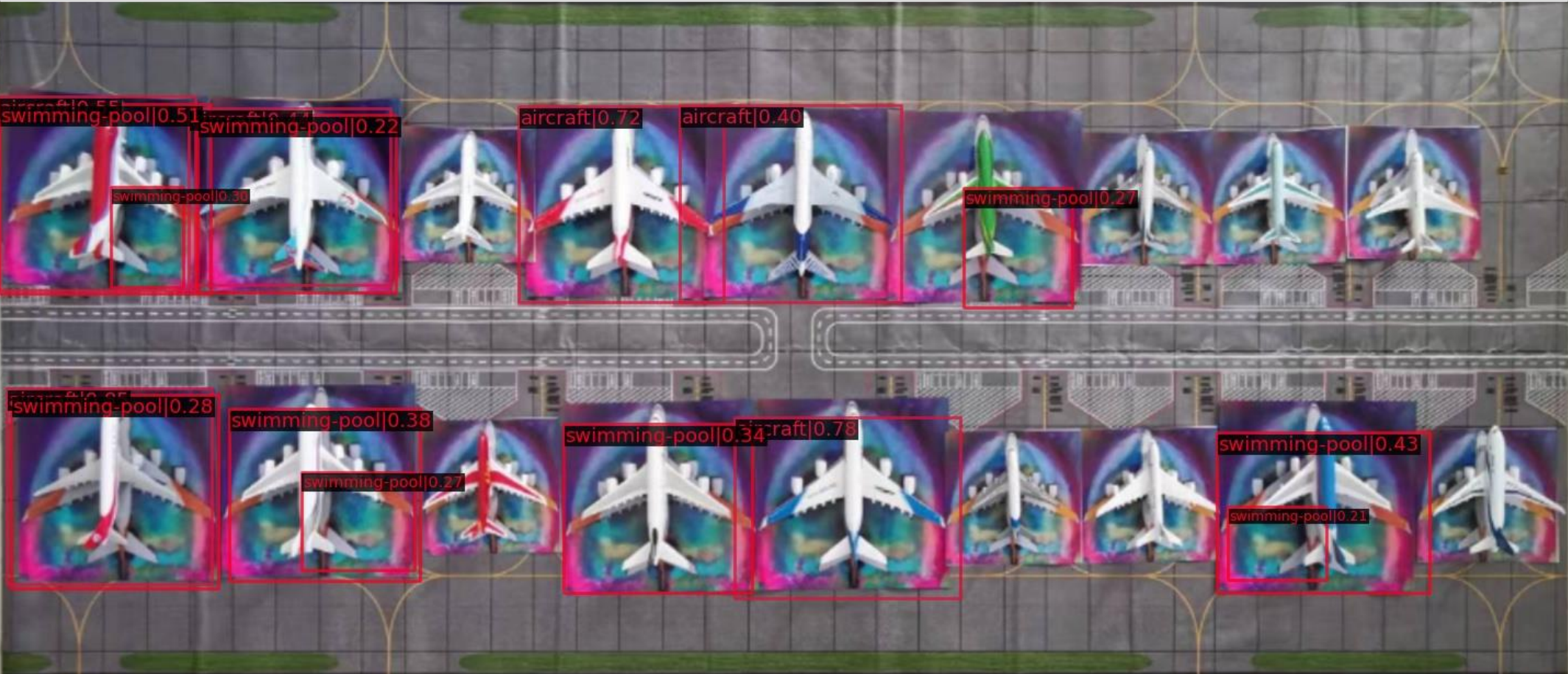}
    \caption{Mask R-CNN}
  \end{subfigure}
  \begin{subfigure}{0.24\linewidth}
    \includegraphics[width=1\linewidth]{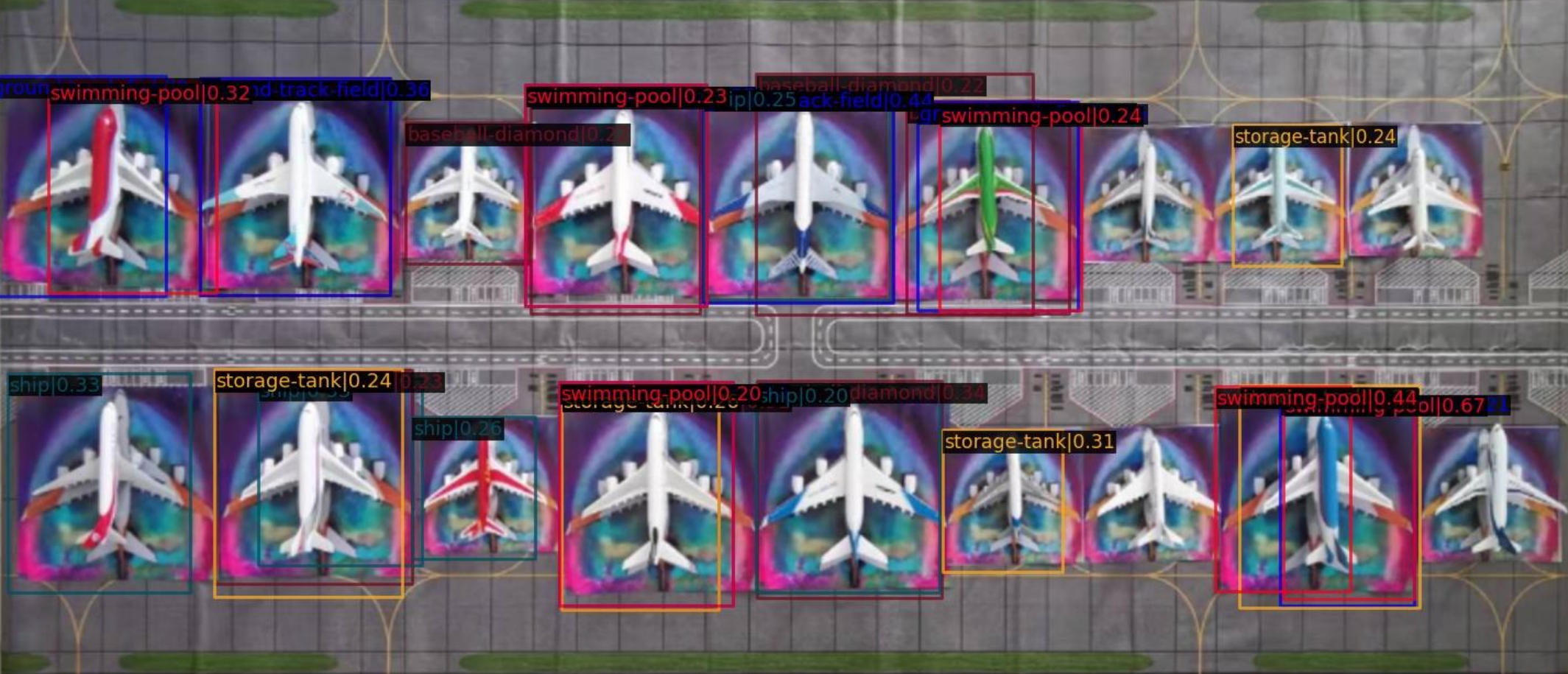}
    \caption{FoveaBox}
  \end{subfigure}
  \begin{subfigure}{0.24\linewidth}
    \includegraphics[width=1\linewidth]{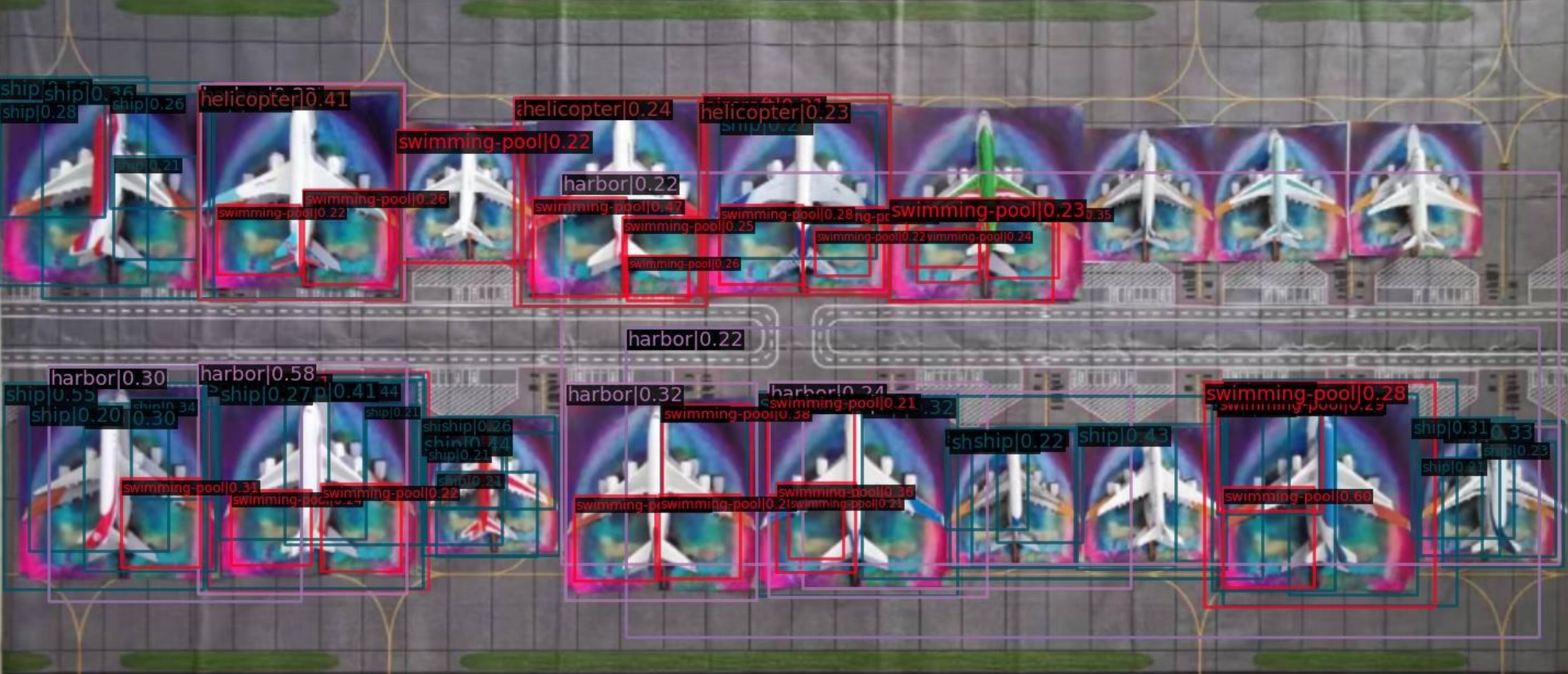}
    \caption{FreeAnchor}
  \end{subfigure}
  \begin{subfigure}{0.24\linewidth}
    \includegraphics[width=1\linewidth]{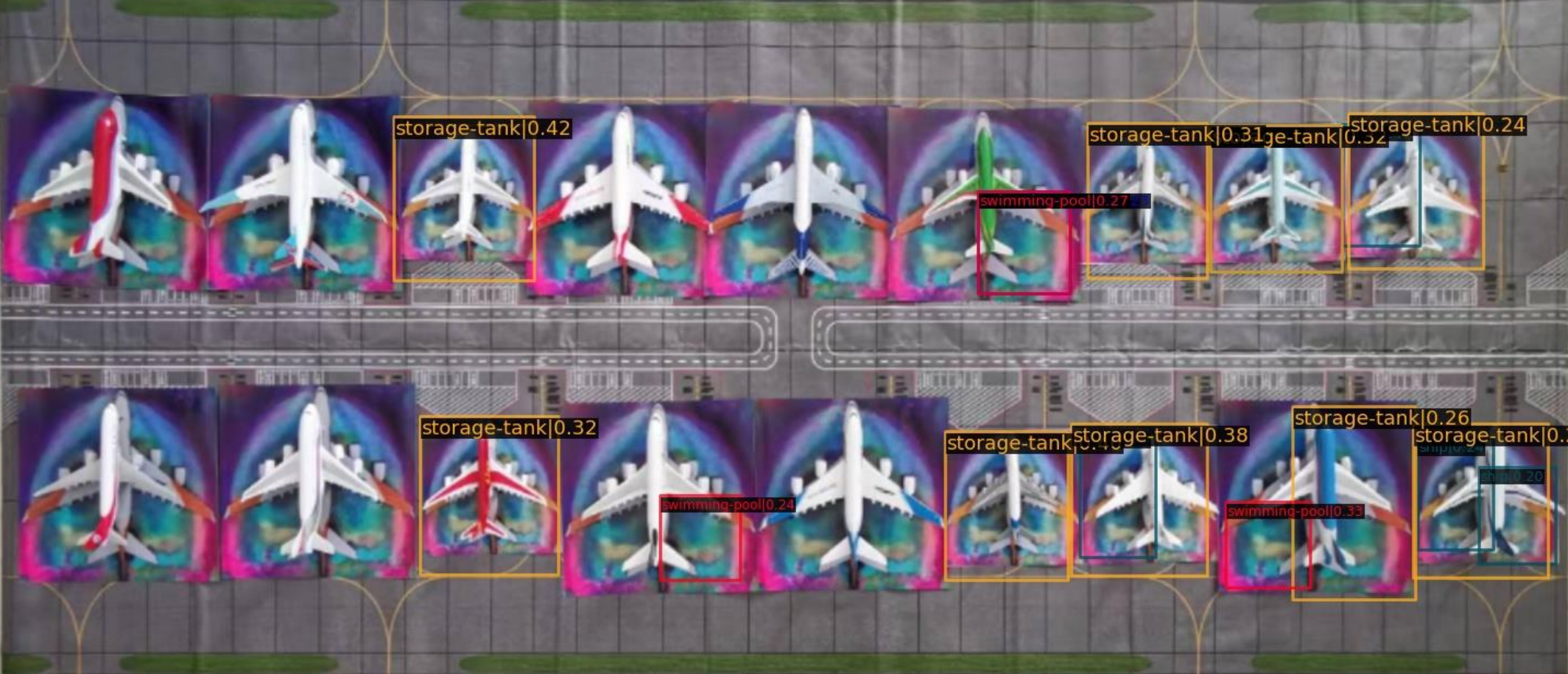}
    \caption{FSAF}
  \end{subfigure}
  \begin{subfigure}{0.24\linewidth}
    \includegraphics[width=1\linewidth]{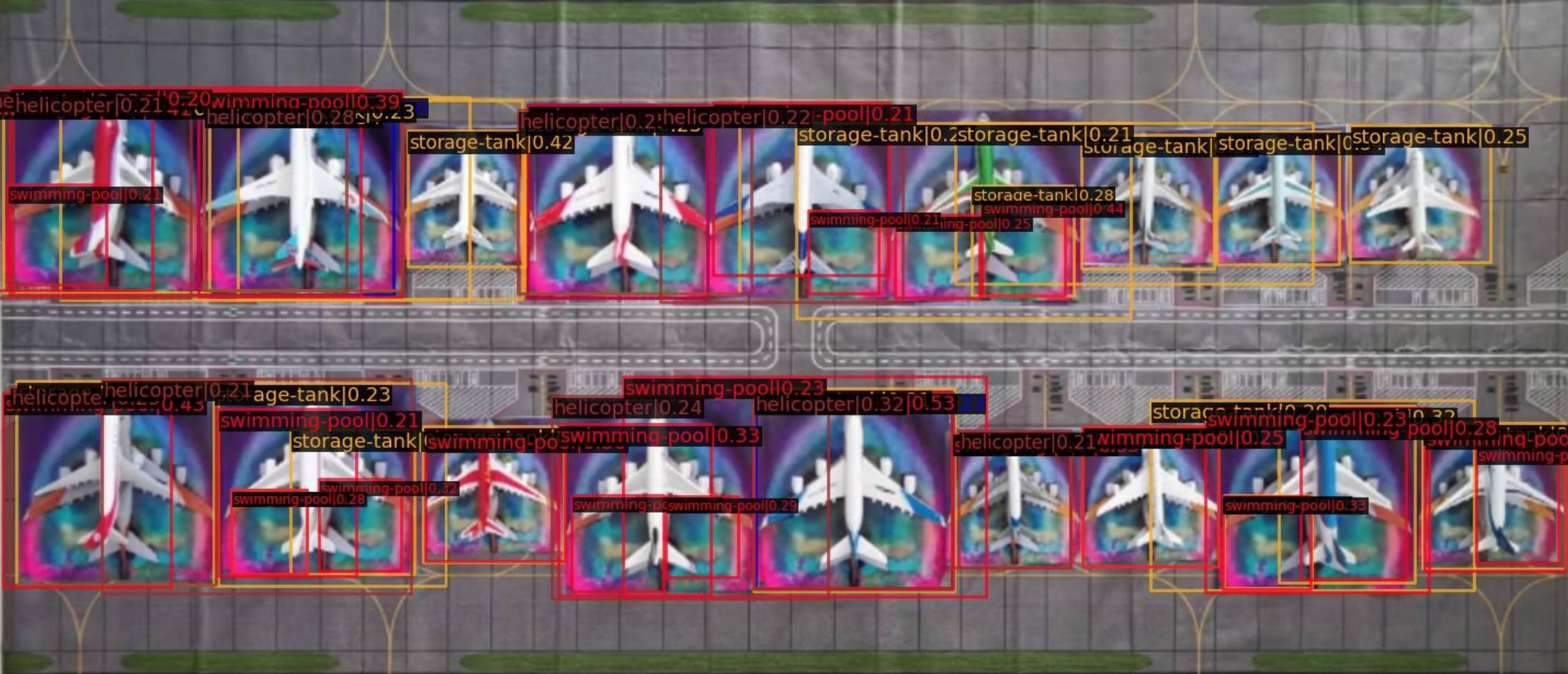}
    \caption{RepPoints}
  \end{subfigure}
  \begin{subfigure}{0.24\linewidth}
    \includegraphics[width=1\linewidth]{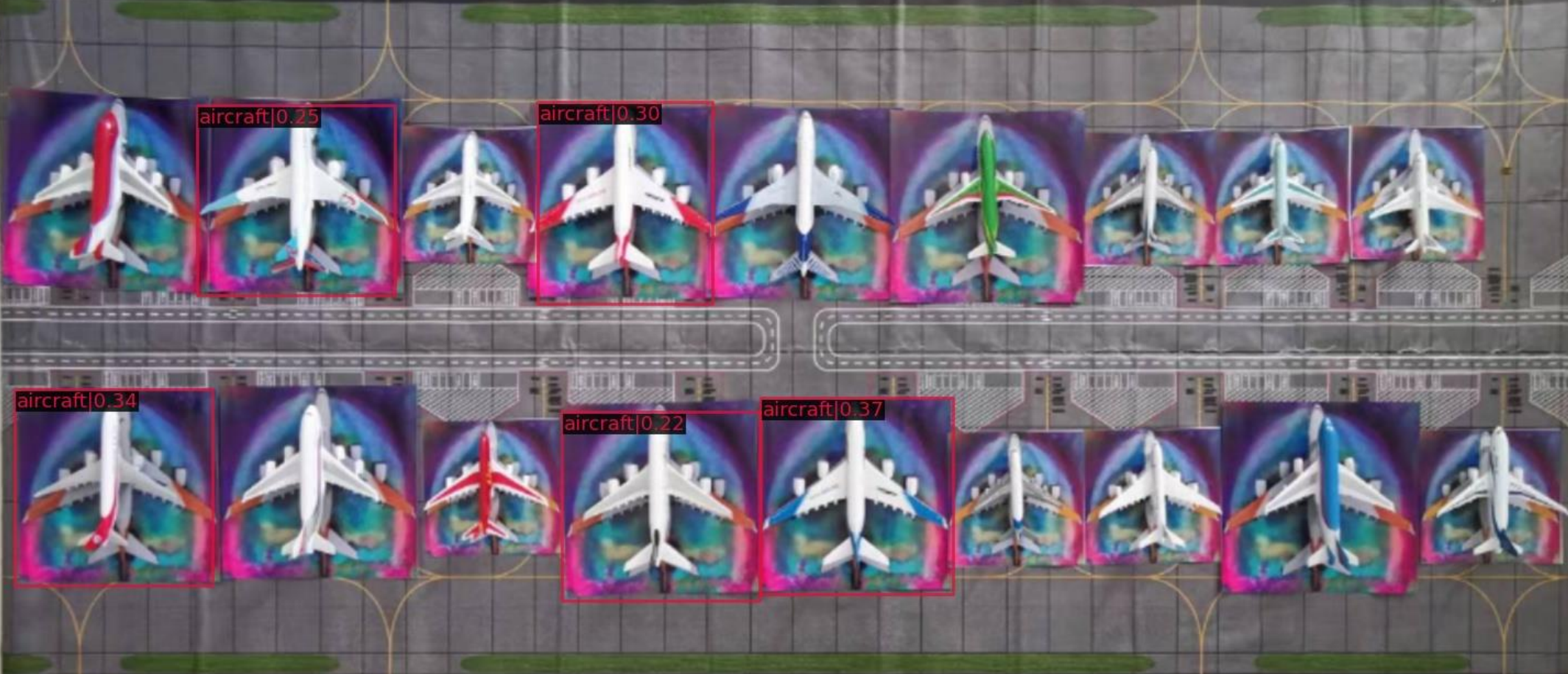}
    \caption{TOOD}
  \end{subfigure}
  \begin{subfigure}{0.24\linewidth}
    \includegraphics[width=1\linewidth]{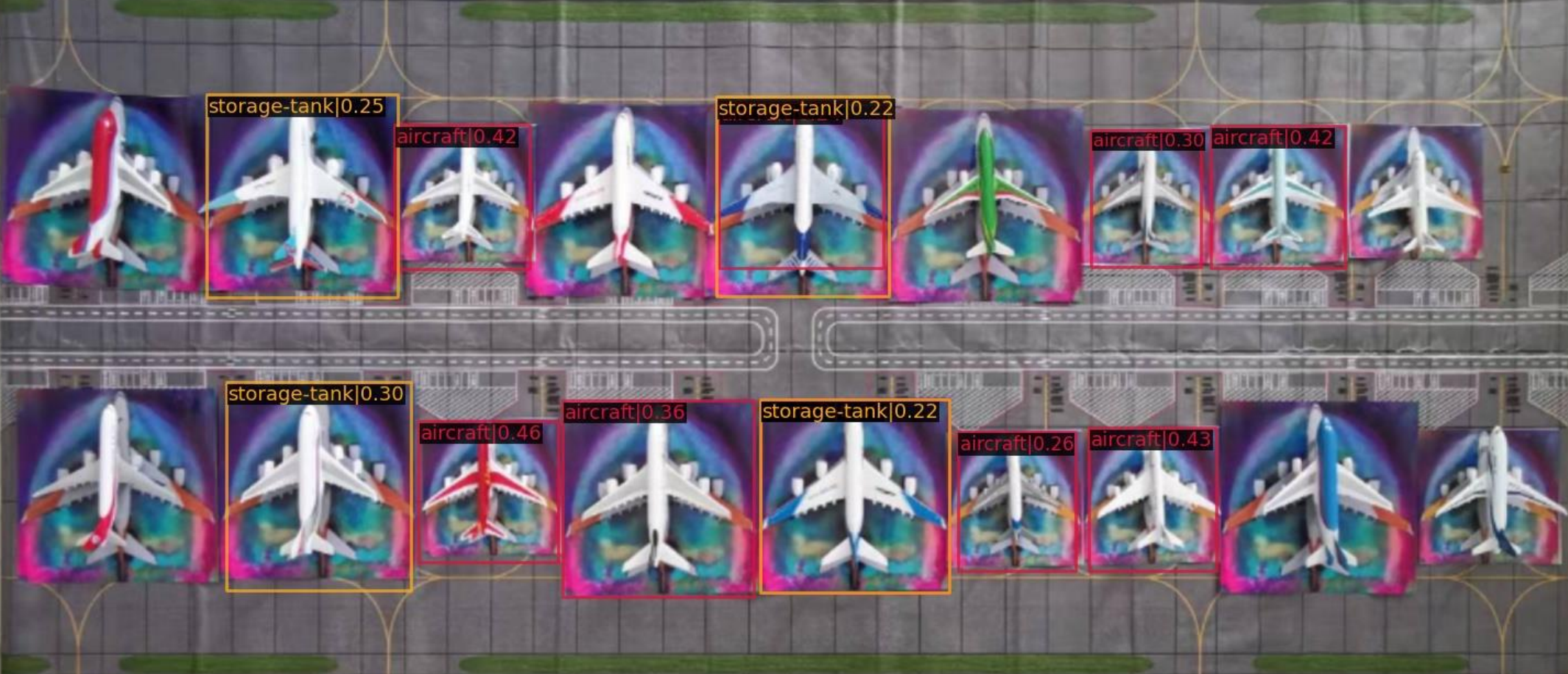}
    \caption{ATSS}
  \end{subfigure}
  \begin{subfigure}{0.24\linewidth}
    \includegraphics[width=1\linewidth]{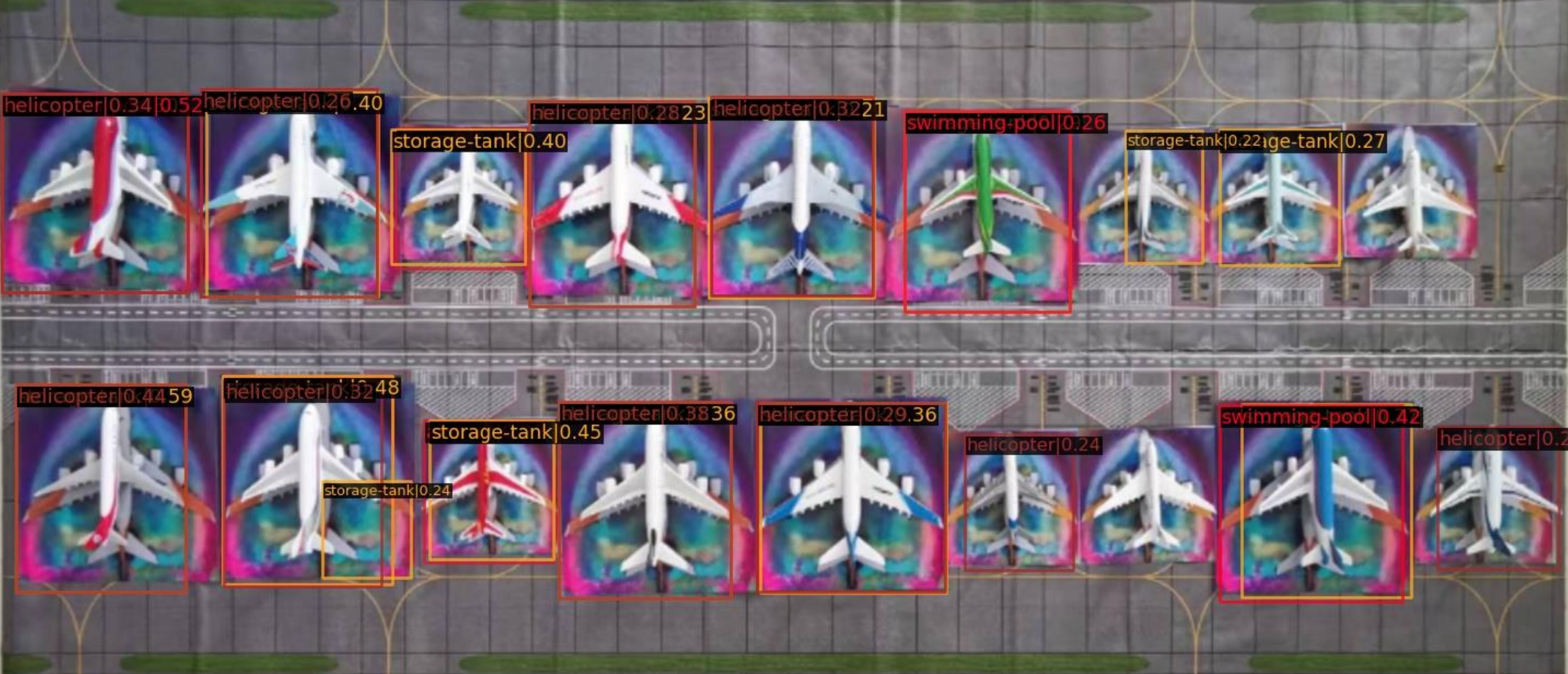}
    \caption{VarifocalNet}
  \end{subfigure}
  \caption{Qualitative results of black-box attacks in the physical world, the proxy model is YOLOv5n, \ie the  transfer attack performance of the contextual adversarial patch trained by YOLOv5n.}
  \label{fig:black_box_physical_attack_qualitative_results}
\end{figure*}

\begin{figure*}[!t]
  \centering
  \begin{subfigure}{0.32\linewidth}
    \includegraphics[width=1\linewidth]{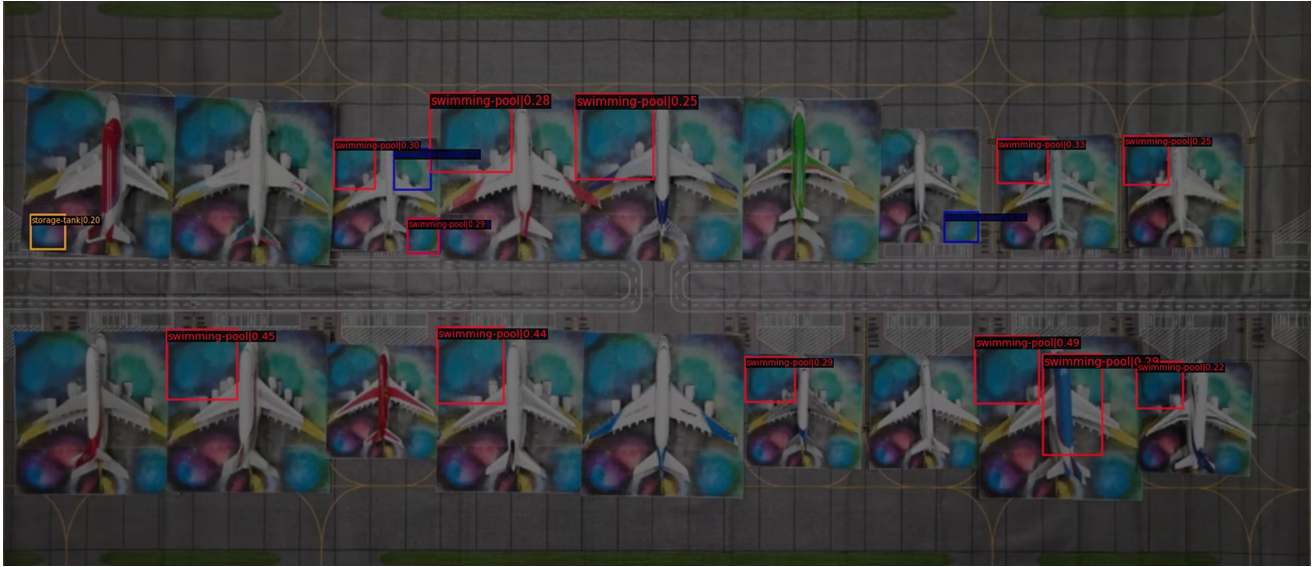}
    \caption{Brightness 1}
  \end{subfigure}
  \begin{subfigure}{0.32\linewidth}
    \includegraphics[width=1\linewidth]{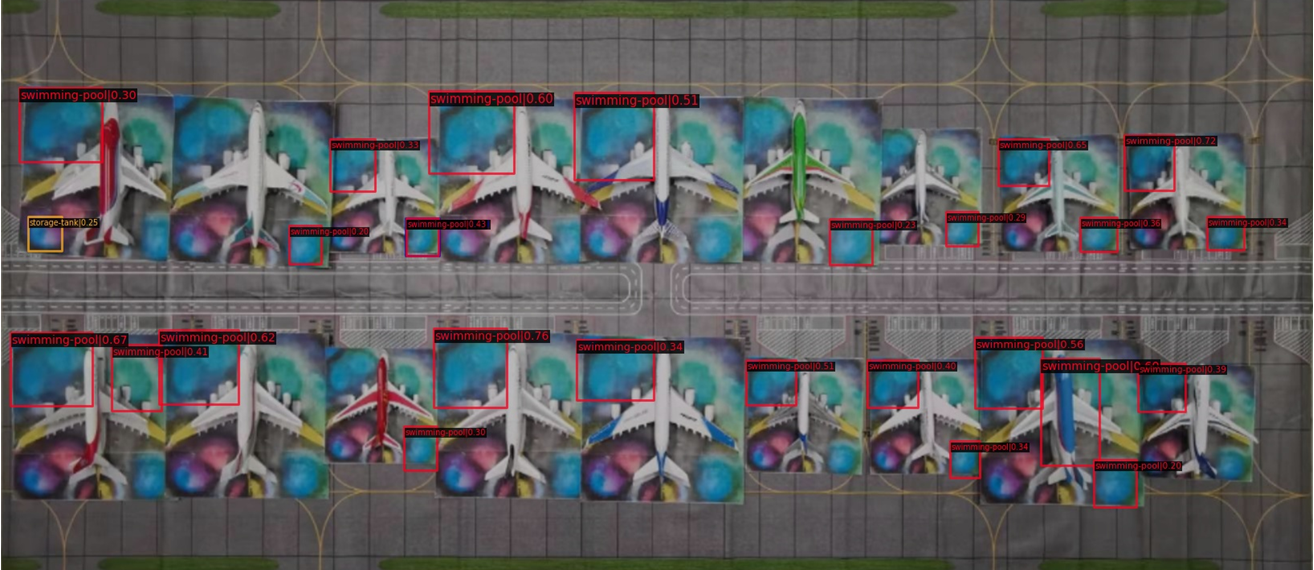}
    \caption{Brightness 2}
  \end{subfigure}
  \begin{subfigure}{0.32\linewidth}
    \includegraphics[width=1\linewidth]{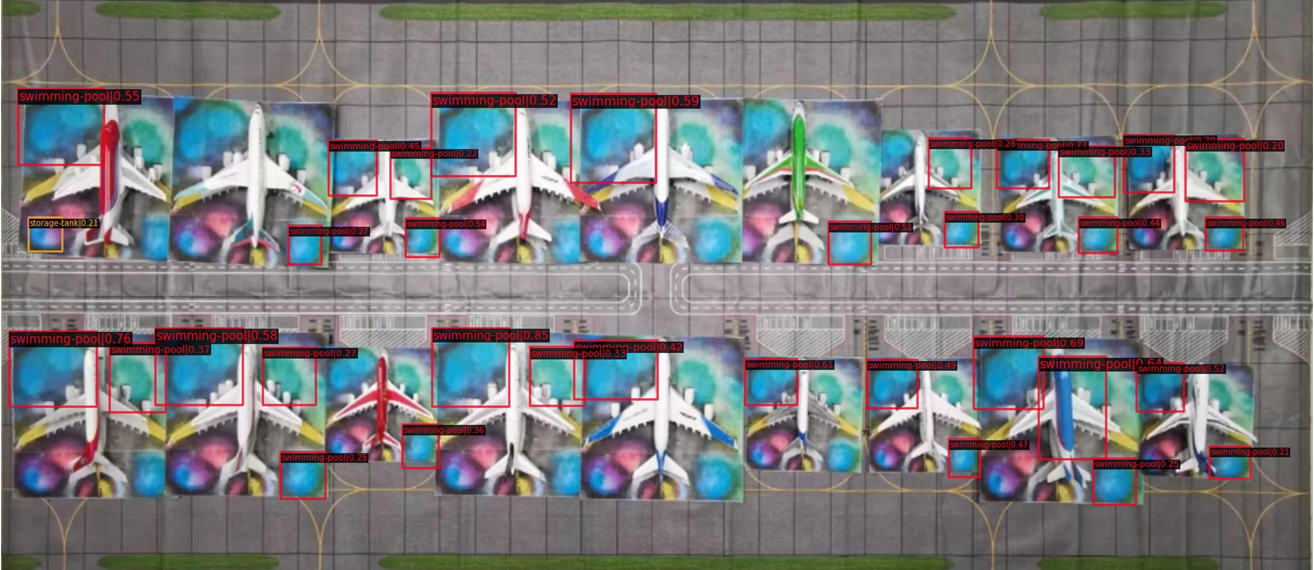}
    \caption{Brightness 3}
  \end{subfigure}
  \begin{subfigure}{0.32\linewidth}
    \includegraphics[width=1\linewidth]{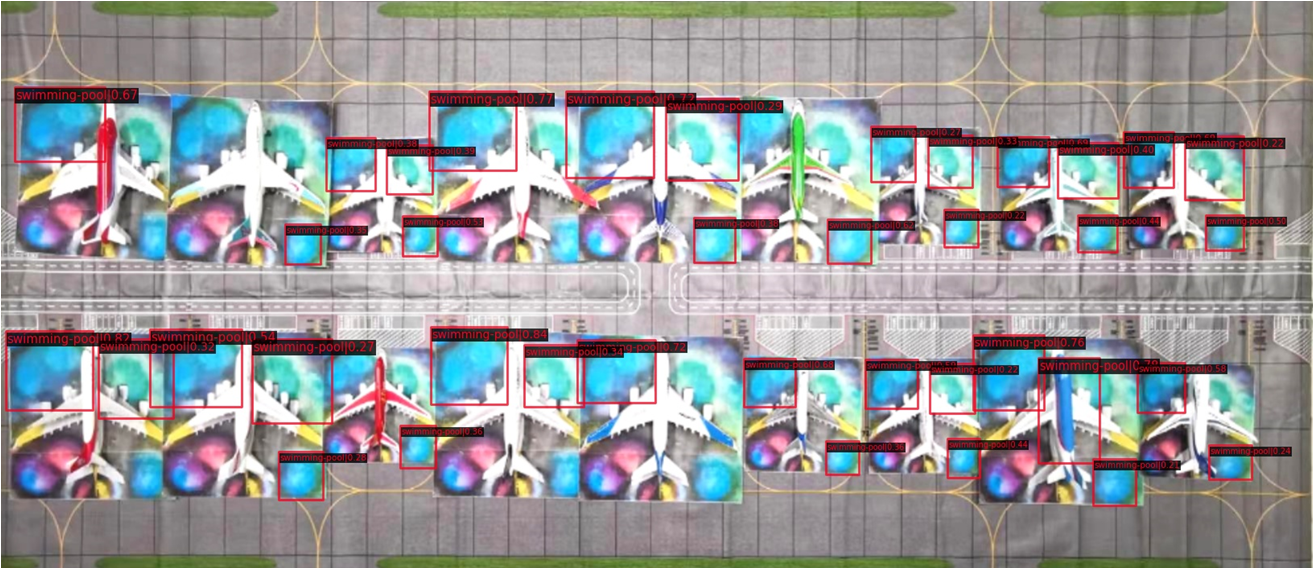}
    \caption{Brightness 4}
  \end{subfigure}
  \begin{subfigure}{0.32\linewidth}
    \includegraphics[width=1\linewidth]{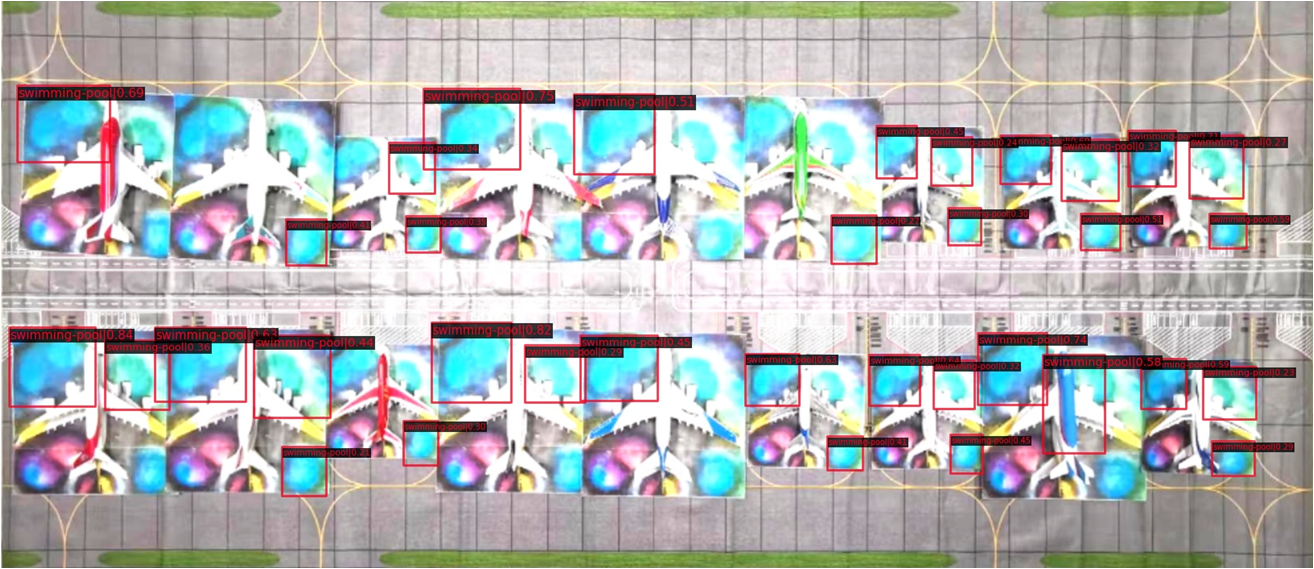}
    \caption{Brightness 5}
  \end{subfigure}
  \begin{subfigure}{0.32\linewidth}
    \includegraphics[width=1\linewidth]{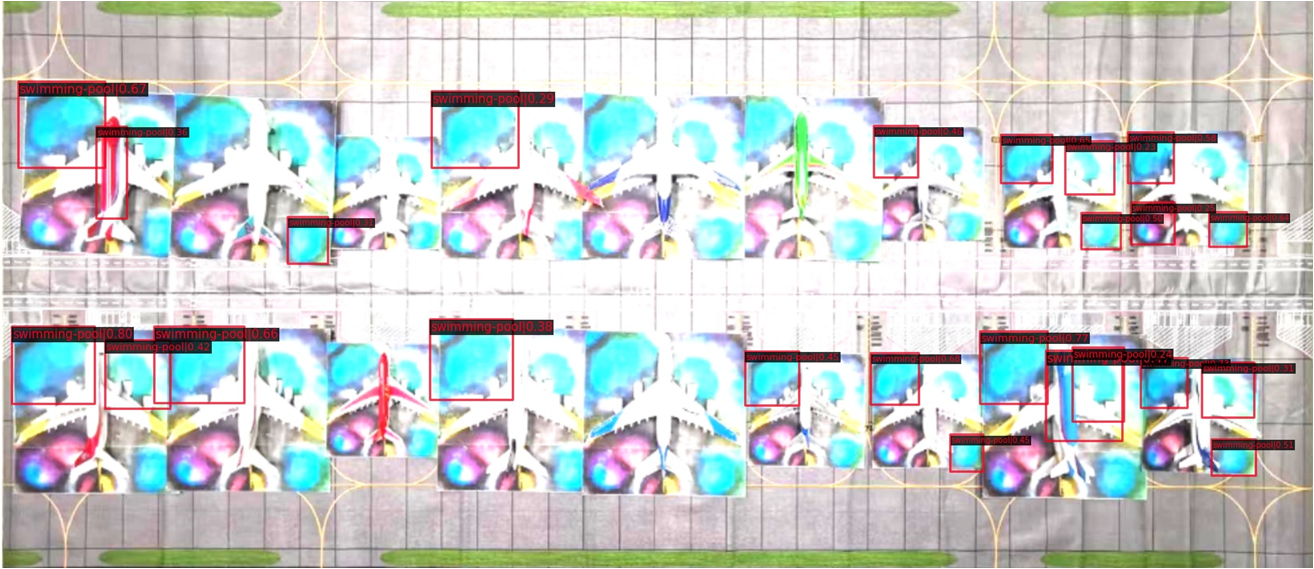}
    \caption{Brightness 6}
  \end{subfigure}
  \caption{Robust physical attacks in varying lighting conditions.}
  \label{fig:varying_lightings}
\end{figure*}

In this section, we perform comprehensive experiments to validate the convincingness of the proposed contextual background attack framework. 
We first outline the experimental settings and then separately report the results of the digital and physical attacks.
Subsequently, we describe the experimental results of the ablation study on total variation.
Finally, we give some possible explanations regarding the unexpectedness of the experimental results.
The video demo of our proposed CBA physical attack method has been released\footnote{\url{https://www.youtube.com/watch?v=wng9LZbQeJA}}\footnote{\url{https://www.youtube.com/shorts/BlBlCNEi_I4}}.
\subsection{Experimental Settings}
\label{section4.1}

\textbf{Datasets}:
Two well-known large-scale datasets, \ie DOTA \cite{xia2018dota} and RSOD\footnote{\url{https://github.com/RSIA-LIESMARS-WHU/RSOD-Dataset-}}, are adopted to train aerial detectors and adversarial patches, respectively, in the experiments.

\textbf{Target models}:
Dozens of mainstream detectors are adopted to validate the effect of the proposed framework, including YOLOv2 \cite{redmon2017yolo9000}, YOLOv3 \cite{redmon2018yolov3}, YOLOv5 \cite{jocher2020yolov5}, SSD \cite{liu2016ssd}, Faster R-CNN \cite{ren2015faster}, Swin Transformer \cite{liu2021swin}, Cascade R-CNN \cite{cai2019cascade}, RetinaNet \cite{lin2017focal}, Mask R-CNN \cite{he2017mask}, FoveaBox \cite{kong2020foveabox}, FreeAnchor \cite{zhang2019freeanchor}, FSAF \cite{zhu2019feature}, RepPoints \cite{yang2019reppoints}, TOOD \cite{feng2021tood}, ATSS \cite{zhang2020bridging}, and VarifocalNet \cite{zhang2021varifocalnet}.

\textbf{Compared methods}:
Two SOTA methods are selected for comparison, including the adversarial patches proposed by Thys \etal \cite{thys2019fooling} and APPA  \cite{lian2022benchmarking}. 
Note that APPA is conducted to generate adversarial patches both on and outside targets of interest, denoting as `APPA (on)' and `APPA (outside)', respectively.
The average precision (AP) is adopted as the quantitative detection metric. 

\textbf{Implementation details}:
During training, $\alpha$ is set to 1.5 to balance the objectiveness loss and total variation.
The learning rate scheduler is started from 0.03, and the thresholds for the intersection of union (IOU) and objective confidence are set as 0.45 and 0.4, respectively.
All the codes are implemented in PyTorch on RTX3090 (24GB) GPUs, and a printer model Color LaserJet Pro MFP M479dw is adopted to print all the adversarial patches generated by different methods.

\subsection{Attack in Digital Domain}
\label{section4.2}

For our proposed CBA, the digital attacks are conducted in the same settings as training, \ie the adversarial patches are also pasted outside the targets detected by aerial detectors in proper size and positions. 
The contextual background patches generated by our CBA are displayed in Fig. \ref{fig:adversarial_aircraft}. 
When trained on different versions of YOLOv5, it generates adversarial patches with similar pattern styles. 
The patches trained by YOLOv3 and YOLOv5 are also somewhat similar, indicating that detectors with similar structures may generate analogous adversarial patches.
However, the generated patches vary greatly when training on other detectors with different structures.

Ten aerial detectors are chosen for the quantitative evaluation.
We set detection results of the clean images as ground truth to calculate the AP, \ie the AP of the clean dataset is $100\%$, by which the targets that the original detector fails to detect won't be counted as a successful attack.
The experimental results are shown in Fig. \ref{fig:table_digital_results}.
It is observed that:

\begin{itemize}
    \item For the attackers, our proposed CBA achieves the best-attacking performance in both white-box and black-box versions for quite a few cases, though the adversarial patch is placed outside the targets of interest, and part area of the patch is sacrificed for physical attack;
    \item For the detectors, YOLOv2 is the easiest to attack with the patches on targets, even in the black-box setting. Various versions of YOLOv5 are robust in diverse attack settings, while Faster R-CNN and SSD are easier to be attacked. In general, Swin Transformer is the most robust detector.
\end{itemize}

\subsection{Attack in Physical World}
\label{section4.3}

\subsubsection{Overall experimental results}

In this paper, the proposed framework is mainly designed to conduct physical attacks, so extensive and rigorous proportionally scaled experiments are performed to verify the effectiveness of the proposed physical attack framework CBA.

In this section, 1:400 proportionally scaled experiments are conducted to verify the attack performance in the physical world. Specifically, we train 20 mainstream aerial detectors as victim models and record the average confidence (threshold set as 0.2, \ie the detected object will be ignored if the detection confidence is lower than 0.2.) of 18 aircraft to compare the physical attack efficacy.
The experimental results are shown in Fig. \ref{fig:table_physical_results}.
It is observed that:

\begin{itemize}
    \item Our CBA results in a considerable number of detectors failing to detect any targets, \ie the values are 0 (highlighted in bold), which barely happens to other methods;
    \item Our CBA can transfer the attack efficacy well between different detectors, even for some hard-to-attack detectors (\eg various versions of YOLOv5), which are only slightly swayed by other methods;
    \item In contrast, different forms of YOLOv5 are still the toughest detectors to be attacked among all approaches, while our elaborated adversarial patches can easily blind them and generalize well between different versions of YOLOv5;
    \item Similar to the digital attack, YOLOv2 is still the weakest detector. However, it seems immune to patches of outside targets in the physical world.
\end{itemize} 

In conclusion, the result area of our CBA in Fig. \ref{fig:table_physical_results} is significantly redder than other areas, indicating that our proposed CBA can achieve excellent attack performance in both white-box and black-box settings and significantly outperform other methods when trained by most detectors.
The fly in the ointment is that the adversarial patch trained by YOLOv2 presents poor attack transferability, and we will discuss this in Sec. \ref{section4.5}.

\subsubsection{Detailed experimental results of white-box attacks}

We report the detailed quantitative and qualitative results in a white-box setting, as shown in Table \ref{table_white_box_attack_results_comparison} and Fig. \ref{fig:white_box_physical_attack_qualitative_results}, respectively. 
It is observed that:

\begin{itemize}
    \item The generated contextual adversarial patches by the proposed CBA can completely blind quite a few aerial detectors, \ie the victim detectors can not recognize any protected targets at all, such as SSD, YOLOv2, YOLOv5n, Cascade R-CNN, RetinaNet, FreeAnchor, FSAF, RepPoints, TOOD, and FoveaBox;
    \item The rest of the detectors can recognize the protected objects correctly, but only with low average confidence under 0.438;
    \item In contrast, most patches generated by the comparison methods can barely misguide the detectors, which can only slightly sway the confidence of the correct detection;
    \item None of the comparison methods can successfully hide all the objects of interest from being perceived.
\end{itemize}

\subsubsection{Detailed experimental results of black-box attacks}

We report the detailed quantitative and qualitative results in a black-box setting, as shown in Table \ref{table_black_box_attack_results_comparison} and Fig. \ref{fig:black_box_physical_attack_qualitative_results}, respectively. 
It is observed that:

\begin{itemize}
    \item Even under the black-box setting, the generated contextual adversarial patches by the proposed CBA can transfer its attack efficacy well between various aerial detectors;
    \item The contextual adversarial patch trained on YOLOv5n successfully protects all the interested targets from being recognized by YOLOv2, Cascade R-CNN, RetinaNet, FSAF, RepPoints, FoveaBox, and VarifocalNet;
    \item Under the attack of our CBA, all the average confidences are lower than 0.208, which significantly outperforms other physical attack methods.
\end{itemize}

\subsubsection{Physical attack robustness}

\begin{figure}
  \centering
  \begin{subfigure}{0.99\linewidth}
    \includegraphics[width=1\linewidth]{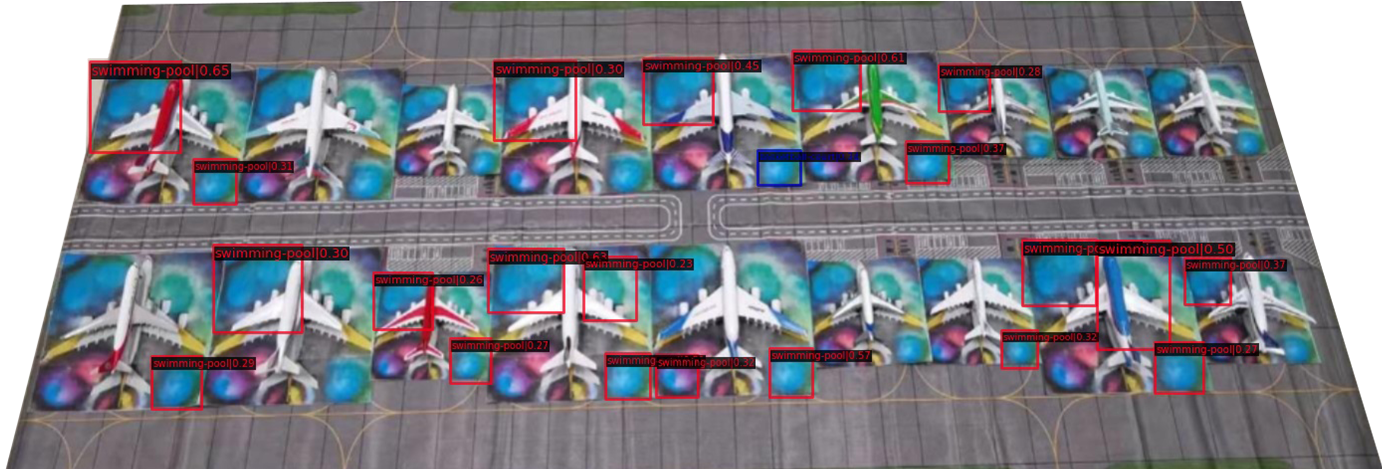}
    \caption{Angle 1}
  \end{subfigure}
  \begin{subfigure}{0.99\linewidth}
    \includegraphics[width=1\linewidth]{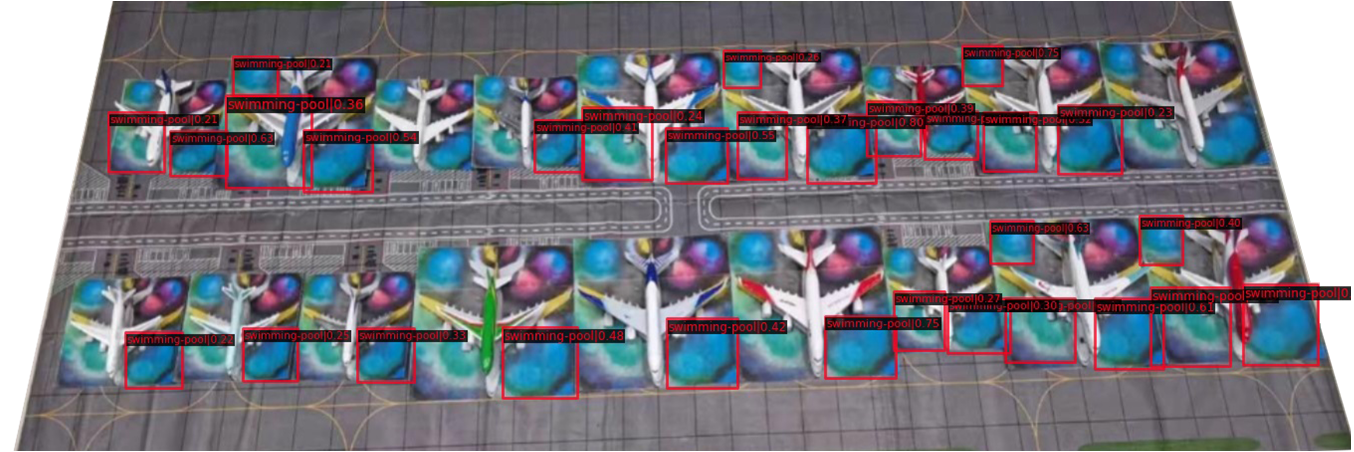}
    \caption{Angle 2}
  \end{subfigure}
  \begin{subfigure}{0.49\linewidth}
    \includegraphics[width=1\linewidth]{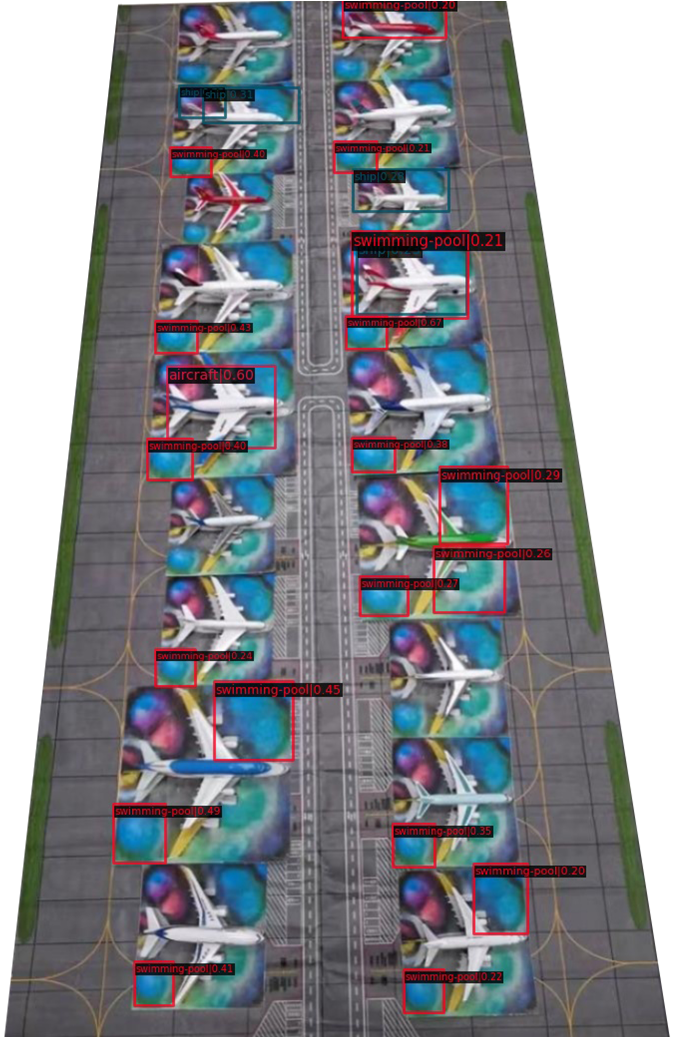}
    \caption{Angle 3}
  \end{subfigure}
  \begin{subfigure}{0.49\linewidth}
    \includegraphics[width=1\linewidth]{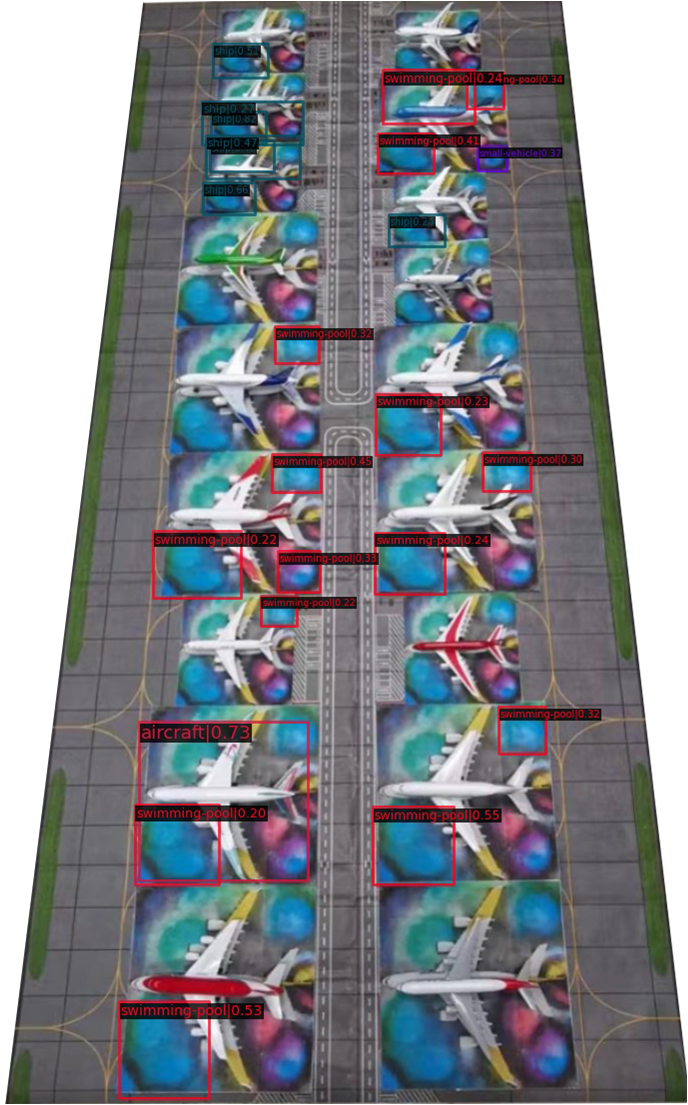}
    \caption{Angle 4}
  \end{subfigure}  
  \caption{Robust physical attacks in varying view angles.}
  \label{fig:varying_angles}
\end{figure}

\begin{figure}[!t]
  \centering
  \includegraphics[width=0.99\linewidth]{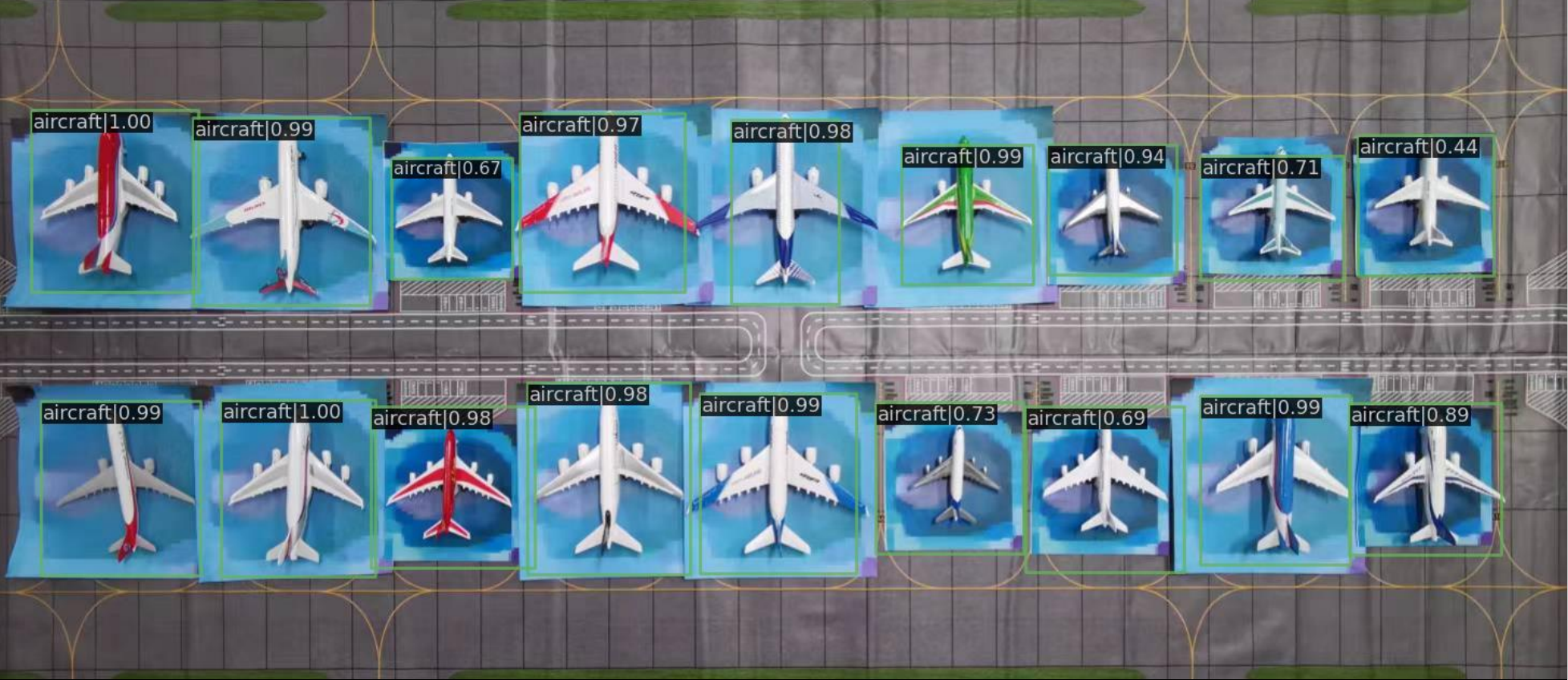}
  \caption{Exclusion study on patch size and location.}
  \label{fig:exclusion_study}
\end{figure}

\begin{figure}
  \centering
  \begin{subfigure}{0.98\linewidth}
    \includegraphics[width=1\linewidth]{cam_clean.pdf}
    \caption{Clean}
  \end{subfigure}
  \begin{subfigure}{0.98\linewidth}
    \includegraphics[width=1\linewidth]{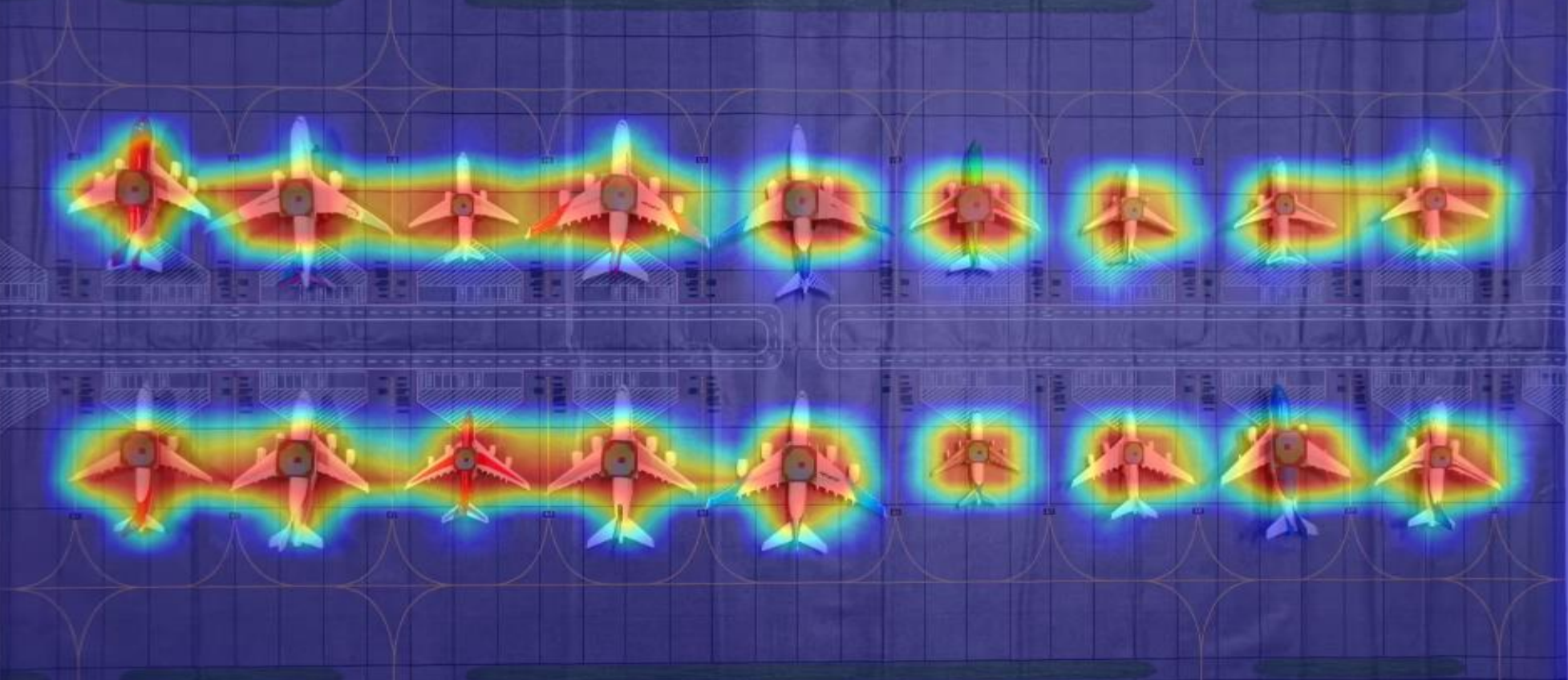}
    \caption{Thys \etal}
  \end{subfigure}
  \begin{subfigure}{0.98\linewidth}
    \includegraphics[width=1\linewidth]{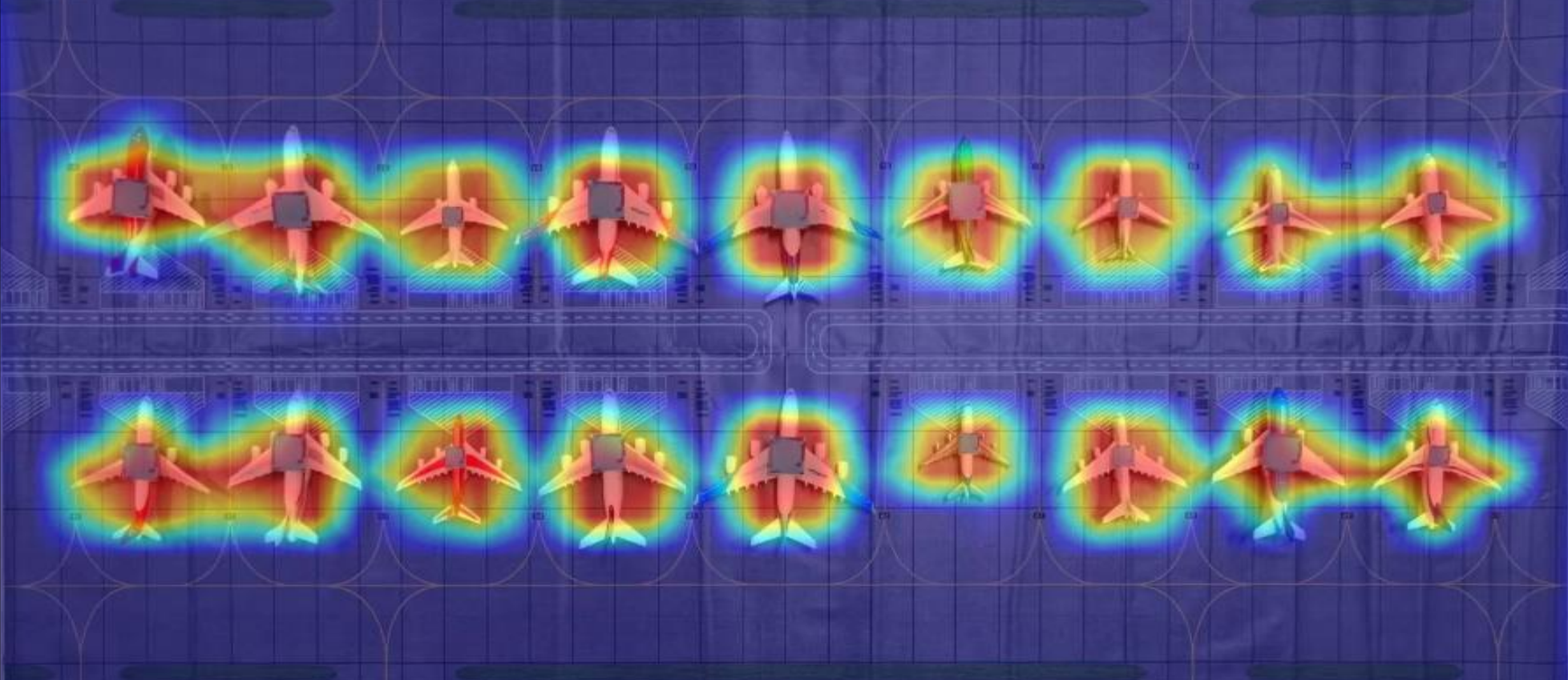}
    \caption{APPA (on)}
  \end{subfigure}
  \begin{subfigure}{0.98\linewidth}
    \includegraphics[width=1\linewidth]{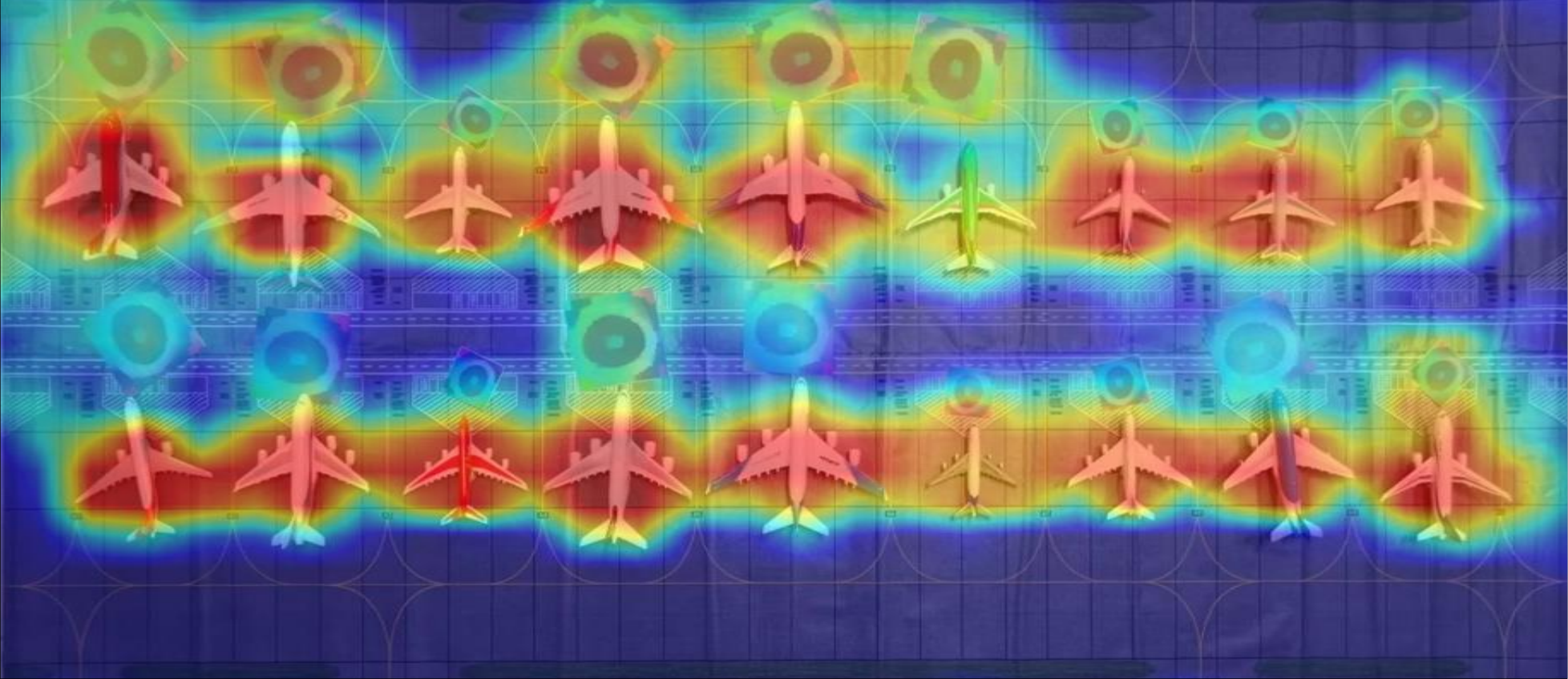}
    \caption{APPA (outside)}
  \end{subfigure}
  \begin{subfigure}{0.98\linewidth}
    \centering
    \includegraphics[width=1\linewidth]{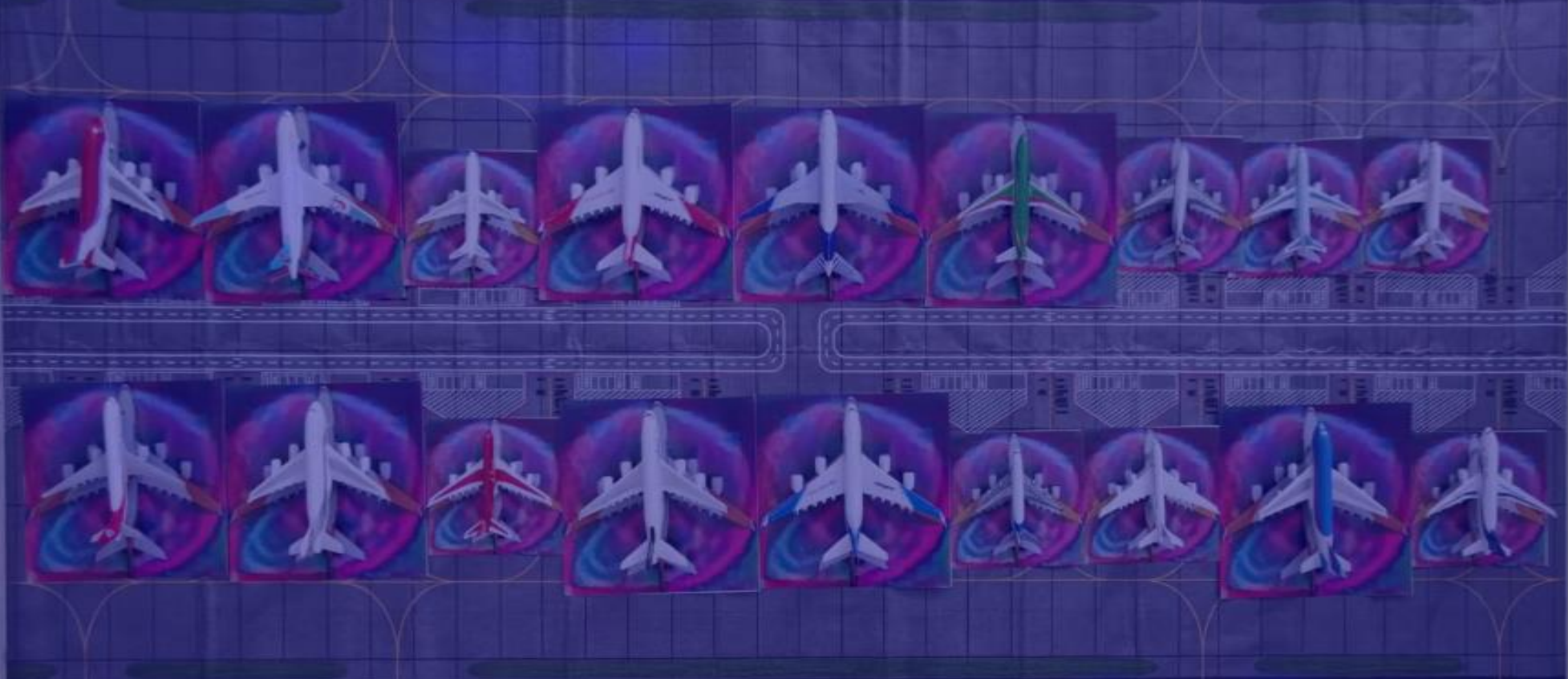}
    \caption{Ours (CBA)}
  \end{subfigure}
  \caption{Visualization of the YOLOv5s' attention before and after different physical attacks. It is observed that our method completely blind the detector, while other methods can barely sway the detector.}
  \label{fig:cam_comparison}
\end{figure}

The robustness of our proposed CBA for the physical attack is further validated by varying imaging angles and lighting conditions. 
The experimental results are shown in Fig. \ref{fig:varying_lightings} and Fig. \ref{fig:varying_angles}, respectively. 
We can observe that our proposed CBA achieves successful attacks over all the aircraft, \ie none of the protected targets are recognized correctly and with prediction confidences higher than 0.2, demonstrating our proposed method's physical attack robustness against dynamic conditions in real-world scenarios.

To exclude the effect of patch location and size, APPA's patches with the same setting as ours are adopted for comparison. It is found from Fig. \ref{fig:exclusion_study} that APPA's patches can barely sway the prediction.
We also visualize the attention map \cite{selvaraju2017grad} of YOLOv5s before and after physical attacks as shown in Fig. \ref{fig:cam_comparison}. It is observed that the proposed CBA can completely blind the powerful detector in the physical world. 

\subsection{Ablation Study}
\label{section4.4}

\begin{figure}
  \centering
  \begin{subfigure}{0.19\linewidth}
    \includegraphics[width=1\linewidth]{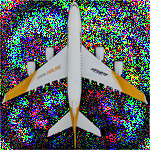}
    \captionsetup{font=footnotesize}
    \caption{$\alpha = 0.0$}
  \end{subfigure}
  \begin{subfigure}{0.19\linewidth}
    \includegraphics[width=1\linewidth]{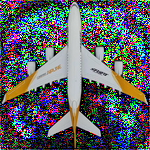}
    \captionsetup{font=footnotesize}
    \caption{$\alpha=0.15$}
  \end{subfigure}
  \begin{subfigure}{0.19\linewidth}
    \includegraphics[width=1\linewidth]{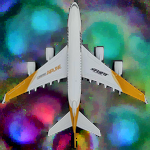}
    \captionsetup{font=footnotesize}
    \caption{$\alpha = 1.5$}
  \end{subfigure}
  \begin{subfigure}{0.19\linewidth}
    \includegraphics[width=1\linewidth]{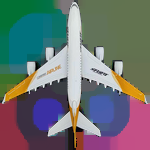}
    \captionsetup{font=footnotesize}
    \caption{$\alpha = 15$}
  \end{subfigure}
  \begin{subfigure}{0.19\linewidth}
    \includegraphics[width=1\linewidth]{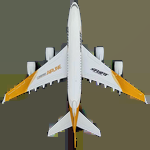}
    \captionsetup{font=footnotesize}
    \caption{$\alpha = 150$}
  \end{subfigure}
  \caption{Adversarial aircraft crafted with different $\alpha$ on SSD.}
  \label{fig:ablation}
\end{figure}

\begin{figure}[!t]
  \centering
  \includegraphics[width=0.99\linewidth]{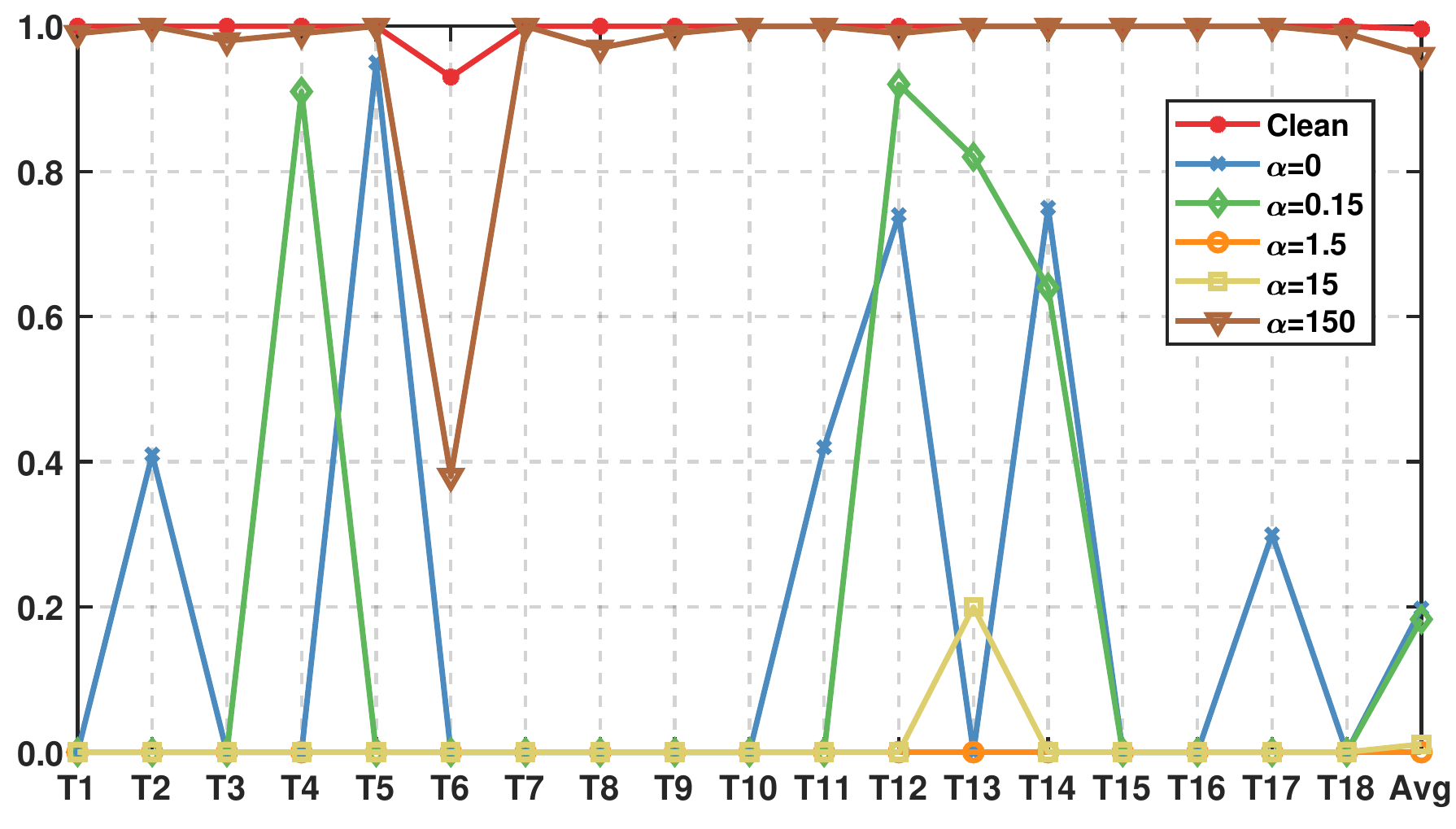}
  \caption{Ablation study on total variation.}
  \label{fig:ablation_study}
\end{figure}

This part discusses the influence of total variation on the physical attack.
We evaluate the effectiveness of smoothness constriction on SSD in physical scenarios.
Specifically, we vary $\alpha$ from $[0.0,0.15,1.5,15,150]$ to generate corresponding adversarial patches, as shown in Fig. \ref{fig:ablation}.
In addition, we perform physical attack experiments in the proportionally scaled physical scenario, including 18 aircraft targets (T1-T18), to quantitatively compare the attack efficacy.
The detection results are shown in Fig. \ref{fig:ablation_study}, and we can observe that:

\begin{itemize}
    \item If $\alpha$ is too small or equal to 0, the  generated patch is not smooth enough, which may cause a significant loss of the attack efficacy during physical-digital transformation;
    \item If $\alpha$ is too large, the pattern of the adversarial patch will be simplified, which has a critical influence on the attack efficacy of the generated patch.
\end{itemize}

Consequently, we choose $\alpha=1.5$ to balance the two parts of the total loss in our experiments.

\subsection{Discussion}
\label{section4.5}

Surprisingly, the adversarial patch against YOLOv2 shows weak transferability.
We try to explain this phenomenon from a different view.
Training adversarial patches are similar to training networks.
The only difference is that pixels in patches update during training adversarial patches while parameters of the network update during training networks.
Therefore, the adversarial patches are influenced by training samples, victim network models, and optimizing strategy.
Consequently, when the training samples and optimizing process are settled, 
the victim model is crucial in generating adversarial patches.
Thus, for poor detectors, such as YOLOv2, a robust attack method can only learn limited information, which may be just enough for the white-box attack but not enough to attack a more powerful model.
Similarly, the above analysis may also explain why adversarial patches of different versions of YOLOv5 own a similar pattern while differing from the styles of other patches. 

\section{Conclusion}

This paper proposes a brand-new physical attack framework against aerial detection based on contextual adversarial patches. 
The target of interest, \ie aircraft, is adopted to mask the contextual adversarial patches, and the pixels outside the mask area are optimized through iterations.
Moreover, the contextual adversarial patches are forced to be outside targets during training, by which the detected targets of interest, both on and outside patches, contribute to improving the attack's effectiveness of the crafted contextual patches.
Extensive and rigorous experiments have been conducted to validate the proposed physical attack framework's effectiveness. 
This demonstrates that our elaborated contextual adversarial patches are gifted with solid fooling efficacy to hide objects on or outside patches.
In addition, the elaborately crafted adversarial patches can dramatically fool aerial detectors in dynamic physical scenarios, such as varying lighting conditions and view angles, and consistently outperform existing methods in both white-box and black-box settings.

\bibliographystyle{IEEEtran}
\bibliography{references}

\end{document}